\documentclass{article} 
\usepackage{iclr2027_conference,times}

\usepackage[utf8]{inputenc} 
\usepackage[T1]{fontenc}    
\usepackage{url}            
\usepackage{booktabs}       
\usepackage{amsfonts}       
\usepackage{nicefrac}       
\usepackage{microtype}      
\usepackage{xcolor}         
\usepackage{hyperref}      
\newcommand{\xhdr}[1]{{\noindent\bfseries #1}.}

\usepackage{xspace}
\newcommand{\method}{\texttt{\upshape ElasticMem}\xspace} 

\usepackage{amsmath}
\usepackage{amssymb}
\usepackage{mathtools}
\usepackage{amsthm}
\usepackage{enumitem}

\usepackage{multirow}
\usepackage{makecell}
\usepackage{pifont}       
\usepackage{bbding}      
\usepackage{subcaption}
\usepackage{tabularx}
\usepackage{longtable}
\usepackage{mdframed}
\usepackage[most]{tcolorbox}
\usepackage{adjustbox}
\usepackage{wrapfig}
\usepackage{titletoc}
\usepackage{booktabs, array,colortbl,graphicx}

\definecolor{table-blue}{RGB}{173, 216, 230}
\definecolor{row-highlight}{RGB}{220, 230, 242}
\definecolor{qwen-header}{RGB}{235, 240, 248}
\definecolor{llama-header}{RGB}{248, 244, 235}
\definecolor{section-gray}{RGB}{245, 245, 245}
\definecolor{group-header}{RGB}{230, 235, 245}

\definecolor{zhz_gray}{rgb}{0.8,0.8,0.8}
\definecolor{darkgreen}{rgb}{0.0, 0.5, 0.0} 
\definecolor{forestgreen}{RGB}{34, 139, 34}
\definecolor{darkred}{rgb}{0.5, 0.0, 0.0}   

\usepackage{algorithm}
\usepackage{algpseudocode}

\definecolor{row-alt}{RGB}{248, 250, 253}      
\definecolor{hl-llama}{RGB}{234, 242, 255}     
\definecolor{hl-qwen}{RGB}{236, 250, 240}      

\definecolor{morandiblue}{RGB}{200,210,225}
\definecolor{morandired}{RGB}{232,205,200}
\definecolor{groupgray}{gray}{0.95}

\definecolor{darkblue}{rgb}{0, 0, 0.5}
\definecolor{darkred}{rgb}{0.72, 0.22, 0.27}
\definecolor{lightblue}{RGB}{129, 209, 241}
\definecolor{forestgreen}{RGB}{34, 139, 34}
 \hypersetup{
    colorlinks=true,
    citecolor=darkblue,
    linkcolor=darkblue,
    urlcolor=darkblue,
}

\newcommand\DoToC{%
  \startcontents
  \begingroup
  \setcounter{tocdepth}{2}
  \printcontents{}{1}{\textbf{Contents of Appendix}\vskip3pt\hrule\vskip5pt}
  \vskip3pt\hrule\vskip5pt
  \endgroup
}

\usepackage{hyperref}
\usepackage{fontawesome5}

\hypersetup{hidelinks}
\usepackage{url}

\title{ElasticMem: Latent Memory as a Learnable Resource for LLM Agents}

\author{%
\textbf{Tao Feng}\textsuperscript{1},\;\;
\textbf{Chongrui Ye}\textsuperscript{1}, \;\;
\textbf{Fangxu Yu}\textsuperscript{2}, \;\;
\textbf{Tianyang Luo}\textsuperscript{\textbf{1}},\;\;
\textbf{Jingjun Xu}\textsuperscript{\textbf{1}},\;\;
\textbf{Xueqiang Xu}\textsuperscript{\textbf{1}}, \\
\textbf{Haozhen Zhang}\textsuperscript{\textbf{3}},\;\; 
\textbf{Weizhi Zhang}\textsuperscript{\textbf{4}},\;\; 
\textbf{Zijie Lei}\textsuperscript{\textbf{1}},\;\;
\textbf{Jiaxuan You}\textsuperscript{\textbf{3}}\;\;
\\
\textsuperscript{1}University of Illinois Urbana-Champaign,\;\;
\textsuperscript{2}University of Maryland, College Park,\;\;\\
\textsuperscript{3}Nanyang Technological University,\;\;
\textsuperscript{3}University of Illinois Chicago
}

\iclrfinalcopy 
\ificlrfinal\else\fi
\begin{document}

\maketitle

\begin{abstract}
Long-term memory is essential for LLM agents to reason coherently across extended interactions, personalize responses, and reuse past experience. However, existing memory-augmented methods typically treat memory as a fixed resource: text-space approaches concatenate retrieved memories into the context window, causing substantial token overhead and sensitivity to noisy evidence, while latent-space approaches reduce textual cost but still rely on rigid retrieval or fixed-capacity memory interfaces. This creates a mismatch between query-dependent memory utility and fixed memory allocation. We propose \method, a memory-augmented LLM framework that learns to use memory as an elastic latent resource. \method builds an offline latent memory bank with retrieval keys and content caches, retrieves memories adaptively from the reasoner's hidden state, assigns each retrieved memory a variable latent budget through a learned policy, and injects selected latent states as soft memory tokens for generation. The full memory-use process is optimized with downstream task rewards through group-relative policy optimization. We evaluate \method on MemorySuite, covering memory-intensive QA and embodied agent control. Across Qwen2.5-3B-Instruct and Qwen2.5-7B-Instruct backbones, \method improves weighted-average QA accuracy by 26.2\% and 24.6\% and ALFWorld success rate by 66.3\% and 27.2\% relative to the strongest baselines, while consuming the fewest tokens on ALFWorld on average. Ablations and qualitative analyses further show that adaptive retrieval and elastic budget allocation help \method prioritize useful evidence and transferable plans beyond rigid cosine similarity.
\end{abstract}

\begin{center}
    \href{https://ulab-uiuc.github.io/ElasticMem/}{\textcolor{darkblue}{\faGlobe\ \texttt{Website}}}
    \quad
    \href{https://github.com/ulab-uiuc/ElasticMem}{\textcolor{darkblue}{\faGithub\ \texttt{Code}}}
\end{center}

\section{Introduction}

\vspace{-1.8mm}

Long-term memory has become essential for enabling large language model (LLM) agents to reason coherently across extended interactions, personalize responses across sessions, and reuse past experience in long-horizon tasks \citep{park2023generative,shinn2023reflexion,zhong2024memorybank,wang2023voyager,hu2025hiagent,zhang2025survey}.
In realistic settings, an agent may need to retrieve user preferences from past conversations \citep{sun2025preference,pan2025memory}, track facts scattered across long dialogue histories \citep{du2025memory,tavakoli2025beyond}, or reuse procedural experience from previous trajectories \citep{zheng2023synapse,mi2026procmem}.
Thus, memory is a core mechanism for extending LLM agents beyond the current context window, yet retrieved memories must compete for the reasoner's limited context capacity \citep{yan2025memory}.

Despite this importance, existing memory-augmented LLM systems largely treat memory as a fixed resource, as summarized in Table \ref{Tab:intro}.
Text-space memory methods store memories as natural-language records and retrieve, summarize, update, or organize them before injecting selected memories into the prompt \citep{zhong2024memorybank,xu2025mem,fang2025lightmem,chhikara2025mem0,kang2025memory,fang2025memp,langchain2024langmem}.
While interpretable, this paradigm couples memory use with prompt length: irrelevant or redundant memories still consume context tokens, and useful memories may be diluted by noisy textual evidence.
Latent-space memory methods reduce textual overhead by compressing memory into continuous vectors or soft tokens \citep{zhang2025memgen,chevalier2023adapting,wang2025m+}.
However, they typically expose a rigid memory interface: retrieval is often based on fixed similarity scores, and each retrieved or compressed memory receives a uniform or largely pre-defined latent capacity \citep{karpukhin2020dense,reimers2019sentence,zhang2026evoking}.
As a result, a memory chunk that is lexically similar to the query may be retrieved even if it provides little useful evidence, while a less similar chunk containing the key fact or transferable plan may be underused.
This leaves a central gap: memory usefulness is query- and task-dependent, yet both which memories are retrieved and how much capacity each receives are fixed by design choices rather than learned from utility.

This mismatch motivates a different view of memory use.
Rather than asking how to retrieve more memories or compress all memories equally, we ask:
\textit{how can an LLM agent learn to retrieve and allocate memory adaptively, so that useful memories receive more representational budget while unhelpful memories are compressed or suppressed?}
Answering this question requires addressing three challenges.
\textbf{\ding{182} Retrieval should adapt to the current query and the reasoner's internal state.}
Most existing memory systems rely on rigid similarity-based retrieval, where memories are selected by a fixed encoder or surface semantic matching.
However, semantic similarity does not necessarily indicate downstream utility: a memory can be topically similar but unhelpful, while a less similar memory may contain the key evidence or transferable plan \citep{gan2024similarity,dai2025seper,xu2025training,shi2023large,liu2024lost}.
\textbf{\ding{183} Memory capacity should be allocated adaptively.}
Different retrieved chunks play different roles: some contain direct evidence, some provide reusable procedures, some are redundant, and some are misleading.
A fixed latent budget cannot distinguish these cases \citep{yan2024corrective}.
\textbf{\ding{184} Retrieval and allocation should be learned from task outcomes.}
Since the usefulness of a memory is only revealed after generation or interaction, the memory system should learn which memories to retrieve and how much capacity to assign based on downstream feedback rather than similarity scores alone \citep{asai2023self,gao2024smartrag}.

\begin{table}[t]
    \caption{\textbf{Comparison with representative memory-augmented LLM paradigms.}
    Text-space memory methods retrieve and inject natural-language memories, which are interpretable but token-expensive.
    Existing latent-space methods reduce textual overhead, but they still rely on fixed-capacity memory interfaces or rigid retrieval mechanisms.
    \method performs reasoner-state adaptive retrieval and allocates elastic latent capacity to each retrieved memory.}
    \vspace{0mm}
    \label{Tab:intro}
    \centering
    \setlength{\tabcolsep}{3pt}
    \renewcommand{\arraystretch}{1.1}
    \small
    \begin{adjustbox}{max width=\linewidth}
    \begin{tabular}{llcccc}
        \toprule
        \textbf{Paradigm}
        & \textbf{Representative Methods}
        & \makecell{\textbf{Memory}\\\textbf{Space}}
        & \makecell{\textbf{Token}\\\textbf{Efficient}}
        & \makecell{\textbf{Reasoner-State}\\\textbf{Adaptive}\\\textbf{Retrieval}}
        & \makecell{\textbf{Elastic}\\\textbf{Capacity}} \\
        \midrule
        \makecell[l]{Textual long-term\\memory}
        & \makecell[l]{MemoryBank~\citep{zhong2024memorybank} /\\Mem0~\citep{chhikara2025mem0} /\\MemoryOS~\citep{kang2025memory}}
        & Text & \textcolor{red}{\ding{55}} & \textcolor{red}{\ding{55}} & \textcolor{red}{\ding{55}} \\
        \addlinespace[2pt]
        \makecell[l]{Textual agent\\memory}
        & \makecell[l]{A-MEM~\citep{xu2025mem} /\\LightMem~\citep{fang2025lightmem} /\\MemP~\citep{fang2025memp}}
        & Text & \textcolor{red}{\ding{55}} & \textcolor{red}{\ding{55}} & \textcolor{red}{\ding{55}} \\
        \addlinespace[2pt]
        \makecell[l]{Fixed latent\\compression}
        & AutoCompressor~\citep{chevalier2023adapting}
        & Latent & \textcolor{forestgreen}{\ding{51}} & \textcolor{red}{\ding{55}} & \textcolor{red}{\ding{55}} \\
        \addlinespace[2pt]
        \makecell[l]{Latent memory\\retrieval}
        & \makecell[l]{MemGen~\citep{zhang2025memgen} /\\M+~\citep{wang2025m+}}
        & Latent & \textcolor{forestgreen}{\ding{51}} & \textcolor{red}{\ding{55}} & \textcolor{red}{\ding{55}} \\
        \cellcolor{cyan!10}\makecell[l]{Elastic latent\\memory}
        & \cellcolor{cyan!10}\method (ours)
        & \cellcolor{cyan!10}Latent & \cellcolor{cyan!10}\textcolor{forestgreen}{\ding{51}} & \cellcolor{cyan!10}\textcolor{forestgreen}{\ding{51}} & \cellcolor{cyan!10}\textcolor{forestgreen}{\ding{51}} \\
        \bottomrule
    \end{tabular}
    \end{adjustbox}
    \vspace{-15pt}
\end{table}
To address these challenges, we propose \method, a memory-augmented LLM framework that treats memory as an elastic latent resource.
\method first constructs an offline latent memory bank by encoding each memory chunk into a retrieval key and a content cache.
At inference time, \method derives a query-conditioned retrieval state from the LoRA-adapted reasoner's hidden representation after selecting a retrieval-control token.
This enables retrieval to adapt to the current query and the model's internal reasoning state, rather than relying only on rigid similarity matching.
After retrieving candidate memories, \method uses a lightweight budget policy to assign each retrieved chunk a variable number of latent tokens.
Evidence-bearing memories can receive larger latent budgets, while redundant memories are compressed or suppressed.
These latent states are injected into the reasoner as soft memory tokens for generation.
Crucially, \method does not train retrieval, allocation, and generation as isolated modules.
Instead, it jointly optimizes the memory-use process with downstream task rewards through group-relative policy optimization \citep{shao2024deepseekmath,guo2025deepseek}.
The reward signal supervises the retrieval-control decision, the memory-budget allocations, the latent projector, and the LoRA-adapted \citep{hu2022lora} reasoner.
As a result, \method learns how much latent capacity to allocate to each retrieved memory based on downstream task feedback, aligning memory use with actual task utility.


We comprehensively evaluate \method on \textsc{MemorySuite}, which spans 4 memory-intensive QA benchmarks (PersonaMem-32K, PersonaMem-128K, LoCoMo, and LongMemEval) and the embodied ALFWorld environment \citep{jiang2025personamem,maharana2024evaluating,wu2024longmemeval,shridhar2020alfworld}.
Experimental results indicate that \method consistently outperforms both text-space and latent-space memory baselines, improving the weighted average QA accuracy by 26.2\% and 24.6\% and the ALFWorld success rate by 66.3\% and 27.2\% over the strongest baseline with Qwen2.5-3B-Instruct and Qwen2.5-7B-Instruct, respectively.
Furthermore, \method achieves the lowest weighted-average token consumption on ALFWorld. Controlled comparisons examine the effects of training objective, parameter capacity, and latent projection.
Ablation studies further validate the key designs in \method, and qualitative analyses show that it allocates larger budgets to memories carrying useful evidence or transferable plans, even when they are not the most similar ones.

\vspace{0mm}

\section{Preliminaries}
\label{sec:preliminaries}

\vspace{0mm}

\subsection{Memory-Augmented Language Models}

\vspace{0mm}

We consider a memory-augmented language model that answers a query $q$
with access to an external memory corpus
$\mathcal{M}=\{m_i\}_{i=1}^{N}$. Each memory item $m_i$ is a chunk of
long-term context, such as a dialogue segment, a document passage, or a
procedural skill card. A standard retrieval-augmented system first
embeds the query and memory chunks, retrieves the top-ranked chunks, and
then prepends the retrieved content to the model input:
\vspace{0mm}
\begin{equation}
    \mathcal{R}(q)
    =
    \operatorname{TopK}_{m_i \in \mathcal{M}}
    \operatorname{sim}(e_q, e_i),
    \label{eq:standard_retrieval}
\end{equation}
where $e_q$ and $e_i$ denote the query and memory embeddings,
respectively.

This retrieval-and-concatenation paradigm provides a simple mechanism
for long-term memory access, but it imposes a rigid interface between
memory and generation. The system typically retrieves a fixed number of
chunks, injects them as plain text, and allocates context-window capacity
in proportion to their original token lengths rather than their
query-specific utility. Consequently, irrelevant or redundant chunks may
consume substantial context, while compact but crucial evidence may
receive insufficient representational capacity.

\vspace{0mm}

\subsection{Latent Memory Interfaces}

\vspace{0mm}

An alternative is to represent memory using continuous latent vectors
rather than natural-language text. Let $E(\cdot)$ denote the token
embedding layer of an LLM with hidden dimension $d$. A latent or soft
memory token $z \in \mathbb{R}^{d}$ can be inserted directly into the
input embedding sequence:
\vspace{0mm}
\begin{equation}
    X
    =
    \bigl[
        E(x_1), \ldots, E(x_T),
        z_1, \ldots, z_B
    \bigr].
    \label{eq:soft_token_input}
\end{equation}
Such latent interfaces avoid re-rendering retrieved memories as text and
can reduce prompt overhead. However, existing latent-memory methods
often still expose a fixed-capacity interface: each retrieved chunk is
assigned the same number of latent tokens regardless of its relevance,
redundancy, or information density. This motivates a more flexible memory interface. Ideally, a model should
adapt memory use at two levels: it should select the memory chunks that
are useful for the current query, and it should allocate more latent
capacity to chunks that are more informative for solving the task.
\vspace{-2.5mm}

\section{\method: Memory Use as a Learnable Resource}
\label{sec:method}

\vspace{-2.5mm}

\subsection{Overview}
\label{sec:method_overview}

\vspace{-1.8mm}

\begin{figure*}[t]
    \centering
    \includegraphics[width=1\linewidth]{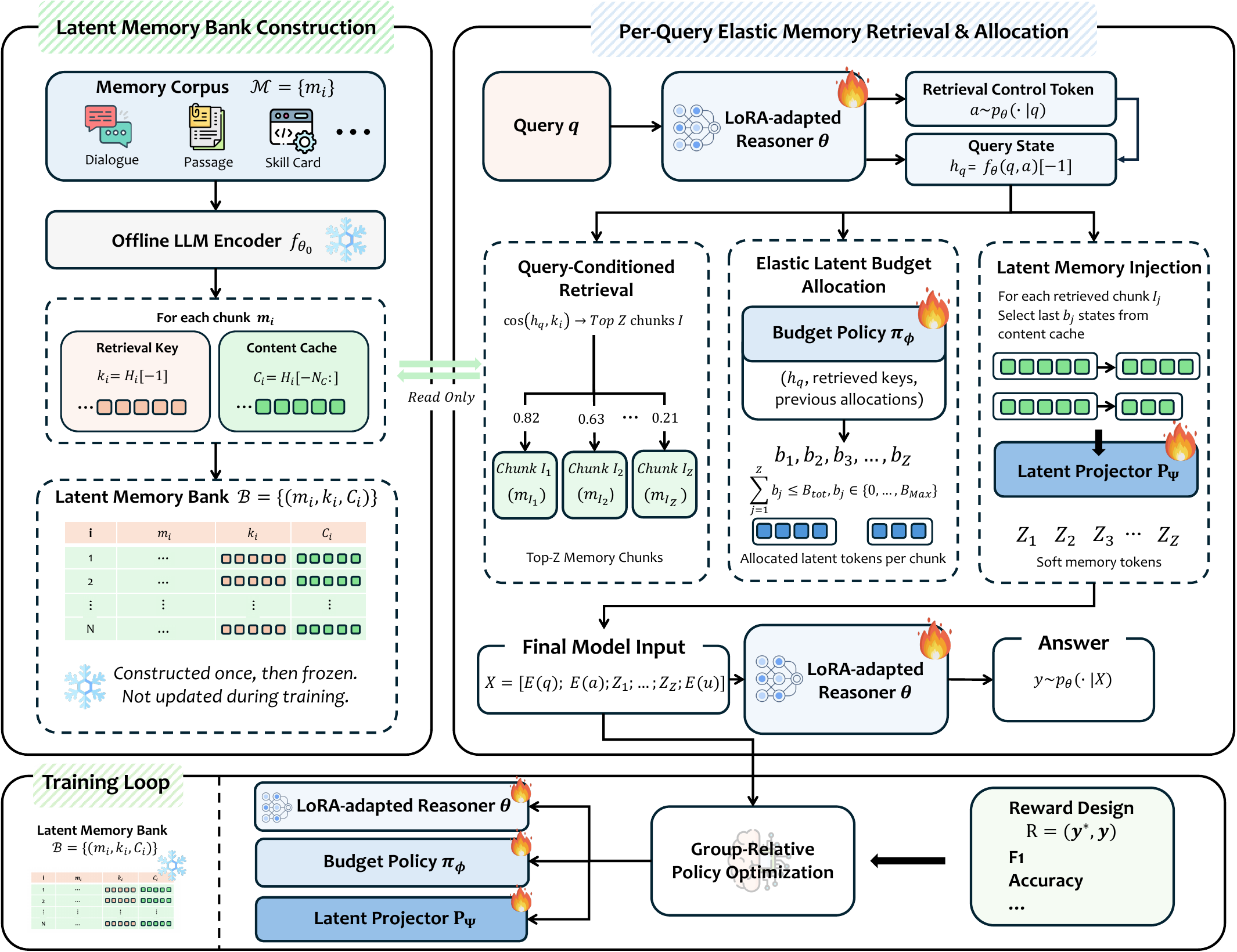}
\caption{
Overview of \method.
\textbf{(1) Latent memory bank construction}: a frozen offline encoder stores each memory chunk as a retrieval key and a latent content cache.
\textbf{(2) Elastic retrieval and allocation}: the LoRA-adapted reasoner derives a query state to retrieve top-ranked chunks, and a budget policy assigns each chunk a latent budget $b_j \in \{0,\ldots,B_{\max}\}$.
\textbf{(3) Latent memory injection}: the selected cached states are projected into soft memory tokens for generation.
\textbf{(4) Reward-based optimization}: GRPO trains the reasoner, budget policy, and projector with task rewards, while the memory bank stays frozen. Sampling denotes training; evaluation uses argmax decisions.
}
\vspace{-10pt}
\label{fig:framework}
\end{figure*}

Given a query $q$ and memory corpus $\mathcal M$, \method learns which memories to retrieve and how much latent capacity to assign to each.
Figure~\ref{fig:framework} summarizes the four stages: offline bank construction, query-conditioned retrieval, elastic allocation, and soft-token injection.
Downstream rewards train the LoRA-adapted reasoner, budget policy, and projector; the memory bank remains frozen.
Appendix~\ref{app:training} provides the training procedure.

\subsection{Latent Memory Bank Construction}
\label{sec:latent_memory_bank}

\vspace{-1.8mm}

We partition the memory corpus into $N$ chunks
$\mathcal{M}=\{m_i\}_{i=1}^{N}$.
Each chunk is encoded once offline by a frozen base LLM encoder $f_{\theta_0}$:
\vspace{-1.5mm}
\begin{equation}
    H_i
    =
    f_{\theta_0}(m_i)
    \in
    \mathbb{R}^{L_i \times d},
    \label{eq:chunk_encoding}
\end{equation}
where $L_i$ is the length of chunk $m_i$ and $d$ is the hidden dimension.
From the hidden states $H_i$, we extract two representations:
\vspace{-1.5mm}
\begin{align}
    k_i
    &=
    H_i[-1]
    \in
    \mathbb{R}^{d},
    \label{eq:retrieval_key}
    \\
    C_i
    &=
    H_i[-N_c:]
    \in
    \mathbb{R}^{N_c \times d}.
    \label{eq:content_cache}
\end{align}
The final hidden state $k_i$ serves as a compact retrieval key, while $C_i$ stores the last $N_c$ hidden states as a latent content cache.
The resulting memory bank is
\vspace{-1.5mm}
\begin{equation}
    \mathcal{B}
    =
    \{(m_i,k_i,C_i)\}_{i=1}^{N}.
    \label{eq:memory_bank}
\end{equation}
Since $\mathcal{B}$ is constructed offline and kept fixed, query-time memory access does not require re-encoding the full memory corpus.

\vspace{-2.5mm}

\subsection{Query-Conditioned Retrieval}
\label{sec:query_conditioned_retrieval}

\vspace{-1.8mm}

Standard retrieval systems usually encode the query with a separate retriever.
This decouples memory access from the reasoner's internal state.
In contrast, \method derives the retrieval query from the reasoner itself, so retrieval is conditioned on how the model interprets the current input.

Given a query $q$, the reasoner first samples a retrieval-control token:
\vspace{-1mm}
\begin{equation}
    a
    \sim
    p_{\theta}(\cdot \mid q),
    \label{eq:retrieval_token}
\end{equation}
where $\theta$ denotes the trainable LoRA-adapted reasoner parameters. The action $a$ is one ordinary token from the full vocabulary, initialized from the pretrained next-token distribution with zero-initialized LoRA updates. Its hidden state, rather than its surface meaning, controls retrieval.
The query and retrieval-control token are then passed through the reasoner, and the last hidden state is used as the query representation:
\vspace{-1.5mm}
\begin{equation}
    h_q
    =
    f_{\theta}(q,a)[-1]
    \in
    \mathbb{R}^{d}.
    \label{eq:query_hidden}
\end{equation}
We score each memory chunk by cosine similarity between $h_q$ and the cached retrieval key:
\vspace{-1.5mm}
\begin{equation}
    s_i
    =
    \cos(h_q,k_i)
    =
    \frac{h_q^{\top}k_i}
    {\|h_q\|_2\|k_i\|_2}.
    \label{eq:retrieval_score}
\end{equation}
The top-$K$ chunks are selected as
\vspace{-0.5mm}
\begin{equation}
    I
    =
    \operatorname{TopK}_{i \in [N]} s_i,
    \label{eq:topz_retrieval}
\end{equation}
where $I=(I_1,\ldots,I_Z)$ denotes the retrieved chunk indices in descending retrieval-score order.
In the ALFWorld evaluation, $q$ is the fixed episode objective. With greedy control-token and budget decisions and a fixed memory bank, retrieval and allocation are task-conditioned and remain unchanged within an episode; the generator additionally conditions on the changing observation and interaction history through $u$.

\vspace{-2.5mm}

\subsection{Elastic Latent Budget Allocation}
\label{sec:elastic_budget}

\vspace{-1.8mm}

After retrieval, \method allocates a variable latent budget to each retrieved chunk.
Let $b_j$ denote the number of soft memory tokens assigned to the $j$-th retrieved chunk $m_{I_j}$.
We impose a per-chunk maximum budget:
\begin{equation}
    b_j \in \{0,1,\ldots,B_{\max}\}, \qquad B_{\max}\leq N_c.
    \label{eq:budget_constraint}
\end{equation}
The option $b_j=0$ allows the model to suppress retrieved chunks that are similar to the query but not useful for the downstream task.
Thus, the budget policy is not merely a compression module; it decides which retrieved memories deserve more representational capacity and which should contribute little or no latent signal.

\paragraph{Budget policy network.}
A lightweight Transformer jointly processes features of the $Z$ retrieved memories: the query state, memory key, their element-wise product, cosine score, and rank embedding.
For each memory, its output parameterizes a categorical distribution over $\{0,\ldots,B_{\max}\}$, from which $b_j$ is sampled.
Joint processing allows the allocator to compare retrieved memories before assigning capacity.
Appendix~\ref{app:budget_architecture} gives the feature construction and policy equations.

Unlike similarity-based reweighting, $\pi_\phi$ is not trained to imitate cosine scores.
It is optimized only through downstream rewards, allowing it to assign larger budgets to memories that improve the final answer or action even when they are not the most similar retrieved chunks.

\vspace{-2.5mm}

\subsection{Latent Memory Injection}
\label{sec:latent_injection}

\vspace{-1.8mm}

Given the allocated budget $b_j$ for retrieved chunk $m_{I_j}$, \method selects the last $b_j$ hidden states from its content cache:
\begin{equation}
    \widetilde{C}_j
    =
    \begin{cases}
    C_{I_j}[-b_j:], & b_j>0,\\
    \emptyset_{0\times d}, & b_j=0.
    \end{cases}
    \label{eq:selected_cache}
\end{equation}
If $b_j=0$, the chunk contributes no latent memory tokens.
Otherwise, the selected states are mapped into the reasoner's input embedding space through a learned projector $P_{\psi}$:
\begin{equation}
    Z_j
    =
    P_{\psi}(\widetilde{C}_j)
    \in
    \mathbb{R}^{b_j \times d}.
    \label{eq:projected_soft_tokens}
\end{equation}
The final input sequence is
\vspace{-1mm}
\begin{equation}
    X
    =
    \bigl[
        E(q);
        E(a);
        Z_1;
        \cdots;
        Z_Z;
        E(u)
    \bigr],
    \label{eq:final_input}
\end{equation}
where $u$ is the generation suffix: answer choices and output constraints for QA, or recent history, the current observation, admissible actions, and output constraints for ALFWorld. Soft tokens are inserted within the user-side input while preserving the chat-template boundaries.
The reasoner generates
\vspace{-1mm}
\begin{equation}
    y
    \sim
    p_{\theta}(\cdot \mid X).
    \label{eq:generation}
\end{equation}
Soft memory positions produce no output tokens and receive no token-level loss.
They influence generation through the reasoner's hidden states and are learned through downstream reward signals.

\vspace{-2.5mm}

\subsection{Reward-Based Optimization}
\label{sec:reward_optimization}

\vspace{-1.8mm}

The usefulness of retrieval and budget decisions is revealed only after the model produces an answer or completes an action.
We therefore optimize the full memory-use process with group-relative policy optimization.
For each query-target pair $(q,y^\star)$, \method samples $G$ trajectories:
\vspace{-0.5mm}
\begin{equation}
    \tau_g
    =
    (a_g,I_g,\mathbf{b}_g,y_g),
    \qquad
    g=1,\ldots,G,
    \label{eq:trajectory}
\end{equation}
where $a_g$ is the retrieval-control token, $I_g=(I_{g,1},\ldots,I_{g,Z})$ is the retrieved memory index sequence, $\mathbf{b}_g=(b_{g,1},\ldots,b_{g,Z})$ is the sampled budget allocation, and $y_g$ is the generated output.
Each trajectory receives a task reward:
\vspace{-0.5mm}
\begin{equation}
    r_g
    =
    R(y_g,y^{\star}),
    \label{eq:reward}
\end{equation}
where $R$ is exact-match accuracy for QA. For ALFWorld training, it is binary agreement between the snapped generated action and the expert action; environment success is the evaluation metric.

We compute a group-relative advantage by normalizing rewards within the sampled group:
\vspace{-1mm}
\begin{equation}
    A_g
    =
    \frac{
        r_g
        -
        \operatorname{mean}
        \left(
            \{r_{g'}\}_{g'=1}^{G}
        \right)
    }{
        \max
        \left(
            \operatorname{std}
            \left(
                \{r_{g'}\}_{g'=1}^{G}
            \right),
            \sigma_{\min}
        \right)
    },
    \label{eq:group_advantage}
\end{equation}
where $\sigma_{\min}$ is a small constant for numerical stability.

We minimize the clipped GRPO objective:
\vspace{-1mm}
\begin{equation}
    \mathcal{L}_{\mathrm{GRPO}}
    =
    -\frac{1}{G}\sum_{g=1}^{G}
    \left[
        \min
        \left(
            \rho_g A_g,\;
            \operatorname{clip}
            \left(
                \rho_g,\,
                1-\epsilon,\,
                1+\epsilon
            \right)
            A_g
        \right)
    \right],
    \label{eq:grpo_loss}
\end{equation}
where the probability ratio is
\vspace{-1mm}
\begin{equation}
    \rho_g
    =
    \frac{
        p_{\theta,\phi,\psi}(\tau_g)
    }{
        p_{\theta_{\mathrm{old}},\phi_{\mathrm{old}},\psi_{\mathrm{old}}}(\tau_g)
    }.
    \label{eq:policy_ratio}
\end{equation}
Including the deterministic retrieval decision, the trajectory likelihood factorizes as
\vspace{-1.5mm}
\begin{equation}
    p_{\theta,\phi,\psi}(\tau_g)
    =
    p_{\theta}(a_g \mid q)\,\mathbf{1}\{I_g=T_\theta(q,a_g)\}
    \left[
        \prod_{j=1}^{Z}
        \pi_{\phi}
        \!\left(
            b_{g,j}
            \mid
            h_{q,g},\, I_g
        \right)
    \right]
    p_{\theta}\!\left(y_g \mid X_g(\psi)\right),
    \label{eq:trajectory_likelihood}
\end{equation}
where $T_\theta(q,a)=\operatorname{TopZ}_i\cos(f_\theta(q,a)[-1],k_i)$, $h_{q,g}=f_{\theta}(q,a_g)[-1]$, and
$X_g(\psi)=\bigl[E(q);\,E(a_g);\,P_{\psi}(\widetilde C_{g,1});\cdots;\,P_{\psi}(\widetilde C_{g,Z});\,E(u)\bigr]$
is the final input sequence whose soft memory tokens are produced by the latent
projector~$P_{\psi}$, with $\widetilde C_{g,j}$ selected from cache $C_{I_{g,j}}$ using budget $b_{g,j}$ as in Eq.~\ref{eq:selected_cache}.
The indicator accounts for deterministic retrieval in the joint likelihood; a likelihood conditional on fixed rollout indices is a different surrogate. Hard top-$Z$ selection is non-differentiable. GRPO reinforces sampled control tokens whose resulting states retrieve useful memories; it does not differentiate through the selected indices. The budget and generation likelihoods train allocation and latent injection. Appendix~\ref{app:control_analysis} provides a same-checkpoint intervention.
The trainable components are the LoRA-adapted reasoner $\theta$, the budget
policy $\pi_{\phi}$, and the latent projector $P_{\psi}$.
The offline memory bank $\mathcal{B}$ remains fixed during training. The detailed training procedure is provided in Appendix~\ref{app:training}.
\vspace{0mm}

\section{Experiments} \label{sec:exp}

\vspace{0mm}

To evaluate the effectiveness of \method in memory-augmented language modeling, we conduct experiments on \textsc{MemorySuite}, which contains two complementary settings: memory-intensive question answering and embodied agentic decision-making
(see Appendix~\ref{appendix:dataset_descriptions}).

\xhdr{Tasks and Metrics}
\textsc{MemorySuite-QA} includes PersonaMem-32K~\citep{jiang2025personamem}, PersonaMem-128K~\citep{jiang2025personamem}, LoCoMo~\citep{maharana2024evaluating}, and LongMemEval~\citep{wu2024longmemeval}.
We report accuracy as the primary metric.
\textsc{MemorySuite-Agentic} uses ALFWorld~\citep{shridhar2020alfworld} and evaluates both seen and unseen splits, using success rate (SR) as the primary metric.
We also report token consumption (\#Tok.); this metric does not measure wall-clock latency or FLOPs.

\xhdr{Baselines and Settings}
We compare \method with representative memory-based baselines from two paradigms.
\textbf{Text-space memory baselines} include MemoryBank~\citep{zhong2024memorybank}, A-MEM~\citep{xu2025mem}, LightMem~\citep{fang2025lightmem}, Mem0~\citep{chhikara2025mem0}, MemoryOS~\citep{kang2025memory}, MemP~\citep{fang2025memp}, and LangMem~\citep{langchain2024langmem}.
\textbf{Latent-space memory baselines} include MemGen~\citep{zhang2025memgen}, AutoCompressor~\citep{chevalier2023adapting}, and M+~\citep{wang2025m+}.
All methods are evaluated with Qwen2.5-3B-Instruct\footnote{\url{https://huggingface.co/Qwen/Qwen2.5-3B-Instruct}} and Qwen2.5-7B-Instruct\footnote{\url{https://huggingface.co/Qwen/Qwen2.5-7B-Instruct}} backbones under a no-few-shot-example setting.

\xhdr{Comparison Protocol}
All latent-space baselines are retrained on each task's training split with the same task supervision, backbone scale, source information, and evaluation protocol, while retaining native objectives and trainable modules.
QA banks contain only each example's raw history; held-out answers and attempts never enter memory.
ALFWorld latent methods share Gemini-generated training skill cards; text methods receive the same segmented raw trajectories and construct memories natively.
All methods share task instructions and chat templates; their memory interfaces differ (Appendix~\ref{app:prompts}).
Table~\ref{tab:controlled_comparison} additionally examines the effects of training objective and increased LoRA capacity; Appendix~\ref{app:comparison_protocol} details the comparison scope.

\begin{table*}[t]
\centering
\small
\setlength{\tabcolsep}{3pt}
\renewcommand{\arraystretch}{1.08}
\caption{\textbf{Main comparison results on PersonaMem-32K, PersonaMem-128K, LoCoMo, LongMemEval, and ALFWorld.}
For Acc. and SR, higher is better; for \#Tok., lower is better. \#Tok. counts task-text input, injected memory, and generated output, distinct from pure latent-memory budget (Appendix~\ref{app:hyper}).
\textbf{Bold} and \underline{underline} denote the best and second-best results.
Task-performance scores are means over five independent training seeds. Avg. columns are weighted by test-set size. Appendix~\ref{app:seed_statistics} summarizes the \method--MemGen means.}
\label{tab:main_results}
\vspace{-2.6mm}
\resizebox{\textwidth}{!}{
\begin{tabular}{>{\centering\arraybackslash}m{1.0cm}>{\raggedright\arraybackslash}m{2.45cm}cc cc cc cc cc cc cc cc}
\toprule
\multirow{3}{*}{\textbf{Model}} &
\multirow{3}{*}{\textbf{Methods}} &
\multicolumn{10}{c}{\textbf{Memory QA Tasks}} &
\multicolumn{6}{c}{\textbf{Agentic Embodied Interactive Tasks}} \\
\cmidrule(lr){3-12} \cmidrule(lr){13-18}
& &
\multicolumn{2}{c}{\textbf{PersonaMem-32K}} &
\multicolumn{2}{c}{\textbf{PersonaMem-128K}} &
\multicolumn{2}{c}{\textbf{LoCoMo}} &
\multicolumn{2}{c}{\textbf{LongMemEval}} &
\multicolumn{2}{c}{\textbf{Avg.}} &
\multicolumn{2}{c}{\textbf{ALF-Seen}} &
\multicolumn{2}{c}{\textbf{ALF-Unseen}} &
\multicolumn{2}{c}{\textbf{Avg.}} \\
\cmidrule(lr){3-4} \cmidrule(lr){5-6} \cmidrule(lr){7-8} \cmidrule(lr){9-10} \cmidrule(lr){11-12}
\cmidrule(lr){13-14} \cmidrule(lr){15-16} \cmidrule(lr){17-18}
& &
\textbf{Acc.} & \textbf{\#Tok.} &
\textbf{Acc.} & \textbf{\#Tok.} &
\textbf{Acc.} & \textbf{\#Tok.} &
\textbf{Acc.} & \textbf{\#Tok.} &
\textbf{Acc.} & \textbf{\#Tok.} &
\textbf{SR} & \textbf{\#Tok.} &
\textbf{SR} & \textbf{\#Tok.} &
\textbf{SR} & \textbf{\#Tok.} \\
\midrule

\multirow{13}{*}{\rotatebox[origin=c]{90}{\shortstack{\textbf{Qwen2.5-3B}\\\textbf{-Instruct}}}} &
\multicolumn{11}{l}{\cellcolor{gray!15}\textit{Memory-based Methods in Text-Space}} &
\multicolumn{6}{l}{\cellcolor{gray!15}} \\
& MemoryBank    & 0.50 & 2,473 & 0.33 & 4,618 & 0.41 & 2,202 & 0.29 & 9,591 & 0.37 & 4,089 & 0.06 & 76,016 & 0.09 & 71,173 & 0.08 & 73,648 \\
& A-MEM         & 0.48 & 4,899 & 0.38 & 3,530 & 0.39 & 3,348 & 0.45 & 15,625 & 0.40 & 5,282 & 0.04 & 17,239 & 0.05 & 14,248 & 0.04 & 15,776 \\
& LightMem      & 0.46 & 851 & 0.35 & 1,306 & 0.40 & 3,468 & 0.44 & 1,248 & 0.39 & 2,239 & 0.14 & 13,252 & 0.17 & \underline{12,617} & 0.15 & 12,941 \\
& Mem0          & 0.42 & 778 & 0.39 & 662 & 0.33 & 836 & 0.33 & 2,843 & 0.36 & 1,062 & 0.14 & 43,055 & 0.12 & 43,677 & 0.13 & 43,359 \\
& MemoryOS      & 0.48 & 4,419 & 0.32 & 9,440 & 0.33 & 4,420 & 0.40 & 13,854 & 0.35 & 7,452 & 0.11 & 68,131 & 0.13 & 65,845 & 0.12 & 67,013 \\
& MemP          & 0.42 & 12,002 & 0.38 & 29,567 & 0.26 & 5,140 & 0.41 & 26,571 & 0.33 & 16,873 & 0.11 & 101,580 & 0.10 & 110,019 & 0.11 & 105,707 \\
& LangMem       & 0.40 & 675 & 0.30 & 1,626 & 0.40 & 936 & 0.28 & 687 & 0.35 & 1,112 & 0.20 & \underline{12,276} & 0.11 & \textbf{12,449} & 0.15 & \underline{12,361} \\

&
\multicolumn{11}{l}{\cellcolor{gray!15}\textit{Memory-based Methods in Latent-Space}} &
\multicolumn{6}{l}{\cellcolor{gray!15}} \\
& MemGen         & \underline{0.62} & \underline{502} & \underline{0.58} & \underline{393} & \underline{0.60} & \underline{198} & \underline{0.55} & \underline{242} & \underline{0.59} & \underline{291} & \underline{0.29} & 14,708 & \underline{0.25} & 14,416 & \underline{0.27} & 14,565 \\
& AutoCompressor & 0.48 & 717 & 0.45 & 1,127 & 0.47 & 234 & 0.44 & 998 & 0.46 & 677 & 0.16 & 21,145 & 0.13 & 21,571 & 0.14 & 21,353 \\
& M+             & 0.56 & \textbf{447} & 0.52 & \textbf{350} & 0.54 & \textbf{154} & 0.50 & \textbf{198} & 0.53 & \textbf{247} & 0.20 & 14,246 & 0.16 & 14,233 & 0.18 & 14,240 \\
& \cellcolor{hl-llama}\textbf{\method} &
\cellcolor{hl-llama}\textbf{0.74} & \cellcolor{hl-llama}537 &
\cellcolor{hl-llama}\textbf{0.77} & \cellcolor{hl-llama}682 &
\cellcolor{hl-llama}\textbf{0.74} & \cellcolor{hl-llama}577 &
\cellcolor{hl-llama}\textbf{0.68} & \cellcolor{hl-llama}433 &
\cellcolor{hl-llama}\textbf{0.74} & \cellcolor{hl-llama}589 &
\cellcolor{hl-llama}\textbf{0.49} & \cellcolor{hl-llama}\textbf{11,028} &
\cellcolor{hl-llama}\textbf{0.41} & \cellcolor{hl-llama}12,846 &
\cellcolor{hl-llama}\textbf{0.45} & \cellcolor{hl-llama}\textbf{11,917} \\
\midrule

\multirow{13}{*}{\rotatebox[origin=c]{90}{\shortstack{\textbf{Qwen2.5-7B}\\\textbf{-Instruct}}}} &
\multicolumn{11}{l}{\cellcolor{gray!15}\textit{Memory-based Methods in Text-Space}} &
\multicolumn{6}{l}{\cellcolor{gray!15}} \\
& MemoryBank    & 0.53 & 2,639 & 0.35 & 4,724 & 0.40 & 2,301 & 0.39 & 10,822 & 0.39 & 4,358 & 0.11 & 36,602 & 0.19 & 39,385 & 0.15 & 37,963 \\
& A-MEM         & 0.52 & 9,731 & 0.37 & 11,098 & 0.42 & 3,289 & 0.50 & 31,489 & 0.42 & 10,408 & 0.06 & 29,543 & 0.09 & 29,531 & 0.07 & 29,537 \\
& LightMem      & 0.52 & 864 & 0.33 & 1,228 & 0.41 & 3,702 & 0.53 & 939 & 0.41 & 2,275 & 0.33 & 11,150 & 0.37 & 10,617 & 0.35 & 10,889 \\
& Mem0          & 0.54 & 794 & 0.39 & 676 & 0.38 & 921 & 0.48 & 3,202 & 0.41 & 1,157 & \underline{0.43} & 26,165 & \underline{0.40} & 27,327 & \underline{0.42} & 26,733 \\
& MemoryOS      & 0.52 & 4,695 & 0.38 & 9,748 & 0.42 & 4,998 & 0.53 & 14,489 & 0.43 & 7,926 & 0.38 & 60,129 & \underline{0.40} & 50,238 & 0.39 & 55,292 \\
& MemP          & 0.52 & 11,964 & 0.40 & 29,581 & 0.22 & 5,382 & 0.48 & 26,580 & 0.34 & 16,985 & 0.24 & 143,768 & 0.22 & 115,492 & 0.23 & 129,940 \\
& LangMem       & 0.68 & 1,359 & 0.42 & 3,097 & 0.35 & 1,478 & 0.42 & 781 & 0.41 & 1,911 & 0.40 & \underline{9,735} & 0.28 & \underline{10,140} & 0.34 & \underline{9,933} \\

&
\multicolumn{11}{l}{\cellcolor{gray!15}\textit{Memory-based Methods in Latent-Space}} &
\multicolumn{6}{l}{\cellcolor{gray!15}} \\
& MemGen         & \underline{0.70} & \underline{502} & \underline{0.66} & \underline{393} & \underline{0.68} & \underline{198} & \underline{0.63} & \underline{242} & \underline{0.67} & \underline{291} & 0.39 & 12,933 & 0.34 & 13,551 & 0.36 & 13,235 \\
& AutoCompressor & 0.56 & 717 & 0.52 & 1,127 & 0.55 & 234 & 0.51 & 998 & 0.54 & 677 & 0.24 & 19,875 & 0.20 & 19,317 & 0.22 & 19,602 \\
& M+             & 0.64 & \textbf{447} & 0.59 & \textbf{350} & 0.62 & \textbf{154} & 0.57 & \textbf{198} & 0.60 & \textbf{247} & 0.32 & 12,520 & 0.28 & 13,382 & 0.30 & 12,942 \\
& \cellcolor{hl-qwen}\textbf{\method} &
\cellcolor{hl-qwen}\textbf{0.90} & \cellcolor{hl-qwen}483 &
\cellcolor{hl-qwen}\textbf{0.84} & \cellcolor{hl-qwen}557 &
\cellcolor{hl-qwen}\textbf{0.84} & \cellcolor{hl-qwen}362 &
\cellcolor{hl-qwen}\textbf{0.76} & \cellcolor{hl-qwen}410 &
\cellcolor{hl-qwen}\textbf{0.83} & \cellcolor{hl-qwen}443 &
\cellcolor{hl-qwen}\textbf{0.56} & \cellcolor{hl-qwen}\textbf{9,035} &
\cellcolor{hl-qwen}\textbf{0.49} & \cellcolor{hl-qwen}\textbf{9,031} &
\cellcolor{hl-qwen}\textbf{0.53} & \cellcolor{hl-qwen}\textbf{9,033} \\
\bottomrule
\end{tabular}
}
\end{table*}

\vspace{0mm}

\subsection{\method Outperforms Memory-based Methods in Text-Space and Latent-Space}
\label{sec:4.1}

\vspace{0mm}

The results in Table~\ref{tab:main_results} show that \method achieves the highest task-performance scores across the evaluated datasets and backbone scales.
Weighted QA accuracy increases from 0.588 to 0.742 with Qwen2.5-3B and from 0.668 to 0.832 with Qwen2.5-7B; weighted ALFWorld SR increases from 0.270 to 0.449 and from 0.416 to 0.529, respectively, relative to the strongest baseline in each aggregate setting. The reported means favor \method over MemGen on all six splits with both backbones.

\xhdr{Performance and Token Consumption}
On QA, \method uses fewer tokens than the text-space baselines, but more than MemGen and M+: its weighted average is 443 tokens, compared with 291 and 247, respectively.
The QA advantage is therefore higher accuracy at a larger token budget than these two latent baselines.
On ALFWorld, \method combines the highest weighted SR with the lowest weighted token consumption on both backbones.
Online latency, including native retrieval and model computation, is lowest for \method among the evaluated latent methods (Appendix~\ref{app:efficiency}). ALFWorld per-episode latency also depends on trajectory length and does not isolate per-step cost.

\subsection{Controlled Comparisons of Training and Model Capacity}
\label{sec:controlled_comparison}

\begin{table}[t]
\centering
\small
\setlength{\tabcolsep}{5pt}
\caption{\textbf{Controlled comparisons with Qwen2.5-7B-Instruct.}
Variants share data splits, source memory information, retrieval count, training examples, and update count.
ALFWorld SR is weighted over Seen and Unseen episodes.
The higher-rank text variant tests additional LoRA capacity; exact parameter matching is not established.}
\label{tab:controlled_comparison}
\begin{tabularx}{\linewidth}{Xlcc}
\toprule
Variant & Objective & LoCoMo Acc. & ALFWorld SR \\
\midrule
LoRA + text memory & SFT & 0.599 & 0.380 \\
LoRA + text memory & GRPO & 0.650 & 0.420 \\
Higher-rank LoRA + text memory & GRPO & 0.682 & 0.449 \\
\method with identity projector & GRPO & 0.739 & 0.460 \\
Full \method & GRPO & \textbf{0.841} & \textbf{0.529} \\
\bottomrule
\end{tabularx}
\end{table}

Within the text-memory configuration, replacing SFT with GRPO improves LoCoMo accuracy from 0.599 to 0.650 and ALFWorld SR from 0.380 to 0.420.
Increasing the text variant's LoRA capacity raises these scores to 0.682 and 0.449, still below 0.841 and 0.529 for full \method.
These comparisons suggest that the advantage is not explained solely by applying GRPO or increasing trainable capacity within the tested text-memory configuration; they do not isolate elastic allocation from all other interface differences.
Within \method, replacing the learned projector with direct latent-state injection reduces the scores to 0.739 and 0.460, supporting the value of a learned mapping between cached states and input embeddings.

\captionsetup[subfigure]{justification=raggedright}
\begin{figure*}[t]
    \vspace{0mm}
    \centering
    \begin{subfigure}{0.32\textwidth}
        \centering
        \includegraphics[width=\linewidth]{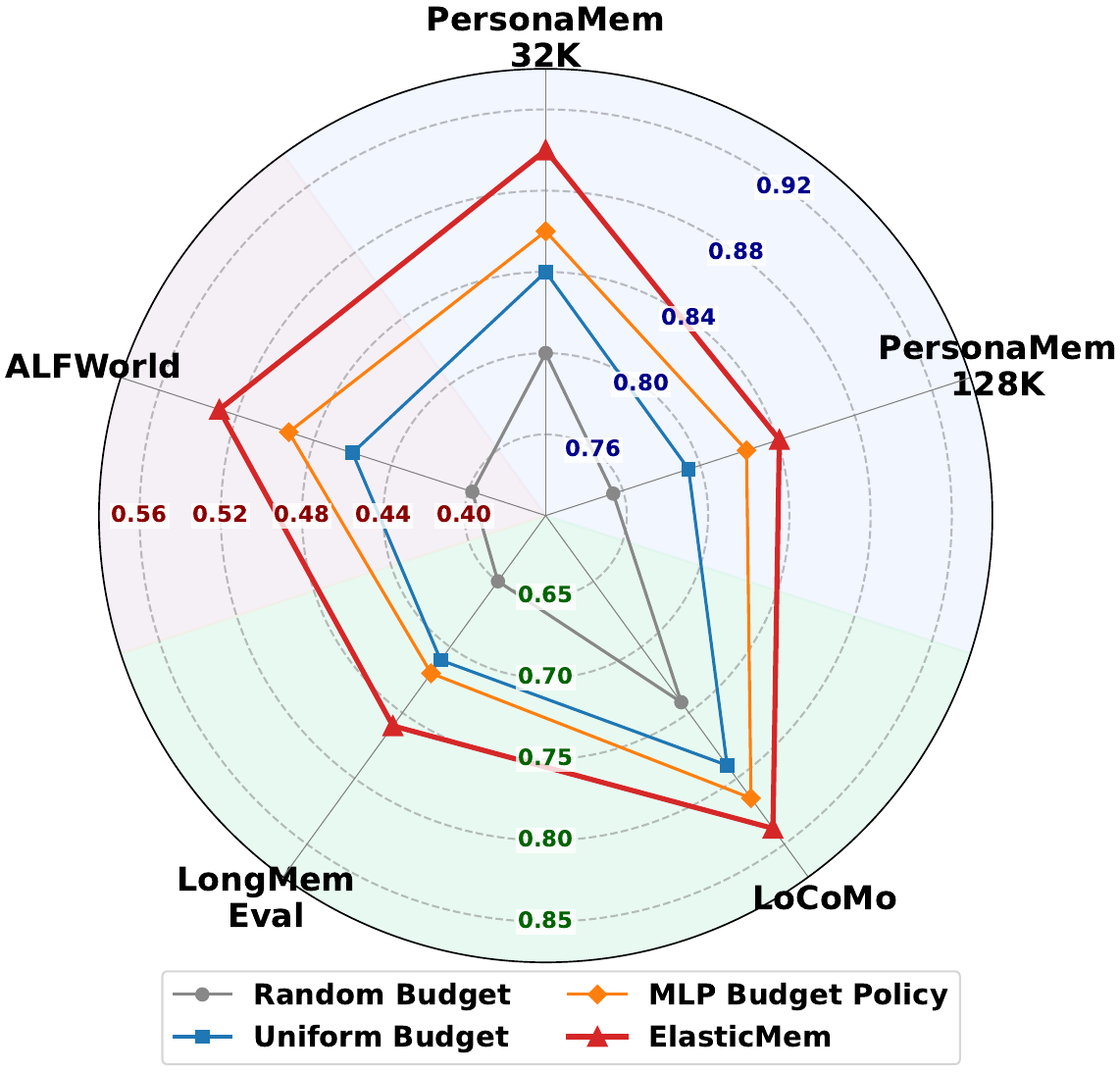}
        \vspace{-5mm}
        \caption*{\hspace{0.1cm}(a)}
        \label{fig:ablation_budget_policy}
    \end{subfigure}
    \begin{subfigure}{0.32\textwidth}
        \centering
        \includegraphics[width=\linewidth]{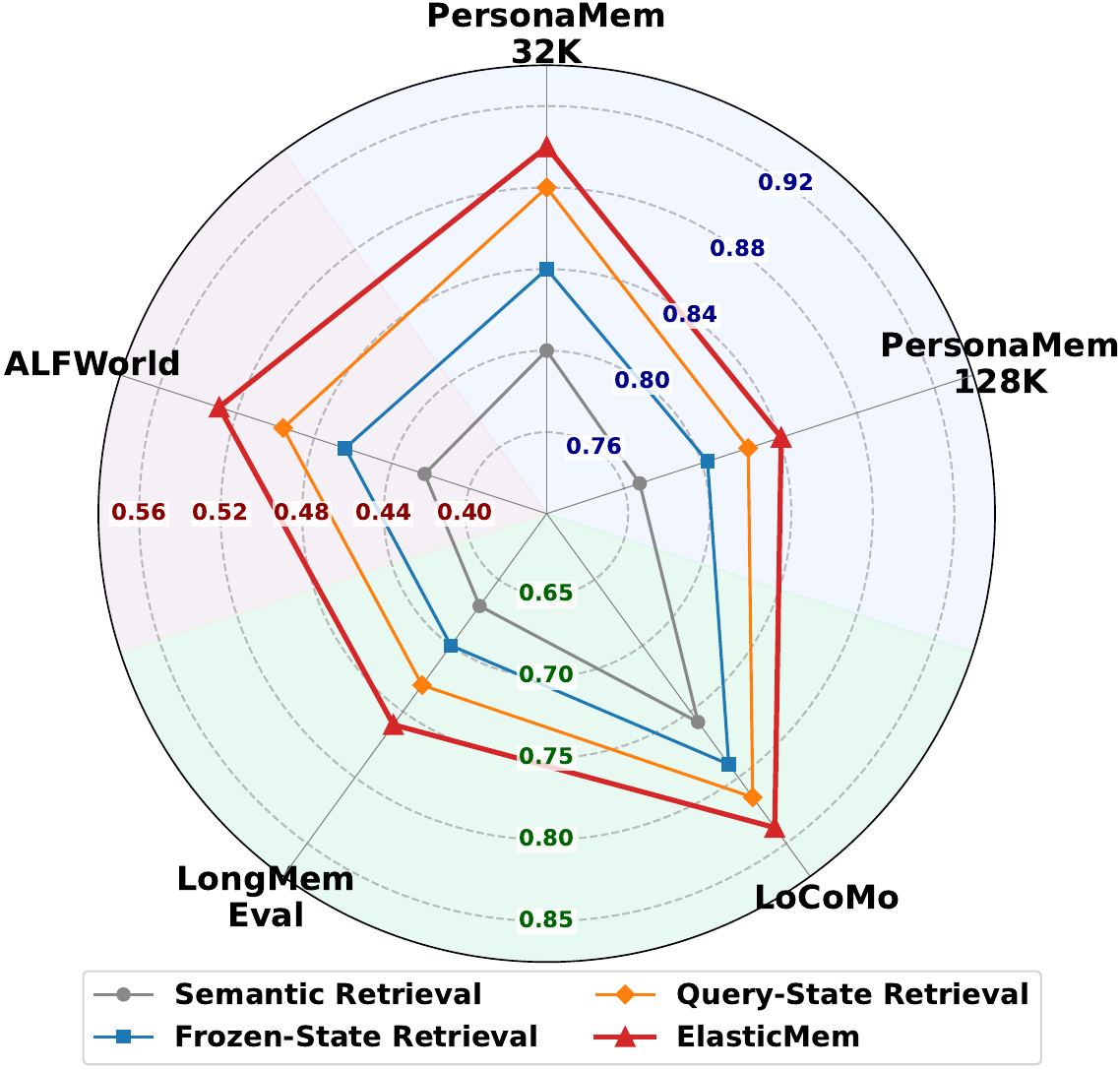}
        \vspace{-5mm}
        \caption*{\hspace{0.2cm}(b)}
        \label{fig:ablation_retrieval}
    \end{subfigure}
    \begin{subfigure}{0.32\textwidth}
        \centering
        \includegraphics[width=\linewidth]{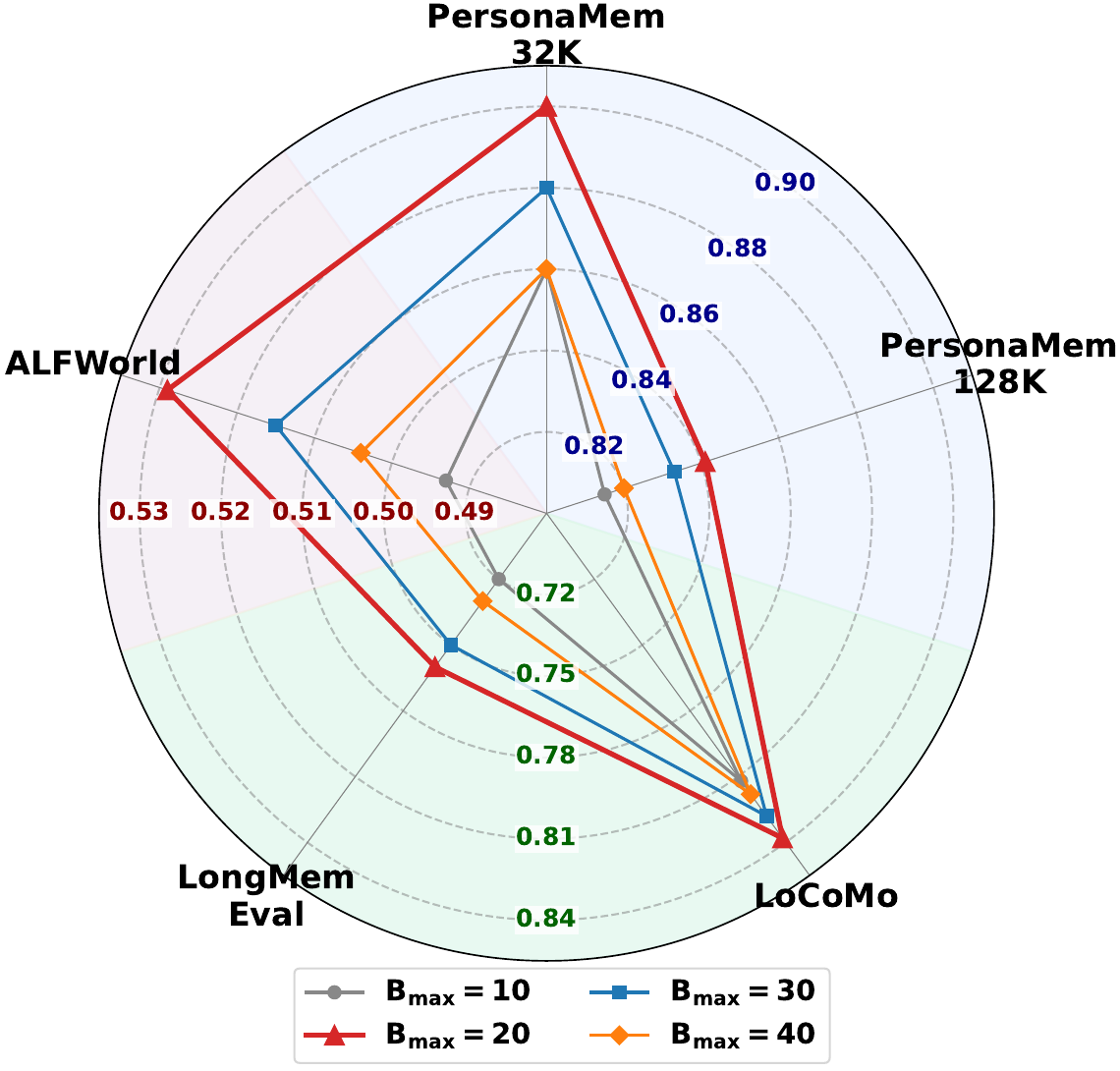}
        \vspace{-5mm}
        \caption*{\hspace{0.3cm}(c)}
        \label{fig:ablation_bmax}
    \end{subfigure}
    \vspace{0mm}
    \caption{\textbf{Ablation studies on Qwen2.5-7B-Instruct.}
(a) \textit{Effect of budget policy}: \method outperforms Random Budget, Uniform Budget, and MLP Budget Policy, showing that the Transformer budget policy better learns how to allocate latent capacity across retrieved memories.
(b) \textit{Effect of retrieval design}: \method outperforms Semantic Retrieval, Frozen-State Retrieval, and Query-State Retrieval, demonstrating the benefit of using a trainable reasoner-state retrieval mechanism with a sampled retrieval-control token.
(c) \textit{Maximum budget}: $N_c=B_{\max}$ in every setting; $B_{\max}=20$ performs best among the tested configurations.
}
    \label{fig:ablation_studies}
    \vspace{-4mm}
\end{figure*}


\subsection{Retrieval and Allocation Ablations}
\label{sec:ablation}


\xhdr{Budget Policy and Retrieval}
Figure~\ref{fig:ablation_studies}(a) compares random, uniform, MLP, and Transformer allocation policies.
The Transformer performs best across the five settings, supporting allocation that conditions jointly on the retrieved set.
Figure~\ref{fig:ablation_studies}(b) compares semantic retrieval, frozen reasoner states, trainable query-only states, and the control-token state.
The control-token variant performs best, supporting the use of a learned query-conditioned retrieval representation.

\xhdr{Maximum Latent Budget}
Among the tested per-chunk limits $B_{\max}\in\{10,20,30,40\}$, Figure~\ref{fig:ablation_studies}(c) favors $B_{\max}=20$.
This is an empirical result under the evaluated training settings: enlarging the allowable allocation space need not improve the policy learned with a finite training budget.
An inference-time clipping intervention is consistent with over-allocation contributing to the degradation (Appendix~\ref{app:budget_diagnostics}); sensitivity to $Z$ and $G$ is reported in Appendix~\ref{app:sensitivity}.
Qualitative examples of nonuniform memory allocation are discussed in Section~\ref{sec:case_study}.

\xhdr{Generalization Beyond Task-Specific Qwen Models}
With learned components frozen, checkpoints trained on PersonaMem-128K or ALFWorld outperform the corresponding MemGen checkpoints on LoCoMo and LongMemEval by 7.3--8.9 accuracy points; only the target memory bank is reconstructed.
Separate evaluations with Llama-3.2-3B and Llama-3.1-8B also favor \method on the two PersonaMem datasets.
The Llama experiments use one run per configuration and demonstrate applicability to another model family, not transfer of the same trained parameters across backbones (Appendix~\ref{app:generalization}).

\subsection{Case Studies of Memory Allocation}
\label{sec:case_study}

\xhdr{Evidence selection in PersonaMem-32K}
A question about a summer mentoring program asks how meditation helped with studying; the answer links meditation to improved focus and reduced anxiety.
\method assigns low budgets to highly ranked but only topically related memories about group projects, peer study, or a restatement of the query.
It assigns larger budgets to lower-ranked memories connecting yoga and meditation with stress reduction, and anxiety and sleep deprivation with cognition.
The resulting answer connects meditation, focus, anxiety, and study performance.
This allocation is consistent with prioritizing useful evidence over surface similarity.

\xhdr{Plan transfer in ALFWorld}
For \emph{``heat some apple and put it on the countertop''}, \method assigns a high budget to a card from a \texttt{cool} task whose plan places an object on a \texttt{shelf}, an open surface resembling the target countertop.
Cards involving \texttt{coffeemachine} or \texttt{stoveburner} destinations receive lower budgets.
Even cards sharing the objective \emph{``heat some mug and put it in coffeemachine''} receive different budgets when their plans search open surfaces versus cabinets and appliances.
These observations suggest sensitivity to procedural structure beyond task labels; they do not establish that any individual memory caused success.

\vspace{-3.5mm}

\section{Additional Related Work}

\noindent\textbf{LLM Agent Memory.}
Text-space memory systems store, organize, and inject natural-language records into the context, including MemoryBank~\citep{zhong2024memorybank}, Mem0~\citep{chhikara2025mem0}, MemoryOS~\citep{kang2025memory}, LangMem~\citep{langchain2024langmem}, ArcMemo~\citep{ho2025arcmemo}, and experience-oriented frameworks such as A-MEM~\citep{xu2025mem}, LightMem~\citep{fang2025lightmem}, and MemP~\citep{fang2025memp}.
Their textual memories consume context tokens.
Latent-space methods such as AutoCompressor~\citep{chevalier2023adapting}, MemGen~\citep{zhang2025memgen}, and M+~\citep{wang2025m+} provide latent alternatives to textual context.
In contrast, \method retrieves memories from the reasoner's state and learns how many latent tokens each memory deserves.

\noindent\textbf{Reward-Based LLM Optimization.}
PPO-style methods~\citep{ouyang2022training,liu2024deepseek, yu2024flow}, preference optimization~\citep{rafailov2023direct}, process supervision~\citep{lightman2023let, yu2026weak}, and group-relative objectives~\citep{shao2024deepseekmath,liu2025understanding,zheng2025group,cui2025entropy} improve model behavior through feedback. We apply task rewards to memory retrieval, allocation, and generation with a frozen external bank.

\vspace{-3.5mm}

\section{Conclusion}

\vspace{-3mm}

We introduce \method, a memory-augmented LLM framework that treats memory as an elastic latent resource, learning from downstream task rewards which memories to retrieve and how much latent capacity each deserves.
Extensive experiments on \textsc{MemorySuite} demonstrate that \method substantially outperforms both text-space and latent-space memory baselines on memory-intensive QA and embodied decision-making, while achieving the lowest weighted-average token consumption on ALFWorld.
\section*{AI use disclosure}
We used generative AI tools solely to polish the writing of this paper. The authors reviewed and edited all AI-assisted text and take full responsibility for the content.

\bibliography{iclr2027_conference}
\bibliographystyle{iclr2027_conference}

\clearpage
\DoToC
\appendix
\section{Comparison Protocol and Additional Experimental Results}
\label{app:comparison_protocol}

\subsection{Scope of the Comparisons}
The main comparison evaluates complete memory systems with the same backbone scale, task partitions, source memory information, and evaluation protocol.
MemGen, AutoCompressor, and M+ are retrained separately for each task while retaining their native objectives and memory mechanisms.
The shared source information consists of benchmark histories for QA and training expert trajectories for ALFWorld; each method's memory representation and construction procedure remain part of the evaluated system.
For ElasticMem and the retrained latent-space baselines, QA answer labels and ALFWorld expert demonstrations provide shared task supervision. Their native training objectives determine how this supervision is used; the comparison does not impose an identical loss, architecture, or total training computation on all methods.
Consequently, the main comparison establishes system-level performance under these settings and does not isolate the contribution of any single component.

\paragraph{Shared information and native training.}
For QA, all methods use the same benchmark histories and training QA partitions; held-out questions, answers, and attempts are excluded from memory construction.
For ALFWorld, source trajectories, trajectory-level segmentation, memory-bank size, construction backbone, and top-$Z$ count are controlled.
ElasticMem and the latent-space baselines receive the same Gemini-generated skill cards and boundaries. Text-space methods apply their native construction procedures to the same segmented raw trajectories.
MemGen retains its reasoner, memory weaver, and learned trigger with RL training; AutoCompressor retains its backbone-scale adaptation for summary representations; M+ retains its native learned memory/retrieval components.
ElasticMem trains one LoRA-adapted reasoner, the budget policy, and the latent projector while keeping the offline encoder fixed.
These are native architectural budgets, not a claim that the baseline objectives or total training FLOPs are identical.

\begin{table}[htbp]
\centering
\small
\caption{\textbf{Scope of the experimental controls.} The controlled variants complement the complete-system comparison. Equal update counts do not imply equal training FLOPs.}
\label{tab:comparison_scope}
\begin{tabularx}{\linewidth}{>{\raggedright\arraybackslash}p{0.23\linewidth}>{\raggedright\arraybackslash}X>{\raggedright\arraybackslash}X}
\toprule
Comparison & Shared conditions & Interpretation \\
\midrule
Main table & Backbone scale, task partitions, source information, evaluation protocol; shared task supervision for latent-space methods & Complete systems with native memory mechanisms and objectives; not a matched-compute component ablation. \\
Text-memory SFT vs. GRPO & Architecture, retrieved text evidence, training examples, update count & Effect of the training objective within the text-memory configuration. \\
Higher-rank text vs. full \method & Backbone, data, retrieval count, GRPO objective, update count & Increased text-memory capacity across different memory interfaces; does not isolate allocation alone. \\
Identity vs. learned projector & Retrieval and allocation pipeline, LoRA adaptation, GRPO objective, update count & Effect of replacing the learned projector with direct latent-state injection. \\
\bottomrule
\end{tabularx}
\end{table}

The higher-rank text-memory variant increases LoRA capacity. Without the exact rank and a verified parameter inventory, this comparison is not treated as a strict parameter-matched control.
For the identity-projector variant, selected states are injected directly as $Z_j=\widetilde C_j$; retrieval, allocation, and reasoner adaptation are retained.
Table~\ref{tab:controlled_comparison} reports these comparisons with Qwen2.5-7B-Instruct on LoCoMo and ALFWorld.

\paragraph{Aggregation weights.}
The QA average uses weights proportional to the test counts $(50,233,314,100)$ for PersonaMem-32K, PersonaMem-128K, LoCoMo, and LongMemEval, respectively, totaling 697 examples.
The corresponding percentages are approximately $(7.17,33.43,45.05,14.35)\%$.
ALFWorld averages use 140 Seen and 134 Unseen episodes, with weights $140/274$ and $134/274$ (approximately $51.09\%$ and $48.91\%$).

\subsection{Five-Seed Comparison with MemGen}
\label{app:seed_statistics}
The main-table experiments are repeated with five independent training seeds. Table~\ref{tab:seed_intervals} gives the detailed \method--MemGen comparison with fixed data splits, memory banks, hyperparameters, and evaluation protocol.
The table reports run means without uncertainty estimates or significance tests. Run-level statistical analysis remains a limitation; differences in means alone do not establish significance.
In particular, MemGen is not the strongest 7B ALFWorld baseline in Table~\ref{tab:main_results}.

\begin{table}[htbp]
\centering
\small
\setlength{\tabcolsep}{4pt}
\caption{\textbf{Five-seed performance comparison.} Reported means across five training seeds; uncertainty estimates are not shown. QA uses accuracy; ALFWorld uses success rate.}
\label{tab:seed_intervals}
\begin{tabular}{lcccc}
\toprule
& \multicolumn{2}{c}{Qwen2.5-3B} & \multicolumn{2}{c}{Qwen2.5-7B} \\
\cmidrule(lr){2-3}\cmidrule(lr){4-5}
Dataset & MemGen & \method & MemGen & \method \\
\midrule
PersonaMem-32K & $0.62$ & $0.74$ & $0.70$ & $0.90$ \\
PersonaMem-128K & $0.58$ & $0.77$ & $0.66$ & $0.84$ \\
LoCoMo & $0.60$ & $0.74$ & $0.68$ & $0.84$ \\
LongMemEval & $0.55$ & $0.68$ & $0.63$ & $0.76$ \\
ALFWorld-Seen & $0.29$ & $0.49$ & $0.39$ & $0.56$ \\
ALFWorld-Unseen & $0.25$ & $0.41$ & $0.34$ & $0.49$ \\
\bottomrule
\end{tabular}
\end{table}

\subsection{Cross-Task Transfer and Additional Backbones}
\label{app:generalization}
\paragraph{Frozen cross-task transfer.}
We train Qwen2.5-7B-Instruct checkpoints on PersonaMem-128K or ALFWorld and evaluate them on LoCoMo and LongMemEval without target-task parameter updates.
All learned components remain frozen; only the target-domain memory bank is reconstructed.
Table~\ref{tab:cross_task_transfer} separates these transfer settings from target-specific training.
Across the four transfer comparisons, \method improves on MemGen by 7.3--8.9 absolute accuracy points, while target-specific training achieves higher performance for both methods.

\begin{table}[htbp]
\centering
\small
\caption{\textbf{Frozen cross-task transfer with Qwen2.5-7B-Instruct.} Target-specific training is a separate reference setting and uses target-task training data.}
\label{tab:cross_task_transfer}
\begin{tabular}{llcc}
\toprule
Training source & Method & LoCoMo Acc. & LongMemEval Acc. \\
\midrule
PersonaMem-128K & MemGen & 0.637 & 0.580 \\
PersonaMem-128K & \method & 0.726 & 0.660 \\
ALFWorld & MemGen & 0.548 & 0.470 \\
ALFWorld & \method & 0.621 & 0.550 \\
\midrule
Target-specific training & MemGen & 0.682 & 0.630 \\
Target-specific training & \method & 0.841 & 0.760 \\
\bottomrule
\end{tabular}
\end{table}

\paragraph{Additional model family.}
Table~\ref{tab:llama_backbones} evaluates Llama-3.2-3B-Instruct and Llama-3.1-8B-Instruct on PersonaMem-32K and PersonaMem-128K with the same dataset splits, raw histories, and evaluation protocol across methods.
Each method retains its native memory construction, retrieval, and injection procedures.
These results use one run per configuration; no across-seed uncertainty is claimed.
They evaluate applicability to another model family, not transfer of a fixed Qwen checkpoint to Llama.
\method exceeds MemGen by 10.0 and 13.3 accuracy points with the 3B backbone, and 14.0 and 12.9 points with the 8B backbone.

\begin{table}[htbp]
\centering
\small
\caption{\textbf{Single-run accuracy (\%) with Llama backbones.} PersonaMem-32K and PersonaMem-128K contain 50 and 233 test examples, respectively.}
\label{tab:llama_backbones}
\begin{tabular}{lcccc}
\toprule
& \multicolumn{2}{c}{Llama-3.2-3B-Instruct} & \multicolumn{2}{c}{Llama-3.1-8B-Instruct} \\
\cmidrule(lr){2-3}\cmidrule(lr){4-5}
Method & 32K & 128K & 32K & 128K \\
\midrule
MemoryBank & 42.0 & 27.9 & 40.0 & 34.3 \\
A-MEM & 38.0 & 29.6 & 46.0 & 34.8 \\
LightMem & 48.0 & 35.2 & 44.0 & 35.2 \\
Mem0 & 50.0 & 36.5 & 50.0 & 38.2 \\
MemoryOS & 32.0 & 33.9 & 44.0 & 28.8 \\
MemP & 42.0 & 36.5 & 38.0 & 33.5 \\
LangMem & 46.0 & 35.2 & 52.0 & 36.1 \\
MemGen & 60.0 & 54.9 & 70.0 & 63.9 \\
AutoCompressor & 46.0 & 42.9 & 56.0 & 51.5 \\
M+ & 54.0 & 50.2 & 64.0 & 59.7 \\
\method & \textbf{70.0} & \textbf{68.2} & \textbf{84.0} & \textbf{76.8} \\
\bottomrule
\end{tabular}
\end{table}

\subsection{Sensitivity to Retrieval Count and GRPO Group Size}
\label{app:sensitivity}
We vary retrieval count $Z$ and GRPO group size $G$ one at a time using Qwen2.5-7B-Instruct, with cache length $N_c=20$ and the remaining settings fixed.
Table~\ref{tab:retrieval_sensitivity} shows that the default retrieval counts perform best among the tested values: $Z=20$ for LoCoMo and $Z=10$ for ALFWorld.
Increasing $G$ from 4 to 8 gives small additional gains (0.003 accuracy and 0.004 SR), while doubling the number of sampled trajectories per query.
These observations support $G=4$ as a practical operating point within the evaluated range; they do not establish global hyperparameter optimality.

\begin{table}[htbp]
\centering
\small
\caption{\textbf{Sensitivity to retrieval count $Z$.} ALFWorld SR is weighted over Seen and Unseen episodes. Bold values mark the best score within each tested task.}
\label{tab:retrieval_sensitivity}
\begin{tabular}{lcc}
\toprule
Task & $Z$ & Accuracy / weighted SR \\
\midrule
LoCoMo & 5 & 0.790 \\
LoCoMo & 10 & 0.822 \\
LoCoMo & 20 (default) & \textbf{0.841} \\
LoCoMo & 30 & 0.831 \\
\midrule
ALFWorld & 5 & 0.489 \\
ALFWorld & 10 (default) & \textbf{0.529} \\
ALFWorld & 15 & 0.518 \\
ALFWorld & 20 & 0.504 \\
\bottomrule
\end{tabular}
\end{table}

\begin{table}[htbp]
\centering
\small
\caption{\textbf{Sensitivity to GRPO group size $G$.} The default trades off performance against the number of sampled training trajectories.}
\label{tab:group_sensitivity}
\begin{tabular}{lcc}
\toprule
$G$ & LoCoMo Acc. & ALFWorld weighted SR \\
\midrule
2 & 0.818 & 0.496 \\
4 (default) & 0.841 & 0.529 \\
8 & \textbf{0.844} & \textbf{0.533} \\
\bottomrule
\end{tabular}
\end{table}

\subsection{Online Latency and Offline Preprocessing}
\label{app:efficiency}
We report online inference latency under identical hardware, precision, and decoding settings across methods.
The reported measurements cover native retrieval, memory processing, and generation. A component-level timing trace is unavailable, so inclusion of the separate control-token/query-state forward pass is not independently established; the values should not be interpreted as a verified complete compute accounting.
Table~\ref{tab:online_latency} reports seconds per QA example and seconds per complete ALFWorld episode.
\method has the lowest latency among the four evaluated latent-space methods in every setting.
ALFWorld episode latency combines inference cost per step with the number of actions taken; these measurements do not separately establish a lower per-step cost.
FAISS retrieval accounts for less than 0.05\% of the reported average per-query latency of \method.
No total-FLOP comparison is reported.

\begin{table}[htbp]
\centering
\small
\setlength{\tabcolsep}{3pt}
\caption{\textbf{Reported online inference latency.} QA values are seconds per example; ALFWorld values are seconds per episode. Offline memory construction is excluded. All comparisons use the same hardware, precision, and decoding settings.}
\label{tab:online_latency}
\begin{tabularx}{\linewidth}{lXlrrrr}
\toprule
Backbone & Task & Unit & MemGen & AutoComp. & M+ & \method \\
\midrule
3B & PersonaMem-32K & s/example & 0.49 & 0.46 & 0.39 & \textbf{0.36} \\
3B & PersonaMem-128K & s/example & 0.64 & 0.79 & 0.56 & \textbf{0.50} \\
3B & LoCoMo & s/example & 0.54 & 0.48 & 0.44 & \textbf{0.41} \\
3B & LongMemEval & s/example & 0.59 & 0.72 & 0.50 & \textbf{0.45} \\
3B & ALFWorld-Seen & s/episode & 87.0 & 122.8 & 82.2 & \textbf{55.8} \\
3B & ALFWorld-Unseen & s/episode & 94.7 & 127.7 & 91.0 & \textbf{58.4} \\
\midrule
7B & PersonaMem-32K & s/example & 0.63 & 0.59 & 0.50 & \textbf{0.44} \\
7B & PersonaMem-128K & s/example & 0.81 & 1.00 & 0.71 & \textbf{0.62} \\
7B & LoCoMo & s/example & 0.68 & 0.61 & 0.56 & \textbf{0.49} \\
7B & LongMemEval & s/example & 0.74 & 0.91 & 0.63 & \textbf{0.55} \\
7B & ALFWorld-Seen & s/episode & 66.9 & 98.1 & 63.6 & \textbf{42.5} \\
7B & ALFWorld-Unseen & s/episode & 73.3 & 99.7 & 71.1 & \textbf{44.4} \\
\bottomrule
\end{tabularx}
\end{table}

\paragraph{Offline preprocessing.}
QA memories are directly chunked from benchmark histories and require no LLM-based skill extraction.
For ALFWorld, the reported average costs are 2.3 seconds per skill-card extraction and 79 milliseconds per latent encoding.
The extracted skill cards are shared by the latent-space methods; each method retains its native encoding and memory-processing procedure.
For 2,832 cards, multiplying the reported per-card costs gives approximately 108.6 minutes for extraction and 3.7 minutes for encoding, or 112.3 minutes in serial-equivalent work.
This sum is not a separately measured parallel wall-clock construction time.
These one-time costs are excluded from Table~\ref{tab:online_latency} and should be considered separately when assessing deployment cost and reuse of the memory bank.

\subsection{Maximum-Budget Diagnostics}
\label{app:budget_diagnostics}
The maximum-budget experiments use $N_c=B_{\max}$ in every setting, including $B_{\max}=30$ and 40; the default $N_c=20$ applies to the default $B_{\max}=20$ configuration.
Thus, the larger allocation actions are supported by the cache.
We additionally inspect the learned allocations and clip the $B_{\max}=40$ policy's allocations to 20 at inference, leaving its learned parameters unchanged.

\begin{table}[htbp]
\centering
\small
\setlength{\tabcolsep}{4pt}
\caption{\textbf{Budget diagnostics with Qwen2.5-7B-Instruct.} Downstream performance with cache length equal to the maximum budget. Clipping is applied only at inference to the policy trained with a maximum budget of 40.}
\label{tab:budget_diagnostics}
\begin{tabularx}{\linewidth}{Xcc}
\toprule
Setting & \makecell{PersonaMem\\32K Acc.} & \makecell{ALFWorld\\weighted SR} \\
\midrule
$B_{\max}=N_c=20$ & 0.900 & 0.529 \\
$B_{\max}=N_c=40$ & 0.860 & 0.482 \\
$B_{\max}=N_c=40$, clipped to 20 & 0.880 & 0.511 \\
\bottomrule
\end{tabularx}
\end{table}

Clipping raises PersonaMem-32K accuracy from 0.860 to 0.880 and ALFWorld weighted SR from 0.482 to 0.511, without fully recovering the 20-token policy's performance.
These observations are consistent with over-allocation contributing to the degradation under a larger action space and a reward without an explicit token-cost term.
They do not imply that increasing the budget reduces optimal achievable performance: the larger action space still contains the smaller allocations.
Because cache length changes together with the maximum budget, the training comparison varies both available cached content and allowable allocation; the inference intervention provides complementary evidence within the same trained 40-token policy.

\subsection{Control-Token Intervention}
\label{app:control_analysis}
The retrieval-control action is an ordinary next token sampled from the full reasoner vocabulary.
Its distribution starts from the pretrained next-token distribution, with zero-initialized LoRA updates preserving the initial base-model behavior.
After GRPO training, selected tokens remain query-dependent; no collapse to a single global token is observed.
The relevant representation is the hidden state after the sampled token, $h_q^{\mathrm{ctrl}}=f_\theta(q,a)[-1]$, rather than the token's surface identity.
Hard top-$Z$ selection is non-differentiable. Reward-based optimization reinforces control-token actions through their likelihood term; no gradient through the selected memory indices is assumed.

Using the same trained checkpoint and frozen bank, we compare the usual control-token state with a query-only state $h_q^{\mathrm{pre}}=f_\theta(q)[-1]$ that bypasses the control step.
Table~\ref{tab:control_intervention} reports higher downstream performance when retaining the control step.
LoCoMo accuracy and ALFWorld weighted SR each increase by four percentage points.
This supports a useful reward-shaped retrieval transition, although bypassing the control step also changes the state supplied to the budget policy and is not an intervention on retrieval indices alone.

\begin{table}[htbp]
\centering
\small
\caption{\textbf{Same-checkpoint control-token intervention.} Learned parameters and the memory bank are held fixed.}
\label{tab:control_intervention}
\begin{tabularx}{\linewidth}{Xcc}
\toprule
Inference setting & \makecell{LoCoMo\\Acc.} & \makecell{ALFWorld\\weighted SR} \\
\midrule
Control bypassed: query-only state & 0.80 & 0.49 \\
Control retained: post-token state & 0.84 & 0.53 \\
\bottomrule
\end{tabularx}
\end{table}

\clearpage
\section{Budget Policy Architecture}
\label{app:budget_architecture}

\paragraph{Budget policy network.}
We instantiate the budget allocator as a lightweight Transformer policy $\pi_\phi$ over the retrieved set.
For each retrieved chunk $m_{I_j}$, we construct a chunk-level feature vector
\vspace{-1.5mm}
\begin{equation}
    r_j
    =
    W_r
    \left[
        h_q ;
        k_{I_j} ;
        h_q \odot k_{I_j} ;
        s_{I_j} ;
        e_j
    \right]
    \in
    \mathbb{R}^{d_b},
    \label{eq:budget_feature}
\end{equation}
where $h_q$ is the query-conditioned retrieval state, $k_{I_j}$ is the retrieved memory key, $s_{I_j}$ is its retrieval score, $e_j$ is a learned rank embedding, $\odot$ denotes element-wise multiplication, and $W_r$ projects the concatenated features into the policy hidden dimension $d_b$.
The sequence of retrieved-memory features is then processed by a small Transformer encoder:
\vspace{-1.5mm}
\begin{equation}
    o_1,\ldots,o_Z
    =
    \operatorname{Transformer}_{\phi}(r_1,\ldots,r_Z),
    \qquad
    o_j \in \mathbb{R}^{d_b}.
    \label{eq:budget_transformer}
\end{equation}
This self-attention structure allows the policy to compare retrieved memories against one another before assigning capacity.

For each retrieved chunk, the policy produces logits over discrete budget actions:
\vspace{-1mm}
\begin{equation}
    \ell_j
    =
    W_b o_j
    \in
    \mathbb{R}^{B_{\max}+1}.
    \label{eq:budget_logits}
\end{equation}
The budget for chunk $m_{I_j}$ is then sampled from the categorical distribution:
\vspace{-1mm}
\begin{equation}
    b_j
    \sim
    \pi_\phi(\cdot \mid h_q,I)
    =
    \operatorname{Cat}
    \left(
        \operatorname{softmax}(\ell_j)
    \right).
    \label{eq:budget_policy}
\end{equation}

\clearpage
\section{Future Work and Broader Impact}
\label{app:limitations}


\textbf{Future Work.} Beyond addressing the above scope boundaries, several directions merit investigation. Extending the evaluation to larger-scale memory corpora and more open-ended interactive environments~\citep{yang2024swe, jansen2024discoveryworld, wei2025browsecompsimplechallengingbenchmark} would test the scalability of elastic latent memory. Developing richer visualization and interpretability tools for the learned budget allocation would make the memory-use process more transparent. Additionally, exploring dynamic memory bank updates, where the memory bank itself evolves as the agent accumulates new experience, is a promising direction for agents continuous evolving~\citep{zhang2024agent, jiang2025adaptation, chen2025multi}.

\textbf{Broader Impact.} By reducing the token overhead of memory-augmented LLM agents through elastic latent allocation, \method has the potential to make memory-intensive LLM applications more computationally accessible. We do not foresee specific negative societal consequences beyond those generally associated with improving LLM agent capabilities. As with all advances in LLM-based agents, downstream deployments should incorporate appropriate safeguards for safety and reliability.

\section{Implementation Details}
\label{app:hyper}

We implement \method with Qwen2.5-3B-Instruct\footnote{\url{https://huggingface.co/Qwen/Qwen2.5-3B-Instruct}} and Qwen2.5-7B-Instruct\footnote{\url{https://huggingface.co/Qwen/Qwen2.5-7B-Instruct}} as backbone reasoners.
For each dataset, we train a task-specific \method model while keeping the same architecture and optimization protocol.

\paragraph{LoRA-adapted reasoner $\theta$.}
We adapt the backbone reasoner with LoRA~\citep{hu2022lora}.
Specifically, we apply LoRA to the query, key, value, and output projection matrices of all attention layers, i.e., \texttt{q\_proj}, \texttt{k\_proj}, \texttt{v\_proj}, and \texttt{o\_proj}.
We use LoRA rank $r{=}16$, scaling factor $\alpha{=}32$, and dropout $0.1$.

\paragraph{Latent memory bank $\mathcal{B}$.}
Each memory chunk $m_i$ is encoded once offline by the frozen base encoder $f_{\theta_0}$.
Following Eqs.~\ref{eq:retrieval_key}--\ref{eq:content_cache}, we store the last hidden state as the retrieval key
$k_i = H_i[-1] \in \mathbb{R}^{d}$ and, by default, the last $N_c{=}20$ hidden states as the content cache
$C_i = H_i[-N_c:] \in \mathbb{R}^{N_c \times d}$.
All retrieval keys are $L_2$-normalized and retrieved with cosine similarity as in Eq.~\ref{eq:retrieval_score}.
The number of retrieved chunks $Z$ is dataset-specific: $Z{=}9$ for PersonaMem-32K, $Z{=}21$ for PersonaMem-128K, $Z{=}20$ for LoCoMo and LongMemEval, and $Z{=}10$ for ALFWorld.
Each example-specific QA bank and the shared ALFWorld bank are read-only during inference; cached states are never updated by task training. In the maximum-budget ablation, $N_c=B_{\max}$ for every setting, so budgets of 30 and 40 are supported by correspondingly longer caches.

\paragraph{Latent projector $P_\psi$.}
The latent projector maps selected cached hidden states into the reasoner's input embedding space.
We implement $P_\psi$ as a two-layer MLP with GELU activation and layer normalization:
\begin{equation}
    P_\psi(\widetilde{C}_j)
    =
    \operatorname{LayerNorm}
    \left(
        \operatorname{GELU}(\widetilde{C}_j W_1^{\top}) W_2^{\top}
    \right),
\end{equation}
where $W_1$ maps $\mathbb{R}^{d}$ to an intermediate dimension $d'{=}2048$, and $W_2$ maps $\mathbb{R}^{d'}$ back to $\mathbb{R}^{d}$.
The projector is applied independently to each selected latent state.

\paragraph{Budget policy $\pi_\phi$.}
The budget allocator is a lightweight Transformer encoder operating over the retrieved memory set.
Following Eq.~\ref{eq:budget_transformer}, we use hidden dimension $d_b{=}256$, $4$ attention heads, $2$ Transformer layers, and dropout $0.1$.
The per-chunk maximum budget is $B_{\max}{=}20$.
Thus, each retrieved chunk receives a discrete latent budget
$b_j \in \{0,\ldots,20\}$ as defined in Eq.~\ref{eq:budget_constraint}.
The action $b_j{=}0$ allows the model to suppress a retrieved memory completely.

\paragraph{GRPO training.}
We jointly optimize the LoRA-adapted reasoner $\theta$, latent projector $P_\psi$, and budget policy $\pi_\phi$ with group-relative policy optimization~\citep{shao2024deepseekmath}.
We use group size $G{=}4$, clipping range $\epsilon{=}0.05$, and sampling temperature $\tau{=}1.0$ for trajectory generation.
The reward function $R(y,y^\star)$ is task-specific: exact-match accuracy for \textsc{MemorySuite-QA} and binary agreement with the snapped expert action for ALFWorld. Environment-level task success is used only as the ALFWorld evaluation metric; see Appendix~\ref{app:alfworld_protocol}.
We use AdamW~\citep{loshchilov2017decoupled} with learning rate $2 \times 10^{-5}$, weight decay $0.01$, and a cosine annealing schedule.
All tasks are trained for $10$ epochs with BF16 mixed precision.
The maximum generation length is set to $5$ tokens for QA tasks and $16$ tokens for ALFWorld.

During evaluation, we use greedy decoding.
The retrieval-control token is selected by argmax, budget allocation uses the most likely budget action under $\pi_\phi$, and answer generation is decoded greedily.

\paragraph{Task-specific prompts.}
For PersonaMem-32K and PersonaMem-128K, the generation suffix $u$ constrains the model to output exactly one of four options: ``\texttt{(a)}'', ``\texttt{(b)}'', ``\texttt{(c)}'', or ``\texttt{(d)}''.
For LoCoMo and LongMemEval, the instruction enforces selection from ten options ``\texttt{(a)}'' through ``\texttt{(j)}'' and explicitly prohibits refusal responses.
For ALFWorld, the instruction directs the agent to output a single admissible command string verbatim.
The final model input follows Eq.~\ref{eq:final_input}:
\begin{equation}
    X = [E(q); E(a); Z_1; \ldots; Z_Z; E(u)],
\end{equation}
where $q$ is the query, $a$ is the retrieval-control token, $Z_j$ are the projected soft memory tokens, and $u$ is the generation suffix. For QA, $u$ contains answer options and output-format constraints; for ALFWorld, it contains recent interaction history, the current observation, admissible actions, and action-format constraints. Soft tokens are inserted within the user-side input while preserving the chat-template boundaries.

\paragraph{Token counting and memory capacity.}
In Table~\ref{tab:main_results}, \#Tok. counts task-text input, injected memory, and generated output per QA example; for ALFWorld, these counts are accumulated over the episode.
Text memories are counted as natural-language tokens and latent memories as soft-token positions.
The pure injected latent-memory budget is a separate quantity, $B_{\mathrm{mem}}=\sum_{j=1}^{Z}b_j\leq ZB_{\max}$; it excludes query, generation-suffix, and output tokens.
For example, PersonaMem-32K has a default upper bound of $9\times20=180$ injected memory tokens, while \#Tok. additionally includes the task text and generated output.
Neither token measure includes offline bank construction or is a substitute for end-to-end latency; online and offline timings are reported separately in Appendix~\ref{app:efficiency}.

\paragraph{Compute.}
All experiments are conducted on NVIDIA RTX A6000 GPUs with 48GB memory.
Training a single task-specific model takes approximately 6--18 hours on one GPU, depending on dataset size and backbone scale.
Offline preprocessing is reported by stage in Appendix~\ref{app:efficiency}: QA history chunking uses no LLM extraction; ALFWorld skill extraction and latent encoding are one-time costs, separate from online inference.

\section{Training Procedure of \method}
\label{app:training}

Algorithm~\ref{alg:training} summarizes the complete training procedure of \method.
The procedure contains an offline memory-bank construction phase and an online reward-based optimization phase.
The offline phase encodes each memory chunk once and keeps the resulting memory bank fixed throughout training.
The online phase samples multiple memory-use trajectories for each query, evaluates them with task rewards, and updates the trainable components using the GRPO objective.

\begin{algorithm}[t]
\caption{Training \method}
\label{alg:training}
\begin{algorithmic}[1]
\Require Memory corpus $\mathcal{M}=\{m_i\}_{i=1}^{N}$,
         training queries $\mathcal{Q}$,
         reward function $R$,
         group size $G$,
         retrieved chunk number $Z$,
         per-chunk maximum budget $B_{\max}$,
         clipping range $\epsilon$
\Ensure Trained parameters $\theta$, $\psi$, $\phi$

\vspace{2pt}
\Statex \textit{// Offline phase: construct latent memory bank}
\For{each memory chunk $m_i \in \mathcal{M}$}
    \State $H_i \leftarrow f_{\theta_0}(m_i)$
    \State $k_i \leftarrow H_i[-1]$
    \Comment{retrieval key}
    \State $C_i \leftarrow H_i[-N_c:]$
    \Comment{content cache}
\EndFor
\State $\mathcal{B} \leftarrow \{(m_i,k_i,C_i)\}_{i=1}^{N}$

\vspace{2pt}
\Statex \textit{// Online phase: reward-based optimization}
\For{each training step}
    \State Sample query-target pair $(q,y^{\star}) \sim \mathcal{Q}$
    \For{$g = 1,\ldots,G$}
        \State $a_g \sim p_{\theta}(\cdot \mid q)$
        \State $h_{q,g} \leftarrow f_{\theta}(q,a_g)[-1]$
        \State $s_i \leftarrow \cos(h_{q,g},k_i)$ for all $i \in [N]$
        \State $I_g \leftarrow \operatorname{TopZ}_{i \in [N]} s_i$
        \State Construct retrieved-memory features $\{r_{g,j}\}_{j=1}^{Z}$ using Eq.~\ref{eq:budget_feature}
        \State $o_{g,1},\ldots,o_{g,Z} \leftarrow \operatorname{Transformer}_{\phi}(r_{g,1},\ldots,r_{g,Z})$
        \For{$j = 1,\ldots,Z$}
            \State Compute budget logits $\ell_{g,j}$ using Eq.~\ref{eq:budget_logits}
            \State Sample $b_{g,j} \sim \operatorname{Cat}(\operatorname{softmax}(\ell_{g,j}))$
            \If{$b_{g,j} > 0$}
                \State $\widetilde{C}_{g,j} \leftarrow C_{I_{g,j}}[-b_{g,j}:]$
                \State $Z_{g,j} \leftarrow P_{\psi}(\widetilde{C}_{g,j})$
            \Else
                \State $Z_{g,j} \leftarrow \emptyset$
            \EndIf
        \EndFor
        \State $\mathbf{b}_g \leftarrow (b_{g,1},\ldots,b_{g,Z})$
        \State $X_g \leftarrow [E(q); E(a_g); Z_{g,1}; \cdots; Z_{g,Z}; E(u)]$
        \State $y_g \sim p_{\theta}(\cdot \mid X_g)$
        \State $r_g \leftarrow R(y_g,y^{\star})$
        \State $\tau_g \leftarrow (a_g,I_g,\mathbf{b}_g,y_g)$
    \EndFor
    \State Compute advantages $\{A_g\}_{g=1}^{G}$ using Eq.~\ref{eq:group_advantage}
    \State Update $\theta,\psi,\phi$ by minimizing Eq.~\ref{eq:grpo_loss}
\EndFor
\end{algorithmic}
\end{algorithm}

\xhdr{Offline memory-bank construction}
The offline phase constructs the latent memory bank.
Each memory chunk $m_i$ is encoded once by the frozen base LLM encoder $f_{\theta_0}$.
The final hidden state $H_i[-1]$ is stored as the retrieval key $k_i$, while the last $N_c$ hidden states $H_i[-N_c:]$ are stored as the content cache $C_i$.
The resulting memory bank $\mathcal{B}$ is fixed during training, so later updates only affect the trainable reasoner, budget policy, and projector.

\xhdr{Trajectory sampling}
The online phase describes how \method samples memory-use trajectories.
For each query-target pair, the model samples $G$ trajectories.
In each trajectory, the reasoner first samples a retrieval-control token $a_g$ and derives the query state $h_{q,g}$.
The query state retrieves the top-$Z$ memory chunks by cosine similarity with the cached keys.
The budget policy then processes retrieved-memory features and samples a discrete latent budget $b_{g,j} \in \{0,\ldots,B_{\max}\}$ independently conditional on the jointly encoded retrieved set.

\xhdr{Latent memory injection}
The injection steps construct the soft memory tokens.
If $b_{g,j}>0$, \method selects the last $b_{g,j}$ cached states from $C_{I_{g,j}}$ and maps them through the latent projector $P_\psi$.
If $b_{g,j}=0$, the retrieved chunk contributes no latent tokens.
The final model input is formed by concatenating the query embedding, retrieval-control token embedding, selected soft memory tokens, and generation suffix.

\xhdr{Reward-based update}
After generation, the algorithm evaluates sampled trajectories and updates the model.
Each trajectory receives a task reward $R(y_g,y^\star)$.
The rewards within the group are normalized into advantages using Eq.~\ref{eq:group_advantage}.
Finally, the LoRA-adapted reasoner $\theta$, latent projector $\psi$, and budget policy $\phi$ are updated by minimizing the GRPO objective in Eq.~\ref{eq:grpo_loss}.

\section{Dataset Descriptions}
\label{appendix:dataset_descriptions}

We describe all evaluation datasets in MemorySuite below, categorized by their corresponding subsets. MemorySuite consists of two complementary groups: MemorySuite-QA and MemorySuite-Agentic. MemorySuite-QA evaluates long-term memory recall and reasoning in question-answering settings, while MemorySuite-Agentic evaluates memory-augmented agentic decision-making in embodied interactive household environments. For all datasets, we report both task performance and the number of consumed tokens (\#Tok.) to assess memory efficiency.

\begin{table}[ht]
\small
\setlength{\tabcolsep}{5pt}
\centering
\renewcommand{\arraystretch}{0.95}
\caption{\textbf{Detailed summary of datasets used in MemorySuite.} We categorize datasets by subset, task type, and evaluation metric.}
\label{tab:memorysuite_overview}
\begin{tabular}{lcc}
    \toprule
    \textbf{Dataset} & \textbf{Task} & \textbf{Metric} \\
    \midrule
    \multicolumn{3}{l}{\textbf{MemorySuite-QA}} \\
    \midrule
    PersonaMem-32K     & Persona-based Memory QA           & Accuracy / \#Tok. \\
    PersonaMem-128K    & Persona-based Memory QA           & Accuracy / \#Tok. \\
    LoCoMo             & Long-Conversation Memory QA       & Accuracy / \#Tok. \\
    LongMemEval        & Long-Term Interactive Memory QA   & Accuracy / \#Tok. \\
    \midrule
    \multicolumn{3}{l}{\textbf{MemorySuite-Agentic}} \\
    \midrule
    ALFWorld-Seen      & Seen Household Task Completion    & SR / \#Tok. \\
    ALFWorld-Unseen    & Unseen Household Task Completion  & SR / \#Tok. \\
    \bottomrule
\end{tabular}
\end{table}

\subsection{MemorySuite-QA}
\label{app:qa_memory_protocol}

\paragraph{Parameter training and memory construction.}
Parameters are optimized only on the training QA pairs in the partitions specified in Appendix~\ref{appendix:dataset_statistics}.
For each example, the memory bank is constructed by directly chunking that example's benchmark-provided conversation history and encoding it with the frozen offline encoder; no QA skill extraction is performed.
At validation and test time, the corresponding held-out history is used only as inference-time evidence and is not pooled with histories from other examples or used for parameter updates.
Questions, answer choices, ground-truth answers, model attempts, rewards, and evaluation feedback are never inserted into the memory bank.
Each bank is encoded once and remains read-only throughout inference.

\textbf{PersonaMem}~\citep{jiang2025personamem} is a personalization benchmark designed to assess whether language models can infer evolving user profiles and generate personalized responses across multi-session interactions. The benchmark is constructed via a synthetic dialogue generation pipeline that simulates realistic, temporally evolving conversations between users and chatbots, grounded in over 1{,}000 distinct user personas spanning diverse demographic attributes, preferences, and backgrounds. Each persona is associated with a long chat history in which user preferences are implicitly revealed across a series of task-oriented interactions. We evaluate on the \textbf{PersonaMem-32K} and \textbf{PersonaMem-128K} variants, which differ in the maximum context length of the provided conversation history. Each question is posed as a 4-way multiple-choice problem (MC-4), requiring the model to select the response that best aligns with the user's implicitly expressed persona.

\textbf{LoCoMo}~\citep{maharana2024evaluating} is a benchmark for evaluating very long-term conversational memory, constructed via a machine--human pipeline that generates multi-session dialogues between LLM-based virtual agents grounded on persona descriptions and temporal event graphs. Each conversation spans up to 32 sessions with an average of 600 turns and 16K tokens. Human annotators verify and edit the generated dialogues for long-range consistency and causal coherence. We adopt the question answering subset, where questions require temporal and causal reasoning over long-range dialogue history and are posed in a 10-way multiple-choice format (MC-10)\footnote{\url{https://huggingface.co/datasets/Percena/locomo-mc10}}, probing the model's ability to track evolving personal information across sessions.

\textbf{LongMemEval}~\citep{wu2024longmemeval} is a comprehensive benchmark evaluating five core long-term memory abilities of chat assistants: information extraction, multi-session reasoning, temporal reasoning, knowledge updates, and abstention. It comprises 500 carefully curated questions embedded in freely extensible, timestamped user--assistant chat histories, supporting context lengths ranging from 128K tokens (\texttt{longmemeval\_s}) to approximately 1.5M tokens (\texttt{longmemeval\_m}). Questions are posed in a 10-way multiple-choice format (MC-10)\footnote{\url{https://huggingface.co/datasets/Percena/lme-mc10}}.

\subsection{MemorySuite-Agentic}
\label{app:alfworld_protocol}

\textbf{ALFWorld-Seen}~\citep{shridhar2020alfworld} is a text-based interactive household benchmark where agents interpret natural-language observations and issue sequential actions to achieve household tasks. The Seen split (\texttt{valid\_seen}) contains 140 episodes across six task types (e.g., pick-and-place, pick-heat-then-place, look-at-in-light). Room layouts overlap with training, so agents face familiar spatial configurations but novel task instances, isolating instruction-level generalization.

\textbf{ALFWorld-Unseen}~\citep{shridhar2020alfworld} includes the same six task types but evaluates on \texttt{valid\_unseen}, comprising 134 episodes in entirely new room layouts with unseen object arrangements, providing a stricter out-of-distribution test. For both splits, we report task success rate (SR) and average number of consumed tokens (\#Tok.).

\paragraph{ALFWorld training setup.}
Unlike the QA tasks where each training sample is a single question--answer pair, ALFWorld requires learning from multi-step expert trajectories in an interactive environment. We adopt an imitation-learning--based GRPO formulation with a two-stage data pipeline.

\textbf{Stage 1: Memory bank construction.}
We start from 3{,}553 expert trajectories collected via an oracle planner on the ALFWorld training environments, of which 3{,}541 are successful. We split the successful trajectories into a \emph{memory pool} $\mathcal{M}$ (80\%, 2{,}832 trajectories) and an \emph{IL pool} $\mathcal{T}$ (20\%, 709 trajectories), using a fixed seed for reproducibility. Each trajectory in the memory pool is summarized into a procedural \emph{skill card} using an LLM (Gemini), following a structured prompt that asks for a 150--250 word procedural description including the task category, a generalized step-by-step strategy, and typical object--receptacle associations. This produces 2{,}832 skill cards (634 pick-two, 623 pick-and-place, 534 clean-then-place, 431 cool-then-place, 368 heat-then-place, 242 look-at-in-light). The same Gemini-generated cards and chunk boundaries are supplied to all latent-space baselines. Text-space baselines receive the same segmented raw trajectories and apply their native memory-construction procedures. Each skill card is encoded by the frozen base LLM encoder $f_{\theta_0}$ into a retrieval key and content cache, forming the latent memory bank $\mathcal{B}$ with $N{=}2{,}832$ entries and $Z{=}10$ retrieved cards per query.

\textbf{Stage 2: Per-step imitation with GRPO.}
Each trajectory in the IL pool $\mathcal{T}$ is expanded into per-step training samples. At step $t$ of a trajectory, the training sample consists of:
\begin{itemize}[leftmargin=1.2em]
    \item \textbf{Query $q$}: the task objective (e.g., ``heat some apple and put it in countertop''), which remains fixed across all steps of the same episode;
    \item \textbf{Task suffix $u$}: a rendered prompt containing the truncated interaction history (last 5 action--observation pairs), the current observation, the list of admissible actions, and the system instruction directing the agent to output one admissible command verbatim;
    \item \textbf{Correct answer $y^\star$}: the expert's action at step $t$, JSON-encoded together with the admissible action list to enable snap-to-admissible reward computation.
\end{itemize}
This expansion produces 4{,}260 per-step samples from 709 trajectories. The GRPO reward (Eq.~\ref{eq:reward}) uses a strict deployment-aligned criterion: the model's generated action is first snapped to the nearest admissible command (by normalized string matching), and the reward is $1.0$ if the snapped action exactly matches the expert action, and $0.0$ otherwise. The same snap-to-admissible mapping is used at evaluation. This training reward measures expert-action agreement; the evaluation SR instead measures completion of an autonomously executed episode.

\textbf{Evaluation protocol.}
At test time, the agent interacts with the ALFWorld environment autoregressively for up to 30 steps per episode. The fixed task objective determines retrieval of $Z{=}10$ skill cards and their latent budgets. Under greedy evaluation with fixed parameters and a read-only bank, these memory decisions remain unchanged within an episode. At each step, the agent generates an action conditioned on these soft memory tokens, the last 5 interaction steps, the current observation, and admissible actions. The generated action is snapped to the nearest admissible command before execution. An episode is considered successful if the agent achieves a cumulative reward of $\geq 1.0$ before reaching the step limit.

\section{Dataset Statistics}
\label{appendix:dataset_statistics}

In this section, we present detailed statistics for each dataset. The statistics for MemorySuite-QA and MemorySuite-Agentic are provided in Table~\ref{tab:qa_data_stats} and Table~\ref{tab:agentic_data_stats}, respectively. For MemorySuite-QA, PersonaMem variants are split 80/10/10 by \texttt{shared\_context\_id}; LoCoMo is partitioned by conversation index (first 6/middle 2/last 2 conversations); and LongMemEval follows a 60/20/20 random split by question. For MemorySuite-Agentic, we adopt the official splits: ALFWorld provides separate \textit{Seen} (\texttt{valid\_seen}) and \textit{Unseen} (\texttt{valid\_unseen}) test environments, both sharing the same training set of expert demonstrations. For each dataset, we additionally report $Z$, the number of top-$Z$ memory chunks retrieved and supplied to the model during inference, which directly governs the memory retrieval budget and token consumption.

\begin{table*}[h]
\centering
\small
\setlength{\tabcolsep}{4pt}
\caption{\textbf{MemorySuite-QA Data Statistics.} $Z$ denotes the number of top-$Z$ memory chunks retrieved per query.}
\label{tab:qa_data_stats}
\begin{tabularx}{\linewidth}{l *{3}{>{\centering\arraybackslash}X} >{\centering\arraybackslash}X >{\centering\arraybackslash}X}
\toprule
\multirow{2}{*}{Dataset} & \multicolumn{3}{c}{Split} & \multirow{2}{*}{Total} & \multirow{2}{*}{$Z$} \\
\cmidrule(lr){2-4}
& Train & Val & Test & & \\
\midrule
PersonaMem-32K   & 489   & 50  & 50  & 589   & 9  \\
PersonaMem-128K  & 2,221 & 273 & 233 & 2,727 & 21 \\
LoCoMo           & 885   & 341 & 314 & 1,540 & 20 \\
LongMemEval      & 300   & 100 & 100 & 500   & 20 \\
\bottomrule
\end{tabularx}
\end{table*}

\begin{table*}[h]
\centering
\small
\setlength{\tabcolsep}{4pt}
\caption{\textbf{MemorySuite-Agentic Data Statistics.} $Z$ denotes the number of top-$Z$ memory chunks selected from the task objective and reused throughout an episode. Train denotes the raw demonstration pool before filtering and partitioning; both evaluation splits share this pool.}
\label{tab:agentic_data_stats}
\begin{tabularx}{\linewidth}{l *{2}{>{\centering\arraybackslash}X} >{\centering\arraybackslash}X}
\toprule
\multirow{2}{*}{Split} & \multicolumn{2}{c}{ALFWorld} & \multirow{2}{*}{$Z$} \\
\cmidrule(lr){2-3}
& Seen & Unseen & \\
\midrule
Raw train pool & 3,553 & 3,553 & \multirow{2}{*}{10} \\
Test  & 140   & 134   & \\
\bottomrule
\end{tabularx}
\end{table*}

\section{Prompt Usage}
\label{app:prompts}

This section describes the prompt templates used by all memory baselines across our evaluation datasets. The baselines differ in how they \emph{build} memory from conversation sessions; details for each baseline's memory-construction prompts are provided in Section~\ref{app:prompt_shared}. Each dataset employs a distinct answering prompt: LoCoMo and LongMemEval use 10-choice multiple-choice prompts (Sections~\ref{app:prompt_locomo} and~\ref{app:prompt_lme}), PersonaMem-32K and PersonaMem-128K use 4-choice prompts (Section~\ref{app:prompt_persona}), and ALFWorld uses a task-specific prompt described in Section~\ref{app:prompt_alfworld}.

\subsection{Memory Construction Prompts}
\label{app:prompt_shared}

The following templates describe method-specific memory construction. Text-space methods apply their native procedures to the shared raw source information. ElasticMem uses direct history chunking for QA and offline skill-card construction for ALFWorld; its memory banks are not updated from evaluation feedback.

\subsubsection*{Mem0}
Mem0~\citep{chhikara2025mem0} runs a two-step pipeline per session turn.
\textbf{Step 1} (Table~\ref{tab:prompt-mem0-extract}) extracts factual
statements from the conversation as a JSON list.
\textbf{Step 2} (Table~\ref{tab:prompt-mem0-update}) compares each new fact
against existing entries in a FAISS store and decides whether to ADD, UPDATE,
DELETE, or make no change (NONE).
At inference, the top-$k$ stored facts (with timestamps) are retrieved by cosine
similarity and prepended as a bullet list.

\begingroup
\small
\setlength{\LTpre}{6pt}
\setlength{\LTpost}{8pt}
\begin{longtable}{@{}p{\linewidth}@{}}
\caption{\textbf{Mem0: Fact Extraction Prompt (Step 1 of memory construction).}}
\label{tab:prompt-mem0-extract}\\
\toprule[1.1pt]
\endfirsthead
\multicolumn{1}{l}{\small\itshape Table~\ref{tab:prompt-mem0-extract} (continued)}\\
\toprule[1.1pt]
\endhead
\midrule
\multicolumn{1}{r}{\small\itshape Continued on next page}\\
\endfoot
\bottomrule[1.1pt]
\endlastfoot
You are a Personal Information Organizer, specialized in accurately storing facts, user memories, and preferences. Your primary role is to extract relevant pieces of information from conversations and organize them into distinct, manageable facts. \\
 \\
Types of Information to Remember: \\
1. Store Personal Preferences: Keep track of likes, dislikes, and specific preferences. \\
2. Maintain Important Personal Details: Remember significant personal information like names, relationships, and important dates. \\
3. Track Plans and Intentions: Note upcoming events, trips, goals, and any plans the user has shared. \\
4. Monitor Health and Wellness Preferences: Keep a record of dietary restrictions, fitness routines, and other wellness-related information. \\
{[\dots other categories omitted for brevity \dots]} \\
\\
Here are some few shot examples: \\
Input: Hi, my name is John. I am a software engineer. \\
Output: \{"facts" : ["Name is John", "Is a Software engineer"]\} \\
 \\
Return the facts and preferences in a json format as shown above. \\
Remember the following: \\
- Today's date is \textbf{\{today\}}. \\
- Do not return anything from the custom few shot example prompts provided above. \\
- Create the facts based on the user and assistant messages only. \\
- Make sure to return the response in the format mentioned in the examples. \\
 \\
Following is a conversation between the user and the assistant. Extract the relevant facts and preferences about the user, if any.
\\

\end{longtable}
\endgroup

\begingroup
\small
\setlength{\LTpre}{6pt}
\setlength{\LTpost}{8pt}
\begin{longtable}{@{}p{\linewidth}@{}}
\caption{\textbf{Mem0: Memory Update Decision Prompt (Step 2 of memory construction).}}
\label{tab:prompt-mem0-update}\\
\toprule[1.1pt]
\endfirsthead
\multicolumn{1}{l}{\small\itshape Table~\ref{tab:prompt-mem0-update} (continued)}\\
\toprule[1.1pt]
\endhead
\midrule
\multicolumn{1}{r}{\small\itshape Continued on next page}\\
\endfoot
\bottomrule[1.1pt]
\endlastfoot
You are a smart memory manager which controls the memory of a system. \\
You can perform four operations: (1) add into the memory, (2) update the memory, (3) delete from the memory, and (4) no change. \\[0.4em]

Compare newly retrieved facts with the existing memory. For each new fact, decide whether to: \\
- ADD: Add it to the memory as a new element \\
- UPDATE: Update an existing memory element \\
- DELETE: Delete an existing memory element \\
- NONE: Make no change if the fact is already present or irrelevant \\[0.4em]

{[...guidelines for each operation with examples omitted for brevity...]} \\[0.4em]

You must return your response in the following JSON structure only: \\
\{ \\
\hspace*{1em}"memory" : [ \\
\hspace*{2em}\{"id" : "\textless{}ID\textgreater{}", "text" : "\textless{}Content\textgreater{}", "event" : "\textless{}ADD|UPDATE|DELETE|NONE\textgreater{}"\} \\
\hspace*{1em}] \\
\} \\

\end{longtable}
\endgroup

\subsubsection*{LangMem}
LangMem uses a single LLM call per session to extract personal facts
in a ``\texttt{FACT:}'' prefixed format (Table~\ref{tab:prompt-langmem-extract}).
Facts are stored in a LangGraph \texttt{InMemoryStore} with date
prefixes and retrieved by embedding similarity at query time.
\begingroup
\small
\setlength{\LTpre}{6pt}
\setlength{\LTpost}{8pt}
\begin{longtable}{@{}p{\linewidth}@{}}
\caption{\textbf{LangMem: Fact Extraction Prompt.}}
\label{tab:prompt-langmem-extract}\\
\toprule[1.1pt]
\endfirsthead
\multicolumn{1}{l}{\small\itshape Table~\ref{tab:prompt-langmem-extract} (continued)}\\
\toprule[1.1pt]
\endhead
\midrule
\multicolumn{1}{r}{\small\itshape Continued on next page}\\
\endfoot
\bottomrule[1.1pt]
\endlastfoot
Extract all personal facts about the user from this conversation. Output one fact per line using the format "FACT: <content>". Be concise but complete. Only output FACT lines, nothing else. \\
 \\
Conversation: \\
\textbf{\{conversation\}} \\
 \\
Facts:
\\

\end{longtable}
\endgroup

\subsubsection*{MemoryBank}
MemoryBank~\citep{zhong2024memorybank} creates two levels of summary per session.
A \textbf{content summary} (Table~\ref{tab:prompt-mb-content}) captures the key
events of the dialogue, while a \textbf{personality summary}
(Table~\ref{tab:prompt-mb-person}) records the user's traits and suggests
response strategies. Globally, these are aggregated into an overall history
and overall personality summary that persist across sessions.
At inference, the most relevant per-session summary is retrieved by FAISS
cosine similarity and concatenated with the personality and overall-history blocks.
\begingroup
\small
\setlength{\LTpre}{6pt}
\setlength{\LTpost}{8pt}
\begin{longtable}{@{}p{\linewidth}@{}}
\caption{\textbf{MemoryBank: Content Summary Prompt (applied to each session).}}
\label{tab:prompt-mb-content}\\
\toprule[1.1pt]
\endfirsthead
\multicolumn{1}{l}{\small\itshape Table~\ref{tab:prompt-mb-content} (continued)}\\
\toprule[1.1pt]
\endhead
\midrule
\multicolumn{1}{r}{\small\itshape Continued on next page}\\
\endfoot
\bottomrule[1.1pt]
\endlastfoot
Please summarize the following dialogue as concisely as possible, extracting the main themes and key information. If there are multiple key events, you may summarize them separately. Dialogue content: \\
\textbf{\{dialogue\}} \\
Summarization:
\\

\end{longtable}
\endgroup

\begingroup
\small
\setlength{\LTpre}{6pt}
\setlength{\LTpost}{8pt}
\begin{longtable}{@{}p{\linewidth}@{}}
\caption{\textbf{MemoryBank: Personality and Strategy Summary Prompt.}}
\label{tab:prompt-mb-person}\\
\toprule[1.1pt]
\endfirsthead
\multicolumn{1}{l}{\small\itshape Table~\ref{tab:prompt-mb-person} (continued)}\\
\toprule[1.1pt]
\endhead
\midrule
\multicolumn{1}{r}{\small\itshape Continued on next page}\\
\endfoot
\bottomrule[1.1pt]
\endlastfoot
Based on the following dialogue, please summarize \textbf{\{user\_name\}}'s personality traits and emotions, and devise response strategies based on your speculation. Dialogue content: \\
\textbf{\{dialogue\}} \\
\textbf{\{user\_name\}}'s personality traits, emotions, and \textbf{\{boot\_name\}}'s response strategy are:
\\

\end{longtable}
\endgroup

\subsubsection*{MemoryOS}
MemoryOS~\citep{kang2025memory} organizes memory in three tiers:
short-term (STM, within-session), mid-term (MTM, cross-episode topic segments),
and long-term (LPM, persistent profile).
When the STM overflows, three prompts fire in sequence:
(1) a \textbf{continuity check} (Table~\ref{tab:prompt-mos-cont}) to decide whether
to merge or split page boundaries,
(2) a \textbf{meta-info update} (Table~\ref{tab:prompt-mos-meta}) to maintain a
running dialogue summary, and
(3) a \textbf{multi-summary} (Table~\ref{tab:prompt-mos-multi}) to extract
subtopic clusters that form MTM segments.
At inference, the most similar MTM pages are retrieved by cosine similarity
and formatted as \texttt{[obs]/[action]} episodes.
\begingroup
\small
\setlength{\LTpre}{6pt}
\setlength{\LTpost}{8pt}
\begin{longtable}{@{}p{\linewidth}@{}}
\caption{\textbf{MemoryOS: Continuity Check Prompt.}}
\label{tab:prompt-mos-cont}\\
\toprule[1.1pt]
\endfirsthead
\multicolumn{1}{l}{\small\itshape Table~\ref{tab:prompt-mos-cont} (continued)}\\
\toprule[1.1pt]
\endhead
\midrule
\multicolumn{1}{r}{\small\itshape Continued on next page}\\
\endfoot
\bottomrule[1.1pt]
\endlastfoot
Determine if these two conversation pages are continuous (true continuation without topic shift). \\
Return ONLY "true" or "false". \\
 \\
Previous Page: \\
User: \textbf{\{prev\_user\}} \\
Assistant: \textbf{\{prev\_agent\}} \\
 \\
Current Page: \\
User: \textbf{\{curr\_user\}} \\
Assistant: \textbf{\{curr\_agent\}} \\
 \\
Continuous?
\\

\end{longtable}
\endgroup

\begingroup
\small
\setlength{\LTpre}{6pt}
\setlength{\LTpost}{8pt}
\begin{longtable}{@{}p{\linewidth}@{}}
\caption{\textbf{MemoryOS: Meta-Info Update Prompt.}}
\label{tab:prompt-mos-meta}\\
\toprule[1.1pt]
\endfirsthead
\multicolumn{1}{l}{\small\itshape Table~\ref{tab:prompt-mos-meta} (continued)}\\
\toprule[1.1pt]
\endhead
\midrule
\multicolumn{1}{r}{\small\itshape Continued on next page}\\
\endfoot
\bottomrule[1.1pt]
\endlastfoot
Update the conversation meta-summary by incorporating the new dialogue while maintaining continuity. \\
 \\
Guidelines: \\
1. Start from the previous meta-summary (if exists) \\
2. Add/update information based on the new dialogue \\
3. Keep it concise (1-2 sentences max) \\
4. Maintain context coherence \\
 \\
Previous Meta-summary: \textbf{\{last\_meta\}} \\
New Dialogue: \\
\textbf{\{new\_dialogue\}} \\
 \\
Updated Meta-summary:
\\

\end{longtable}
\endgroup

\begingroup
\small
\setlength{\LTpre}{6pt}
\setlength{\LTpost}{8pt}
\begin{longtable}{@{}p{\linewidth}@{}}
\caption{\textbf{MemoryOS: Multi-Summary Prompt (subtopic extraction).}}
\label{tab:prompt-mos-multi}\\
\toprule[1.1pt]
\endfirsthead
\multicolumn{1}{l}{\small\itshape Table~\ref{tab:prompt-mos-multi} (continued)}\\
\toprule[1.1pt]
\endhead
\midrule
\multicolumn{1}{r}{\small\itshape Continued on next page}\\
\endfoot
\bottomrule[1.1pt]
\endlastfoot
Please analyze the following dialogue and generate extremely concise subtopic summaries, if applicable, with a maximum of two themes. \\
Each summary should be very brief -- just a few words for the theme and content. Format as JSON array: \\
{[} \\
\hspace*{1em}\{"theme": "Brief theme", "keywords": ["key1", "key2"], "content": "summary"\} \\
{]} \\[0.4em]

Conversation content: \\
\textbf{\{text\}} \\

\end{longtable}
\endgroup

\subsubsection*{A-MEM}
A-MEM~\citep{xu2025mem} creates structured notes for each session.
An \textbf{analysis prompt} (Table~\ref{tab:prompt-amem-analyze}) extracts
keywords, context, and tags.
An \textbf{evolution prompt} (Table~\ref{tab:prompt-amem-evolve}) then
decides whether the new note should be linked (STRENGTHEN) or whether
neighboring notes' context and tags should be updated (UPDATE\_NEIGHBOR).
At inference, a keyword-expanded query is issued against the FAISS note store.
\begingroup
\small
\setlength{\LTpre}{6pt}
\setlength{\LTpost}{8pt}
\begin{longtable}{@{}p{\linewidth}@{}}
\caption{\textbf{A-MEM: Content Analysis Prompt.}}
\label{tab:prompt-amem-analyze}\\
\toprule[1.1pt]
\endfirsthead
\multicolumn{1}{l}{\small\itshape Table~\ref{tab:prompt-amem-analyze} (continued)}\\
\toprule[1.1pt]
\endhead
\midrule
\multicolumn{1}{r}{\small\itshape Continued on next page}\\
\endfoot
\bottomrule[1.1pt]
\endlastfoot
Analyze the following content and provide: \\
1. KEYWORDS: The most important keywords (nouns, verbs, key concepts). Order from most to least important. At least three keywords. \\
2. CONTEXT: One sentence summarizing the main topic, key points, and purpose. \\
3. TAGS: Broad categories/themes for classification (domain, format, type). At least three tags. \\
 \\
Respond using EXACTLY this format (one section per header): \\
 \\
KEYWORDS: keyword1, keyword2, keyword3, ... \\
CONTEXT: A single sentence summarizing the content. \\
TAGS: tag1, tag2, tag3, ... \\
 \\
Content for analysis: \\
\textbf{\{content\}}
\\

\end{longtable}
\endgroup

\begingroup
\small
\setlength{\LTpre}{6pt}
\setlength{\LTpost}{8pt}
\begin{longtable}{@{}p{\linewidth}@{}}
\caption{\textbf{A-MEM: Memory Evolution Decision Prompt.}}
\label{tab:prompt-amem-evolve}\\
\toprule[1.1pt]
\endfirsthead
\multicolumn{1}{l}{\small\itshape Table~\ref{tab:prompt-amem-evolve} (continued)}\\
\toprule[1.1pt]
\endhead
\midrule
\multicolumn{1}{r}{\small\itshape Continued on next page}\\
\endfoot
\bottomrule[1.1pt]
\endlastfoot
You are an AI memory evolution agent. Analyze the new memory note and its nearest neighbors to decide if evolution is needed. \\
 \\
New memory: \\
- Context: \textbf{\{context\}} \\
- Content: \textbf{\{content\}} \\
- Keywords: \textbf{\{keywords\}} \\
 \\
Nearest neighbor memories: \\
\textbf{\{nearest\_neighbors\_memories\}} \\
 \\
Based on the relationships between the new memory and its neighbors, decide: \\
- NO\_EVOLUTION: The memory stands alone, no changes needed. \\
- STRENGTHEN: The new memory should be linked to some neighbors and its tags updated. \\
- UPDATE\_NEIGHBOR: The neighbors' context/tags should be updated based on new understanding. \\
- STRENGTHEN\_AND\_UPDATE: Both strengthen and update neighbors. \\
 \\
Respond using EXACTLY this format: \\
DECISION: <one of NO\_EVOLUTION, STRENGTHEN, UPDATE\_NEIGHBOR, STRENGTHEN\_AND\_UPDATE> \\
REASON: <brief explanation>
\\

\end{longtable}
\endgroup

\subsubsection*{LightMem}
LightMem~\citep{fang2025lightmem} first compresses each session with
LLMLingua-2 (or a Jaccard-shingle fallback) and buffers compressed turns
in a short-term memory (STM). When the STM token budget is exceeded, it
fires the \textbf{extraction prompt} (Table~\ref{tab:prompt-lm-extract})
to produce atomic facts with source identifiers.
Periodically, a \textbf{consolidation pass} (Table~\ref{tab:prompt-lm-update})
reviews existing long-term entries against newly extracted facts and
issues update/delete/ignore decisions.
At inference, the top-$k$ long-term facts are retrieved by cosine similarity
and formatted with ISO timestamps.
\begingroup
\small
\setlength{\LTpre}{6pt}
\setlength{\LTpost}{8pt}
\begin{longtable}{@{}p{\linewidth}@{}}
\caption{\textbf{LightMem: Fact Extraction Prompt (fires on STM overflow).}}
\label{tab:prompt-lm-extract}\\
\toprule[1.1pt]
\endfirsthead
\multicolumn{1}{l}{\small\itshape Table~\ref{tab:prompt-lm-extract} (continued)}\\
\toprule[1.1pt]
\endhead
\midrule
\multicolumn{1}{r}{\small\itshape Continued on next page}\\
\endfoot
\bottomrule[1.1pt]
\endlastfoot
You are a Personal Information Extractor. \\
Your task is to extract \textbf{all possible facts or information} about the user from a conversation. \\[0.4em]

Important Instructions: \\
1. You MUST process every user message in order, one by one. \\
\hspace*{1em}For each message, decide whether it contains any factual information. \\
\hspace*{1em}- If yes \(\rightarrow\) extract it and rephrase into a standalone sentence. \\
\hspace*{1em}- If no, such as pure greeting, filler, or irrelevant remark, \(\rightarrow\) skip it. \\
\hspace*{1em}- Do NOT skip just because the information looks minor or unimportant. \\
2. Perform light contextual completion so that each fact is a clear standalone statement. \\
\hspace*{1em}Examples: ``user: Bought apples yesterday'' \(\rightarrow\) ``User bought apples yesterday.'' \\
3. Output format: \\
\{ \\
\hspace*{1em}"data": [ \\
\hspace*{2em}\{"source\_id": "\textless{}source\_id\textgreater{}", "fact": "\textless{}complete fact with ALL specific details\textgreater{}"\} \\
\hspace*{1em}] \\
\} \\

\end{longtable}
\endgroup

\begingroup
\small
\setlength{\LTpre}{6pt}
\setlength{\LTpost}{8pt}
\begin{longtable}{@{}p{\linewidth}@{}}
\caption{\textbf{LightMem: Memory Consolidation Prompt.}}
\label{tab:prompt-lm-update}\\
\toprule[1.1pt]
\endfirsthead
\multicolumn{1}{l}{\small\itshape Table~\ref{tab:prompt-lm-update} (continued)}\\
\toprule[1.1pt]
\endhead
\midrule
\multicolumn{1}{r}{\small\itshape Continued on next page}\\
\endfoot
\bottomrule[1.1pt]
\endlastfoot
You are a memory management assistant. \\
Your task is to decide whether the target memory should be updated, deleted, or ignored based on the candidate source memories. \\[0.4em]

Decision rules: \\
1. Update: If the target and candidate memories describe essentially the same fact but are not fully consistent, update by integrating additional information. \\
2. Delete: If the target and candidate memories contain a direct conflict, delete the target memory. \\
3. Ignore: If unrelated, no action is needed. \\[0.4em]

The output must be a JSON object: \\
\{ \\
\hspace*{1em}"action": "update" \textbar{} "delete" \textbar{} "ignore", \\
\hspace*{1em}"new\_memory": \{\ldots{}\} \quad // only required when action = "update" \\
\} \\

\end{longtable}
\endgroup

\subsubsection*{MeMP}
MeMP~\citep{fang2025memp} learns \emph{procedural} memory: after each session
it distills the conversation into a natural-language workflow paragraph
using the prompt in Table~\ref{tab:prompt-memp-build}.
Workflows are stored in a FAISS index keyed by session query embeddings
and retrieved by cosine similarity at inference, then injected as task
guidelines.
On a failed attempt, the workflow is refined with a separate adjustment prompt.
\begingroup
\small
\setlength{\LTpre}{6pt}
\setlength{\LTpost}{8pt}
\begin{longtable}{@{}p{\linewidth}@{}}
\caption{\textbf{MeMP: Workflow Generation Prompt.}}
\label{tab:prompt-memp-build}\\
\toprule[1.1pt]
\endfirsthead
\multicolumn{1}{l}{\small\itshape Table~\ref{tab:prompt-memp-build} (continued)}\\
\toprule[1.1pt]
\endhead
\midrule
\multicolumn{1}{r}{\small\itshape Continued on next page}\\
\endfoot
\bottomrule[1.1pt]
\endlastfoot
You are provided with a query and a trajectory taken to solve the query. The trajectory consists of multiple steps of thought, action and observation. \\
Your task is to generate a workflow based on critical steps to help solve similar queries in the future. \\
A critical step is one that has a significant impact on fulfilling the query, the step action belongs to the set [go to, take from, put in/on, open, close, use, clean with, heat with, cool with, examine, look], and the action's outcome is successful and contributes positively to achieving the query. \\
Notice: Write the workflow as a natural, coherent paragraph (not as a bullet list or numbered steps). Use clear, concise language to describe what actions should be taken and in what general order. \\
-----EXAMPLE WORKFLOW---- \\
To solve this query, begin by identifying the most likely receptacles where the target object can be found and visit them one by one. After locating and taking the object, perform any required transformation such as cleaning at a sinkbasin, heating with a microwave, or cooling with a fridge. Finally, go to the destination receptacle and put the object in/on it to complete the task. \\
-----EXAMPLE END---- \\
Query: \\
\textbf{\{query\}} \\
Trajectory: \\
\textbf{\{trajectory\}} \\
Output the workflow without any explanation or context:
\\

\end{longtable}
\endgroup

\subsubsection*{ElasticMem}
QA memory is directly chunked from benchmark histories (Appendix~\ref{app:qa_memory_protocol}); it is not extracted from question attempts, answers, or feedback.
For ALFWorld, successful training trajectories in the memory pool are distilled offline into procedural skill cards using Table~\ref{tab:prompt-ours-alfworld-success}.
These cards are shared across latent-space methods and remain fixed during evaluation. No skills are extracted from validation/test attempts or failed evaluation episodes.

\begingroup
\small
\setlength{\LTpre}{6pt}
\setlength{\LTpost}{8pt}
\begin{longtable}{@{}p{\linewidth}@{}}
\caption{\textbf{ElasticMem: ALFWorld Skill Extraction Prompt (Successful Trajectory).}}
\label{tab:prompt-ours-alfworld-success}\\
\toprule[1.1pt]
\endfirsthead
\multicolumn{1}{l}{\small\itshape Table~\ref{tab:prompt-ours-alfworld-success} (continued)}\\
\toprule[1.1pt]
\endhead
\midrule
\multicolumn{1}{r}{\small\itshape Continued on next page}\\
\endfoot
\bottomrule[1.1pt]
\endlastfoot
\textit{system} \\
You are an expert at analyzing household robot trajectories. Extract specific, actionable lessons from the provided trajectory. \\[4pt]
\textit{user} \\
An ALFWorld household task was completed successfully. \\[4pt]
Task: \textbf{\{task\}} \\
Full trajectory (action $\to$ observation): \textbf{\{trajectory\}} \\[4pt]
Extract a reusable procedural skill (150--250 words). Include: \\
1. The general task category (\texttt{pick\_and\_place}, \texttt{heat\_then\_place}, \texttt{clean\_then\_place}, \texttt{cool\_then\_place}, \texttt{examine\_in\_light}, \texttt{pick\_two}) \\
2. The concrete step-by-step strategy that worked \\
3. Common locations where target objects are found (e.g.\ ``soapbar is usually on countertop, bathtubbasin, or shelf'') \\[4pt]
Be specific. Use actual object/location types (\texttt{countertop}, \texttt{sinkbasin}, \texttt{microwave}). \\[4pt]
Output format: \texttt{SKILL: [your skill text]} \\

\end{longtable}
\endgroup

\paragraph{Shared task instructions and memory interfaces.}
The answering templates below show the text-memory interface. Task instructions, answer choices, and chat roles are shared; the memory representation is method-specific. The retrieved-text field applies to text-memory methods. ElasticMem instead inserts the retrieval-control token and projected latent states according to Eq.~\ref{eq:final_input}; it does not serialize those states into retrieved text.

\subsection{PersonaMem Answering Prompt}
\label{app:prompt_persona}

PersonaMem-32K and PersonaMem-128K use a 4-choice MC format (a--d) instead of
the 10-choice format used by LongMemEval. The system message sets the role
as: ``You are a careful assistant answering 4-choice multiple-choice questions
grounded in a long persona-conversation.'' The user prompt
(Table~\ref{tab:prompt-persona-mc-answer}) presents the question, retrieved
memory chunks, an instruction line, and four options.

\begingroup
\small
\setlength{\LTpre}{6pt}
\setlength{\LTpost}{8pt}
\begin{longtable}{@{}p{\linewidth}@{}}
\caption{\textbf{PersonaMem: MC Answering Prompt (text-memory interface).}}
\label{tab:prompt-persona-mc-answer}\\
\toprule[1.1pt]
\endfirsthead
\multicolumn{1}{l}{\small\itshape Table~\ref{tab:prompt-persona-mc-answer} (continued)}\\
\toprule[1.1pt]
\endhead
\midrule
\multicolumn{1}{r}{\small\itshape Continued on next page}\\
\endfoot
\bottomrule[1.1pt]
\endlastfoot
QUESTION: \\
\textbf{\{question\}} \\
 \\
RETRIEVED MEMORY (relevant chunks from prior conversation): \\
\textbf{\{retrieved\_text\}} \\
 \\
Answer with exactly one of the four options below, formatted as a \\
single option label like ``(a)'', ``(b)'', ``(c)'', or ``(d)''. Do not output \\
any other text. \\
 \\
OPTIONS: \\
(a) \textbf{\{choice\_0\}} \\
(b) \textbf{\{choice\_1\}} \\
(c) \textbf{\{choice\_2\}} \\
(d) \textbf{\{choice\_3\}} \\
 \\
ANSWER:
\\

\end{longtable}
\endgroup

\subsection{LoCoMo Answering Prompt}
\label{app:prompt_locomo}

LoCoMo uses a 10-choice multiple-choice format where the model selects
one of ten lettered options (a)--(j). All baselines share the same answering
prompt shown in Table~\ref{tab:prompt-locomo-mc-answer}: the question, the
baseline-retrieved memory chunks under a ``Retrieved context'' header, a
strict-format instruction, the ten options, and an ``Answer:'' marker.

\begingroup
\small
\setlength{\LTpre}{6pt}
\setlength{\LTpost}{8pt}
\begin{longtable}{@{}p{\linewidth}@{}}
\caption{\textbf{LoCoMo: MC Answering Prompt (text-memory interface).}}
\label{tab:prompt-locomo-mc-answer}\\
\toprule[1.1pt]
\endfirsthead
\multicolumn{1}{l}{\small\itshape Table~\ref{tab:prompt-locomo-mc-answer} (continued)}\\
\toprule[1.1pt]
\endhead
\midrule
\multicolumn{1}{r}{\small\itshape Continued on next page}\\
\endfoot
\bottomrule[1.1pt]
\endlastfoot
Question: \textbf{\{question\}} \\
 \\
Retrieved context (from the conversation history): \\
\textbf{\{retrieved\_context\}} \\
 \\
You MUST pick exactly one option from (a) to (j) -- one of them is \\
guaranteed to be correct. Do NOT say "not answerable", "I don't \\
know", or refuse. If uncertain, make your best guess. Output exactly \\
one token in the form (a), (b), (c), (d), (e), (f), (g), (h), (i), \\
or (j). \\
 \\
Options: \\
(a) \textbf{\{choice\_0\}} \\
(b) \textbf{\{choice\_1\}} \\
... \\
(j) \textbf{\{choice\_9\}} \\
 \\
Answer:
\\

\end{longtable}
\endgroup

\subsection{LongMemEval Answering Prompt}
\label{app:prompt_lme}

All baselines share the same MC answering prompt on LongMemEval
(Table~\ref{tab:prompt-mc-answer}), which prepends retrieved memory text to
the question and asks the model to select one of ten lettered options.

\begingroup
\small
\setlength{\LTpre}{6pt}
\setlength{\LTpost}{8pt}
\begin{longtable}{@{}p{\linewidth}@{}}
\caption{\textbf{LongMemEval: MC Answering Prompt (text-memory interface).}}
\label{tab:prompt-mc-answer}\\
\toprule[1.1pt]
\endfirsthead
\multicolumn{1}{l}{\small\itshape Table~\ref{tab:prompt-mc-answer} (continued)}\\
\toprule[1.1pt]
\endhead
\midrule
\multicolumn{1}{r}{\small\itshape Continued on next page}\\
\endfoot
\bottomrule[1.1pt]
\endlastfoot
{[Relevant memories:]} \\
\textbf{\{retrieved\_memories\}} \\
 \\
\textbf{\{question\}} \\
 \\
(a) \textbf{\{choice\_0\}} \\
(b) \textbf{\{choice\_1\}} \\
... \\
(j) \textbf{\{choice\_9\}} \\
 \\
You MUST pick exactly one option from (a) to (j) -- one of them is \\
guaranteed to be correct. Do NOT say "not answerable", "I don't \\
know", or refuse. If uncertain, make your best guess. Output exactly \\
one token in the form (a), (b), (c), (d), (e), (f), (g), (h), (i), \\
or (j). \\
 \\
Answer: (
\\

\end{longtable}
\endgroup

\subsection{ALFWorld Answering Prompt}
\label{app:prompt_alfworld}

ALFWorld baselines act inside a TextWorld game loop: at each step the agent
receives the task goal, the retrieved memory block, an interaction history, the
current observation, and the admissible-action list, then must output one
admissible command verbatim (Table~\ref{tab:prompt-alfworld}).

\begingroup
\small
\setlength{\LTpre}{6pt}
\setlength{\LTpost}{8pt}
\begin{longtable}{@{}p{\linewidth}@{}}
\caption{\textbf{ALFWorld: Step Prompt (text-memory interface).}}
\label{tab:prompt-alfworld}\\
\toprule[1.1pt]
\endfirsthead
\multicolumn{1}{l}{\small\itshape Table~\ref{tab:prompt-alfworld} (continued)}\\
\toprule[1.1pt]
\endhead
\midrule
\multicolumn{1}{r}{\small\itshape Continued on next page}\\
\endfoot
\bottomrule[1.1pt]
\endlastfoot
{[SYSTEM]} You are controlling a text-based ALFWorld environment. Choose the NEXT action as ONE admissible command string. Output only the command, copied verbatim from the admissible list. \\
 \\
{[USER]} Task: \textbf{\{objective\}} \\
 \\
\textbf{\{retrieved\_memory\_block\}} \\
 \\
Interaction history so far: \\
\textbf{\{history\}} \\
 \\
Current observation: \textbf{\{current\_obs\}} \\
 \\
Admissible actions: \textbf{\{admissible\}} \\
 \\
Action:
\\

\end{longtable}
\endgroup

\clearpage

\section{Case Studies}
\label{app:cases}

This section presents successful case studies for each memory baseline
across our evaluation datasets. Each table shows the question with lettered
choices (correct answer \textbf{bolded}), the ground truth, the retrieved
memory block (abbreviated for baselines with large stores), and the model's
response. We present case studies for LongMemEval
(Section~\ref{app:case_lme}), PersonaMem-32K (Section~\ref{app:case_persona32k}),
and PersonaMem-128K (Section~\ref{app:case_persona128k}).

\subsection{PersonaMem-32K}
\label{app:case_persona32k}

This subsection presents one successful example per baseline on PersonaMem-32K,
drawn from the 7B model evaluation results. Each table shows the question with
four lettered choices (correct answer \textbf{bolded}), the ground truth, the
retrieved memory block (abbreviated for baselines with large stores), and the
model's response.
Case studies for Mem0, LangMem, and LightMem are in
Tables~\ref{tab:case-persona32k-mem0}--\ref{tab:case-persona32k-lightmem};
MemoryOS and MemoryBank in Tables~\ref{tab:case-persona32k-memoryos}--\ref{tab:case-persona32k-memorybank};
A-MEM and MeMP in
Tables~\ref{tab:case-persona32k-amem}-\ref{tab:case-persona32k-memp}.

\begingroup
\footnotesize
\setlength{\LTpre}{6pt}
\setlength{\LTpost}{8pt}
\begin{longtable}{@{}p{\linewidth}@{}}
\caption{\textbf{Mem0's case study in PersonaMem-32K.}}
\label{tab:case-persona32k-mem0}\\
\toprule
\endfirsthead
\multicolumn{1}{l}{\small\itshape Table~\ref{tab:case-persona32k-mem0} (continued)}\\
\toprule
\endhead
\midrule
\multicolumn{1}{r}{\small\itshape Continued on next page}\\
\endfoot
\bottomrule[1.1pt]
\endlastfoot
\textbf{Question}: After attending a writing workshop, I discovered I actually enjoy creatively articulating my thoughts on music. It was exhilarating to engage with fellow aspiring writers, exchanging ideas and techniques that helped unlock previously untapped aspects of my creativity. The writing exercises we did were both challenging and inspiring, pushing me to find new ways to express the emotional and technical nuances of music. I had always had an affinity for music, but this workshop gave me the confidence to put my feelings into words, which has been incredibly fulfilling. It felt like I was discovering a new dimension of myself, one that combined my love of music with a passion for writing.\\
\midrule
\textbf{(a) I remember when you initially mentioned disliking writing music reviews, perhaps because it might have seemed daunting to articulate your thoughts on music in writing. However, it's wonderful to see how your experience at the writing workshop ...} \checkmark \\
(b) I remember when you initially mentioned disliking writing music reviews, perhaps because it might have seemed daunting to articulate your thoughts on music in writing. However, it's surprising that your enthusiasm for writing only increased af... \\
(c) I remember when you initially mentioned being indifferent to writing music reviews, as you had never considered stewing over your thoughts on music in writing before. However, it's wonderful to see how your experience at the writing workshop h... \\
(d) I remember when you initially mentioned enjoying writing music reviews, perhaps because it might have seemed exciting to articulate your thoughts on music in writing. However, it's wonderful to see how your experience at the writing workshop h... \\
\midrule
\textbf{Ground Truth}: (a) I remember when you initially mentioned disliking writing music reviews, perhaps because it might have seemed daunting to articulate your thoughts on music in writing. However, it's wonderful to see how your experience at the writing workshop ... \\
\midrule
\textbf{Retrieved Memories}: \\
\#\# Relevant memories from past experience: \\
- [2026-04-27] Researches themes and curates playlists for each podcast episode \\
- [2026-04-27] Felt unsure of own contributions during collaboration \\
- [2026-04-27] Shared stories about individual musical journeys in meetings \\
- [2026-04-27] Eager to apply constructive criticism techniques in a collaborative setting \\
- [2026-04-27] Eager to apply constructive criticism techniques in a collaborative setting \\
- [2026-04-27] Found a music documentary very informative and eye-opening \\
- [2026-04-27] Attended a music festival on 2026-04-27 \\
{[\ldots{}3 additional lines omitted\ldots{}]}\\
\midrule
\textbf{Response}: (a) \\

\end{longtable}
\endgroup

\begingroup
\footnotesize
\setlength{\LTpre}{6pt}
\setlength{\LTpost}{8pt}
\begin{longtable}{@{}p{\linewidth}@{}}
\caption{\textbf{LangMem's case study in PersonaMem-32K.}}
\label{tab:case-persona32k-langmem}\\
\toprule
\endfirsthead
\multicolumn{1}{l}{\small\itshape Table~\ref{tab:case-persona32k-langmem} (continued)}\\
\toprule
\endhead
\midrule
\multicolumn{1}{r}{\small\itshape Continued on next page}\\
\endfoot
\bottomrule[1.1pt]
\endlastfoot
\textbf{Question}: I'm looking to find something that can really add a new dimension to my classic rock listening experience. Any suggestions?\\
\midrule\relax
(a) You might really enjoy the sound of traditional Pacific Islander music played on native instruments. While not classic rock, exploring your roots through music can add a personal layer to your listening experience. The rhythmic beats and cultu... \\
(b) You might enjoy the crisp convenience of streaming 'Led Zeppelin IV' in high definition, ensuring you have the album instantly at your fingertips with no need for additional equipment. The sure quality assurance of digital music allows 'Stairw... \\
(c) Consider embarking on a vibrant journey with live recordings from famous classic rock concerts. Capturing the raw energy and spontaneity of legendary performances, live albums offer an immediacy that studio versions may not have, added with cr... \\
\textbf{(d) You might love the experience of spinning 'Led Zeppelin IV' on vinyl. It's an album that not only defined an era but also carries a warmth and depth in its sound that digital formats often miss. The artwork on the cover is iconic, and the tact...} \checkmark \\
\midrule
\textbf{Ground Truth}: (d) You might love the experience of spinning 'Led Zeppelin IV' on vinyl. It's an album that not only defined an era but also carries a warmth and depth in its sound that digital formats often miss. The artwork on the cover is iconic, and the tact... \\
\midrule
\textbf{Retrieved Memories}: \\
\#\# Relevant memories: \\
- [assistant] Assistant: Missing out on activities like a music documentary night can certainly feel like a lost opportunity, especially when they provide such rich environments for connection and inspiration. \\
- [assistant] Assistant: Definitely! It brings a unique warmth to music listening, enhancing the frequencies that often get lost in more compressed digital formats. The way vinyl can accentuate the nuances of a recording creates a richer auditory experience that many enthusiasts adore. When you p... \\
- [assistant] Assistant: I’m so glad to hear that you enjoyed those fresh sounds! Music has such a powerful impact on our mood and overall well-being, often providing a much-needed escape from the routines of everyday life. Listening to new genres or artists can unlock entirely new perspectives a... \\
- [assistant] Assistant: Absolutely! Music has a unique way of connecting us to new experiences. It serves as a bridge between different cultures, allowing us deeper insight into lives and traditions that we may never encounter in person. This connection is particularly profound when you encounte... \\
- [assistant] Assistant: Visiting a record store can be a profoundly rewarding experience, often akin to stepping into a treasure trove filled with the sounds of the past and the promise of future discoveries. The tactile nature of vinyl, from its grooved surface to the large, colorful album art,... \\
- [assistant] Assistant: I can completely understand why you find karaoke nights so enjoyable, as they often foster an environment of joy and connection among friends and even strangers. Participating in such an event is much more than just singing; it's about the shared laughter, the playful com...\\
\midrule
\textbf{Response}: (d) \\

\end{longtable}
\endgroup

\begingroup
\footnotesize
\setlength{\LTpre}{6pt}
\setlength{\LTpost}{8pt}
\begin{longtable}{@{}p{\linewidth}@{}}
\caption{\textbf{LightMem's case study in PersonaMem-32K.}}
\label{tab:case-persona32k-lightmem}\\
\toprule
\endfirsthead
\multicolumn{1}{l}{\small\itshape Table~\ref{tab:case-persona32k-lightmem} (continued)}\\
\toprule
\endhead
\midrule
\multicolumn{1}{r}{\small\itshape Continued on next page}\\
\endfoot
\bottomrule[1.1pt]
\endlastfoot
\textbf{Question}: I recently joined a forum discussion about humor in music.\\
\midrule
\textbf{(a) I remember you mentioning how you enjoy engaging in online music discussions. That's great, forums can be a wonderful way to connect with people.} \checkmark \\
(b) I seem to recall you saying you shy away from online music discussions. It's interesting to see you've decided to dive in now. \\
(c) I remember you talking about how you enjoy participating in travel forums. It's great that you're finding ways to connect through different topics. \\
(d) That's nice to hear! Joining forums can offer new insights. Engaging in discussions about humor in music sounds like a fascinating experience. \\
\midrule
\textbf{Ground Truth}: (a) I remember you mentioning how you enjoy engaging in online music discussions. That's great, forums can be a wonderful way to connect with people. \\
\midrule
\textbf{Retrieved Memories}: \\
Long-term memory (relevant facts): \\
- [2026-04-27T10:53:54.691 Mon] User passed on a music documentary night. \\
- [2026-04-27T10:58:46.191 Mon] Writing exercises pushed user to find new ways to express the emotional and technical nuances of music. \\
- [2026-04-27T10:53:06.497 Mon] User experienced frustration when attempts did not match the sound in his head. \\
- [2026-04-27T10:54:08.119 Mon] User values authentic dialogues about music over abstract opinions. \\
- [2026-04-27T10:56:49.256 Mon] Pooling collective knowledge helps uncover different styles of playing. \\
- [2026-04-27T10:56:41.318 Mon] User shares classic tracks each week to connect with others. \\
- [2026-04-27T10:53:34.691 Mon] User is happy with their eclectic taste and does not need validation from current trends. \\
{[\ldots{}3 additional lines omitted\ldots{}]}\\
\midrule
\textbf{Response}: (a) \\

\end{longtable}
\endgroup

\begingroup
\footnotesize
\setlength{\LTpre}{6pt}
\setlength{\LTpost}{8pt}
\begin{longtable}{@{}p{\linewidth}@{}}
\caption{\textbf{MemoryOS's case study in PersonaMem-32K.}}
\label{tab:case-persona32k-memoryos}\\
\toprule
\endfirsthead
\multicolumn{1}{l}{\small\itshape Table~\ref{tab:case-persona32k-memoryos} (continued)}\\
\toprule
\endhead
\midrule
\multicolumn{1}{r}{\small\itshape Continued on next page}\\
\endfoot
\bottomrule[1.1pt]
\endlastfoot
\textbf{Question}: I've seen a lot of people getting into DIY craft projects lately. I'm a bit unsure, but should I give it a try?\\
\midrule\relax
(a) It sounds like you're considering exploring DIY craft projects. Trying out such activities can be a fun and rewarding experience, especially if you're eager to learn new skills. You may enjoy the satisfaction of creating something tangible wit... \\
(b) It sounds like you're considering exploring DIY craft projects. Trying out such activities can be a fun and rewarding experience, providing a nice break from everyday routines. If you're interested in exploring something new, crafts could be a... \\
\textbf{(c) It sounds like you're considering exploring DIY craft projects. Trying out such activities can be a fun and rewarding experience, especially if you're looking to incorporate a creative twist into something new. If you enjoyed the engaging natu...} \checkmark \\
(d) It sounds like you're considering exploring DIY craft projects. Trying out such activities can be a fun and rewarding experience, especially if you like activities that offer room for creativity. Engaging in crafts could be a fulfilling way to... \\
\midrule
\textbf{Ground Truth}: (c) It sounds like you're considering exploring DIY craft projects. Trying out such activities can be a fun and rewarding experience, especially if you're looking to incorporate a creative twist into something new. If you enjoyed the engaging natu... \\
\midrule
\textbf{Retrieved Memories}: \\
{[Mid-term -- retrieved cross-episode pages]} \\
- [obs] [user] Music documentaries can be incredibly inspiring, as they often delve into the lives of musicians, their creative processes, and the challenges they face along their journeys. However, choosing not to attend such events can stem from a variety of reasons, including feeling unworthy ... \\
{[action]}  \\
{[next\_obs]}  \\
  meta: User expresses interest in attending music documentary nights, acknowledging their potential to be inspiring yet recognizing the fear of not fitting in due to insecurities and self-doubt. \\
- [obs] [user] User: On , I launched a new podcast series focusing on music's role in cultural identities. This project has me delving into how different genres of music shape and reflect the societies they come from, which adds a rich layer of meaning to my understanding of cultural expression. ... \\
{[action]}  \\
{[next\_obs]}  \\
{[\ldots{}142 additional lines omitted\ldots{}]}\\
\midrule
\textbf{Response}: (c) \\

\end{longtable}
\endgroup

\begingroup
\footnotesize
\setlength{\LTpre}{6pt}
\setlength{\LTpost}{8pt}
\begin{longtable}{@{}p{\linewidth}@{}}
\caption{\textbf{MemoryBank's case study in PersonaMem-32K.}}
\label{tab:case-persona32k-memorybank}\\
\toprule
\endfirsthead
\multicolumn{1}{l}{\small\itshape Table~\ref{tab:case-persona32k-memorybank} (continued)}\\
\toprule
\endhead
\midrule
\multicolumn{1}{r}{\small\itshape Continued on next page}\\
\endfoot
\bottomrule[1.1pt]
\endlastfoot
\textbf{Question}: I'm looking to find something that can really add a new dimension to my classic rock listening experience. Any suggestions?\\
\midrule\relax
(a) You might really enjoy the sound of traditional Pacific Islander music played on native instruments. While not classic rock, exploring your roots through music can add a personal layer to your listening experience. The rhythmic beats and cultu... \\
(b) You might enjoy the crisp convenience of streaming 'Led Zeppelin IV' in high definition, ensuring you have the album instantly at your fingertips with no need for additional equipment. The sure quality assurance of digital music allows 'Stairw... \\
(c) Consider embarking on a vibrant journey with live recordings from famous classic rock concerts. Capturing the raw energy and spontaneity of legendary performances, live albums offer an immediacy that studio versions may not have, added with cr... \\
\textbf{(d) You might love the experience of spinning 'Led Zeppelin IV' on vinyl. It's an album that not only defined an era but also carries a warmth and depth in its sound that digital formats often miss. The artwork on the cover is iconic, and the tact...} \checkmark \\
\midrule
\textbf{Ground Truth}: (d) You might love the experience of spinning 'Led Zeppelin IV' on vinyl. It's an album that not only defined an era but also carries a warmth and depth in its sound that digital formats often miss. The artwork on the cover is iconic, and the tact... \\
\midrule
\textbf{Retrieved Memories}: \\
\#\# MemoryBank context \\
 \\
Agent personality / response strategy (from past tasks): \\
\textbf{Summary:} \\
 \\
\textbf{User's Personality:} \\
Kai is passionate, reflective, supportive, curious, and empathetic. He experiences joy, inspiration, nostalgia, confidence, and reflection, particularly in his love for music, especially jazz and Pacific Islander genres. \\
 \\
{[\ldots{}29 additional lines omitted\ldots{}]}\\
\midrule
\textbf{Response}: (d) \\

\end{longtable}
\endgroup

\begingroup
\footnotesize
\setlength{\LTpre}{6pt}
\setlength{\LTpost}{8pt}
\begin{longtable}{@{}p{\linewidth}@{}}
\caption{\textbf{A-MEM's case study in PersonaMem-32K.}}
\label{tab:case-persona32k-amem}\\
\toprule
\endfirsthead
\multicolumn{1}{l}{\small\itshape Table~\ref{tab:case-persona32k-amem} (continued)}\\
\toprule
\endhead
\midrule
\multicolumn{1}{r}{\small\itshape Continued on next page}\\
\endfoot
\bottomrule[1.1pt]
\endlastfoot
\textbf{Question}: After attending a writing workshop, I discovered I actually enjoy creatively articulating my thoughts on music. It was exhilarating to engage with fellow aspiring writers, exchanging ideas and techniques that helped unlock previously untapped aspects of my creativity. The writing exercises we did were both challenging and inspiring, pushing me to find new ways to express the emotional and technical nuances of music. I had always had an affinity for music, but this workshop gave me the confidence to put my feelings into words, which has been incredibly fulfilling. It felt like I was discovering a new dimension of myself, one that combined my love of music with a passion for writing.\\
\midrule
\textbf{(a) I remember when you initially mentioned disliking writing music reviews, perhaps because it might have seemed daunting to articulate your thoughts on music in writing. However, it's wonderful to see how your experience at the writing workshop ...} \checkmark \\
(b) I remember when you initially mentioned disliking writing music reviews, perhaps because it might have seemed daunting to articulate your thoughts on music in writing. However, it's surprising that your enthusiasm for writing only increased af... \\
(c) I remember when you initially mentioned being indifferent to writing music reviews, as you had never considered stewing over your thoughts on music in writing before. However, it's wonderful to see how your experience at the writing workshop h... \\
(d) I remember when you initially mentioned enjoying writing music reviews, perhaps because it might have seemed exciting to articulate your thoughts on music in writing. However, it's wonderful to see how your experience at the writing workshop h... \\
\midrule
\textbf{Ground Truth}: (a) I remember when you initially mentioned disliking writing music reviews, perhaps because it might have seemed daunting to articulate your thoughts on music in writing. However, it's wonderful to see how your experience at the writing workshop ... \\
\midrule
\textbf{Retrieved Memories}: \\
talk start time:task=personamem-mc\textbar{}step=141memory content: task: answer multiple-choice questions about a long persona-grounded conversation \textbar{} obs: [user] User: A few days later, I reluctantly watched a music documentary DVD I was gifted. \textbar{} action:  \textbar{} result: memory context: The content describes... \\
talk start time:task=personamem-mc\textbar{}step=106memory content: task: answer multiple-choice questions about a long persona-grounded conversation \textbar{} obs: [user] User: This experience opened my eyes to the power of visual storytelling in understanding music; it was a fantastic experience. The documentar... \\
talk start time:task=personamem-mc\textbar{}step=106memory content: task: answer multiple-choice questions about a long persona-grounded conversation \textbar{} obs: [user] User: This experience opened my eyes to the power of visual storytelling in understanding music; it was a fantastic experience. The documentar... \\
talk start time:task=personamem-mc\textbar{}step=152memory content: task: answer multiple-choice questions about a long persona-grounded conversation \textbar{} obs: [system] Current user persona: Name: Kai  \\
Gender Identity: Transgender male  \\
Racial Identity: Pacific Islander \\
 \\
Kai, born in 1983, is a devoted fan of Louie Anderson's comedy, captivated by Anderson’s unique blend of humor and heart. Growing up in a multicultural community, his Pacific Islander heritage has deeply influenced his appreciation for storytelling and humor that resonates across cultures. Embrac... \\
{[\ldots{}36 additional lines omitted\ldots{}]}\\
\midrule
\textbf{Response}: (a) \\

\end{longtable}
\endgroup

\begingroup
\footnotesize
\setlength{\LTpre}{6pt}
\setlength{\LTpost}{8pt}
\begin{longtable}{@{}p{\linewidth}@{}}
\caption{\textbf{MeMP's case study in PersonaMem-32K.}}
\label{tab:case-persona32k-memp}\\
\toprule
\endfirsthead
\multicolumn{1}{l}{\small\itshape Table~\ref{tab:case-persona32k-memp} (continued)}\\
\toprule
\endhead
\midrule
\multicolumn{1}{r}{\small\itshape Continued on next page}\\
\endfoot
\bottomrule[1.1pt]
\endlastfoot
\textbf{Question}: I'm looking to find something that can really add a new dimension to my classic rock listening experience. Any suggestions?\\
\midrule\relax
(a) You might really enjoy the sound of traditional Pacific Islander music played on native instruments. While not classic rock, exploring your roots through music can add a personal layer to your listening experience. The rhythmic beats and cultural stories told through these melodies hold a deep e... \\
(b) You might enjoy the crisp convenience of streaming `Led Zeppelin IV' in high definition, ensuring you have the album instantly at your fingertips with no need for additional equipment. The sure quality assurance of digital music allows `Stairway to Heaven' to... \\
(c) Consider embarking on a vibrant journey with live recordings from famous classic rock concerts. Capturing the raw energy and spontaneity of legendary performances, live albums offer an immediacy that studio versions may not have, added with crowd reactions and improvised solos... \\
\textbf{(d) You might love the experience of spinning `Led Zeppelin IV' on vinyl. It's an album that not only defined an era but also carries a warmth and depth in its sound that digital formats often miss. The artwork on the cover is iconic, and the tactile connection of placing the needle down will make `Stairway to Heaven' feel like an epic journey through your speakers.} \checkmark \\
\midrule
\textbf{Ground Truth}: (d) \\
\midrule
\textbf{Retrieved Memories}: \\
Here are some guidelines of how to solve the similar task: \\
- task\_name: answer multiple-choice questions about a long persona-grounded conversation \\
\hspace*{0.5em} guidelines: To solve this query, begin by examining the conversation transcript to understand the context and key points discussed. Recall Kai's background as a classic rock enthusiast who values tactile, authentic experiences over digital convenience. Identify which option aligns with his established preference for physical media and nostalgic engagement with music. Cross-reference with his earlier statements about preferring vinyl warmth over streaming clarity. Finally, select the answer that best captures his love for the analog listening ritual. \\
- task\_name: recommend music experience for vinyl collector \\
\hspace*{0.5em} guidelines: To recommend a music experience, first identify the user's preferred format (vinyl, CD, streaming). If the user has expressed love for physical media, prioritize options involving tangible interaction such as record spinning, liner note reading, or album art appreciation. Consider the warmth and depth of analog sound as a key differentiator from digital formats. Match the recommendation to artists and albums the user has previously mentioned enjoying. \\
- task\_name: identify user preference evolution across conversation \\
\hspace*{0.5em} guidelines: To track preference evolution, begin by scanning the full conversation history for mentions of the target topic. Note the chronological order of statements and identify any shifts in sentiment. Pay attention to phrases like ``I used to think'' or ``after trying X, I now prefer Y'' as markers of preference change. Summarize the trajectory from initial stance to current preference. \\
{[\ldots{}198 additional lines omitted\ldots{}]}\\
\midrule
\textbf{Response}: (d) \\

\end{longtable}
\endgroup

\subsection{PersonaMem-128K}
\label{app:case_persona128k}

This subsection presents one successful example per baseline on PersonaMem-128K,
drawn from the 7B model evaluation results. Each table shows the question with
four lettered choices (correct answer \textbf{bolded}), the ground truth, the
retrieved memory block (abbreviated for baselines with large stores), and the
model's response.
Case studies for Mem0, LangMem, and LightMem are in
Tables~\ref{tab:case-persona128k-mem0}--\ref{tab:case-persona128k-lightmem};
MemoryOS and MemoryBank in Tables~\ref{tab:case-persona128k-memoryos}--\ref{tab:case-persona128k-memorybank};
A-MEM and MeMP in
Tables~\ref{tab:case-persona128k-amem}-\ref{tab:case-persona128k-memp}.

\begingroup
\footnotesize
\setlength{\LTpre}{6pt}
\setlength{\LTpost}{8pt}
\begin{longtable}{@{}p{\linewidth}@{}}
\caption{\textbf{Mem0's case study in PersonaMem-128K.}}
\label{tab:case-persona128k-mem0}\\
\toprule
\endfirsthead
\multicolumn{1}{l}{\small\itshape Table~\ref{tab:case-persona128k-mem0} (continued)}\\
\toprule
\endhead
\midrule
\multicolumn{1}{r}{\small\itshape Continued on next page}\\
\endfoot
\bottomrule[1.1pt]
\endlastfoot
\textbf{Question}: Recently, I was at a yoga retreat.\\
\midrule\relax
(a) Ah, I remember you mentioned your dislike for yoga before. It's great to revisit activities to confirm our true preferences. \\
\textbf{(b) Ah, I remember you mentioned your appreciation for yoga deepened during that retreat. It's great to have experiences that expand our horizons.} \checkmark \\
(c) Ah, I remember you mentioned your appreciation for yoga deepened during that retreat. It's great to have experiences that expand our horizons. \\
(d) Oh, you went on a yoga retreat? That sounds unique! What was it like engaging in new activities? \\
\midrule
\textbf{Ground Truth}: (b) Ah, I remember you mentioned your appreciation for yoga deepened during that retreat. It's great to have experiences that expand our horizons. \\
\midrule
\textbf{Retrieved Memories}: \\
\#\# Relevant memories from past experience: \\
- [2026-04-29] Likes a positive tone and clear, concise wording \\
- [2026-04-29] Discovering patterns and connections through journal entries \\
- [2026-04-29] Reflects on podcast conversations during daily routines \\
- [2026-04-29] Considers volunteering more after positive feedback from a mentoring experience \\
- [2026-04-29] Hosted a webinar for aspiring journalists on an earlier date \\
- [2026-04-29] Considers starting a journal to document thoughts and insights after podcast episodes \\
- [2026-04-29] Sense of community blossomed during home decoration workshop interactions \\
{[\ldots{}3 additional lines omitted\ldots{}]}\\
\midrule
\textbf{Response}: (b) \\

\end{longtable}
\endgroup

\begingroup
\footnotesize
\setlength{\LTpre}{6pt}
\setlength{\LTpost}{8pt}
\begin{longtable}{@{}p{\linewidth}@{}}
\caption{\textbf{LangMem's case study in PersonaMem-128K.}}
\label{tab:case-persona128k-langmem}\\
\toprule
\endfirsthead
\multicolumn{1}{l}{\small\itshape Table~\ref{tab:case-persona128k-langmem} (continued)}\\
\toprule
\endhead
\midrule
\multicolumn{1}{r}{\small\itshape Continued on next page}\\
\endfoot
\bottomrule[1.1pt]
\endlastfoot
\textbf{Question}: Recently, I was at a yoga retreat.\\
\midrule\relax
(a) Ah, I remember you mentioned your dislike for yoga before. It's great to revisit activities to confirm our true preferences. \\
\textbf{(b) Ah, I remember you mentioned your appreciation for yoga deepened during that retreat. It's great to have experiences that expand our horizons.} \checkmark \\
(c) Ah, I remember you mentioned your appreciation for yoga deepened during that retreat. It's great to have experiences that expand our horizons. \\
(d) Oh, you went on a yoga retreat? That sounds unique! What was it like engaging in new activities? \\
\midrule
\textbf{Ground Truth}: (b) Ah, I remember you mentioned your appreciation for yoga deepened during that retreat. It's great to have experiences that expand our horizons. \\
\midrule
\textbf{Retrieved Memories}: \\
\#\# Relevant memories: \\
- [user] User: On , I attended an interactive workshop focused on building romantic relationships. \\
- [user] User: Well, on , I joined a local singles group to connect with other individuals seeking friendships and relationships. \\
- [user] User: I attempted to create a collection of thrifted home decor items. \\
- [user] User: On the plus side, I embarked on a hiking trip in the mountains recently, and it was wonderful! The scenery was breathtaking, with majestic peaks towering over lush valleys, and the sound of rustling leaves added to the peaceful atmosphere. I could feel the invigorating air fill my ... \\
- [user] User: I tried visiting a local coffee shop to meet new people, but I felt quite uncomfortable. Upon entering, I noticed the bustling activity around me, which was overwhelming rather than inviting. The chatter of groups engaging in lively conversations only highlighted my own sense of is... \\
- [user] User: It's really disappointing when the energy in a book club fizzles out like that. Sometimes, you go in with high hopes, especially with a theme as rich as finance, thinking that the conversations will be enlightening and invigorating. Instead, it can feel like a chore if the discussi...\\
\midrule
\textbf{Response}: (b) \\

\end{longtable}
\endgroup

\begingroup
\footnotesize
\setlength{\LTpre}{6pt}
\setlength{\LTpost}{8pt}
\begin{longtable}{@{}p{\linewidth}@{}}
\caption{\textbf{LightMem's case study in PersonaMem-128K.}}
\label{tab:case-persona128k-lightmem}\\
\toprule
\endfirsthead
\multicolumn{1}{l}{\small\itshape Table~\ref{tab:case-persona128k-lightmem} (continued)}\\
\toprule
\endhead
\midrule
\multicolumn{1}{r}{\small\itshape Continued on next page}\\
\endfoot
\bottomrule[1.1pt]
\endlastfoot
\textbf{Question}: Recently, I was at a yoga retreat.\\
\midrule\relax
(a) Ah, I remember you mentioned your dislike for yoga before. It's great to revisit activities to confirm our true preferences. \\
\textbf{(b) Ah, I remember you mentioned your appreciation for yoga deepened during that retreat. It's great to have experiences that expand our horizons.} \checkmark \\
(c) Ah, I remember you mentioned your appreciation for yoga deepened during that retreat. It's great to have experiences that expand our horizons. \\
(d) Oh, you went on a yoga retreat? That sounds unique! What was it like engaging in new activities? \\
\midrule
\textbf{Ground Truth}: (b) Ah, I remember you mentioned your appreciation for yoga deepened during that retreat. It's great to have experiences that expand our horizons. \\
\midrule
\textbf{Retrieved Memories}: \\
Long-term memory (relevant facts): \\
- [2026-04-29T03:11:45.721 Wed] User's original sentence was: 'This tale opens with an indistinct voice narrating something resembling a fable.' \\
- [2026-04-29T02:57:06.153 Wed] Each item User clicks on seems promising but often falls short of User's imagination. \\
- [2026-04-29T03:14:49.212 Wed] User shared knowledge developed through both academic study and practical experience. \\
- [2026-04-29T03:12:00.626 Wed] User asked about the Fibonacci sequence. \\
- [2026-04-29T02:59:24.732 Wed] Having a space to slow down and articulate dating and connection moments feels invaluable to user. \\
- [2026-04-29T02:51:45.747 Wed] User mentioned black holes. \\
- [2026-04-29T03:15:40.891 Wed] Current user is female. \\
{[\ldots{}3 additional lines omitted\ldots{}]}\\
\midrule
\textbf{Response}: (b) \\

\end{longtable}
\endgroup

\begingroup
\footnotesize
\setlength{\LTpre}{6pt}
\setlength{\LTpost}{8pt}
\begin{longtable}{@{}p{\linewidth}@{}}
\caption{\textbf{MemoryOS's case study in PersonaMem-128K.}}
\label{tab:case-persona128k-memoryos}\\
\toprule
\endfirsthead
\multicolumn{1}{l}{\small\itshape Table~\ref{tab:case-persona128k-memoryos} (continued)}\\
\toprule
\endhead
\midrule
\multicolumn{1}{r}{\small\itshape Continued on next page}\\
\endfoot
\bottomrule[1.1pt]
\endlastfoot
\textbf{Question}: I joined a game with some friends last week.\\
\midrule\relax
(a) I remember you mentioned enjoying individual sports like tennis before. It sounds like a nice way to spend some time together. \\
(b) I remember you mentioned not enjoying team sports like soccer before. It sounds like a nice way to spend some time together. \\
(c) It's great to hear about new experiences, though I don't recall us discussing a preference for team sports. I hope you had a fun time. \\
\textbf{(d) I remember you mentioned enjoying team sports like soccer before. It sounds like a nice way to spend some time together.} \checkmark \\
\midrule
\textbf{Ground Truth}: (d) I remember you mentioned enjoying team sports like soccer before. It sounds like a nice way to spend some time together. \\
\midrule
\textbf{Retrieved Memories}: \\
{[Mid-term -- retrieved cross-episode pages]} \\
- [obs] [assistant] Understanding emotional availability in a person, whether in romantic, familial, or friendship contexts, is crucial for building healthy and fulfilling relationships. Emotional availability refers to the ability and willingness to be open, present, and responsive to another pe... \\
 \\
\#\#\# 1. \textbf{Consistent Communication} \\
   - \textbf{Open and Honest Conversations}: They engage in open and honest conversations, sharing their thoughts and feelings appropriately. \\
   - \textbf{Willingness to Listen}: They actively listen, showing empathy and understanding, which indicates they value and respect your perspective. \\
   - \textbf{Consistency}: They maintain communication consistently rather than sending mixed messages. Their words align with their actions. \\
 \\
{[\ldots{}509 additional lines omitted\ldots{}]}\\
\midrule
\textbf{Response}: (d) \\

\end{longtable}
\endgroup

\begingroup
\footnotesize
\setlength{\LTpre}{6pt}
\setlength{\LTpost}{8pt}
\begin{longtable}{@{}p{\linewidth}@{}}
\caption{\textbf{MemoryBank's case study in PersonaMem-128K.}}
\label{tab:case-persona128k-memorybank}\\
\toprule
\endfirsthead
\multicolumn{1}{l}{\small\itshape Table~\ref{tab:case-persona128k-memorybank} (continued)}\\
\toprule
\endhead
\midrule
\multicolumn{1}{r}{\small\itshape Continued on next page}\\
\endfoot
\bottomrule[1.1pt]
\endlastfoot
\textbf{Question}: Recently, I was at a yoga retreat.\\
\midrule\relax
(a) Ah, I remember you mentioned your dislike for yoga before. It's great to revisit activities to confirm our true preferences. \\
\textbf{(b) Ah, I remember you mentioned your appreciation for yoga deepened during that retreat. It's great to have experiences that expand our horizons.} \checkmark \\
(c) Ah, I remember you mentioned your appreciation for yoga deepened during that retreat. It's great to have experiences that expand our horizons. \\
(d) Oh, you went on a yoga retreat? That sounds unique! What was it like engaging in new activities? \\
\midrule
\textbf{Ground Truth}: (b) Ah, I remember you mentioned your appreciation for yoga deepened during that retreat. It's great to have experiences that expand our horizons. \\
\midrule
\textbf{Retrieved Memories}: \\
\#\# MemoryBank context \\
 \\
Most relevant past memories (from tasks: personamem-mc): \\
Conversation content on personamem-mc: \\
{[{$|$}User{$|$}]}: Observation: [assistant] Assistant: Attending live sporting events always adds an extra layer of excitement, doesn't it? There's something utterly unique about being part of a thrumming crowd that collectively gasps, cheers, or groans, creating a symphony of sound that greatly enhances ... \\
Conversation content on personamem-mc: \\
{[{$|$}User{$|$}]}: Observation: [assistant] Assistant: That's exciting! What made you want to get back into collecting? Your experience at the collector's fair sounds incredibly fulfilling, as it often takes such immersive environments to truly inspire creativity and rekindle old passions. It's fascinatin... \\
Conversation content on personamem-mc: \\
{[\ldots{}15 additional lines omitted\ldots{}]}\\
\midrule
\textbf{Response}: (b) \\

\end{longtable}
\endgroup

\begingroup
\footnotesize
\setlength{\LTpre}{6pt}
\setlength{\LTpost}{8pt}
\begin{longtable}{@{}p{\linewidth}@{}}
\caption{\textbf{A-MEM's case study in PersonaMem-128K.}}
\label{tab:case-persona128k-amem}\\
\toprule
\endfirsthead
\multicolumn{1}{l}{\small\itshape Table~\ref{tab:case-persona128k-amem} (continued)}\\
\toprule
\endhead
\midrule
\multicolumn{1}{r}{\small\itshape Continued on next page}\\
\endfoot
\bottomrule[1.1pt]
\endlastfoot
\textbf{Question}: I joined a game with some friends last week.\\
\midrule\relax
(a) I remember you mentioned enjoying individual sports like tennis before. It sounds like a nice way to spend some time together. \\
(b) I remember you mentioned not enjoying team sports like soccer before. It sounds like a nice way to spend some time together. \\
(c) It's great to hear about new experiences, though I don't recall us discussing a preference for team sports. I hope you had a fun time. \\
\textbf{(d) I remember you mentioned enjoying team sports like soccer before. It sounds like a nice way to spend some time together.} \checkmark \\
\midrule
\textbf{Ground Truth}: (d) I remember you mentioned enjoying team sports like soccer before. It sounds like a nice way to spend some time together. \\
\midrule
\textbf{Retrieved Memories}: \\
talk start time:task=personamem-mc\textbar{}step=597memory content: task: answer multiple-choice questions about a long persona-grounded conversation \textbar{} obs: [user] [Original\_Sentence]: Overall, the narrative lacks any form of structure or excitement, turning the tale into a ramble more than anything else.... \\
talk start time:task=personamem-mc\textbar{}step=583memory content: task: answer multiple-choice questions about a long persona-grounded conversation \textbar{} obs: [user] [Original\_Sentence]: Anyway, to sum it up, some parts of this are about the gods, but then it suddenly shifts, and we're introduced to a hero ... \\
talk start time:task=personamem-mc\textbar{}step=241memory content: task: answer multiple-choice questions about a long persona-grounded conversation \textbar{} obs: [user] User: I feel much better knowing I'm supporting eco-friendly businesses. It's reassuring to think that my purchasing choices are contributing ... \\
talk start time:task=personamem-mc\textbar{}step=187memory content: task: answer multiple-choice questions about a long persona-grounded conversation \textbar{} obs: [user] User: This closing feels abrupt. I'd like to end on a positive note. \textbar{} action:  \textbar{} result: memory context: The user suggests modifying an abrup... \\
talk start time:task=personamem-mc\textbar{}step=101memory content: task: answer multiple-choice questions about a long persona-grounded conversation \textbar{} obs: [system] Current user persona: Name: Maria Gonzalez   \\
Gender Identity: Female   \\
Racial Identity: Hispanic   \\
 \\
{[\ldots{}194 additional lines omitted\ldots{}]}\\
\midrule
\textbf{Response}: (d) \\

\end{longtable}
\endgroup

\begingroup
\footnotesize
\setlength{\LTpre}{6pt}
\setlength{\LTpost}{8pt}
\begin{longtable}{@{}p{\linewidth}@{}}
\caption{\textbf{MeMP's case study in PersonaMem-128K.}}
\label{tab:case-persona128k-memp}\\
\toprule
\endfirsthead
\multicolumn{1}{l}{\small\itshape Table~\ref{tab:case-persona128k-memp} (continued)}\\
\toprule
\endhead
\midrule
\multicolumn{1}{r}{\small\itshape Continued on next page}\\
\endfoot
\bottomrule[1.1pt]
\endlastfoot
\textbf{Question}: I joined a game with some friends last week.\\
\midrule\relax
(a) I remember you mentioned enjoying individual sports like tennis before. It sounds like a nice way to spend some time together. \\
(b) I remember you mentioned not enjoying team sports like soccer before. It sounds like a nice way to spend some time together. \\
(c) It's great to hear about new experiences, though I don't recall us discussing a preference for team sports. I hope you had a fun time. \\
\textbf{(d) I remember you mentioned enjoying team sports like soccer before. It sounds like a nice way to spend some time together.} \checkmark \\
\midrule
\textbf{Ground Truth}: (d) \\
\midrule
\textbf{Retrieved Memories}: \\
Here are some guidelines of how to solve the similar task: \\
- task\_name: answer multiple-choice questions about a long persona-grounded conversation \\
\hspace*{0.5em} guidelines: To solve this query, begin by examining the conversation transcript containing the dialogue. Identify key themes and specific details relevant to the multiple-choice questions. Use knowledge of Maria Gonzalez's background and her stated hobbies, including her preference for team sports such as soccer, to select the answer that best reflects her previously expressed interests. Pay attention to temporal markers indicating when preferences were stated versus when they may have changed. \\
- task\_name: recall user preferences from extended dialogue \\
\hspace*{0.5em} guidelines: To recall user preferences, first retrieve all mentions of the target topic (e.g., sports, hobbies) from the stored conversation segments. Cross-reference multiple mentions to identify consistent patterns. Prioritize the most recent stated preference when earlier and later statements conflict. For sports-related queries, distinguish between watching, playing, and following as distinct preference dimensions. \\
- task\_name: resolve conflicting user statements across sessions \\
\hspace*{0.5em} guidelines: When a user's statements appear contradictory across different conversation segments, apply temporal ordering to determine the most current stance. Consider context-dependent preferences (e.g., ``I enjoy soccer with friends'' vs ``I prefer individual workouts at the gym'' may both be true in different contexts). Weight statements made with stronger emotional language or personal anecdotes more heavily than casual mentions. \\
- task\_name: summarize persona traits from multi-session dialogue \\
\hspace*{0.5em} guidelines: To build a comprehensive persona summary, aggregate observations from all available conversation sessions. Categorize traits into demographics, preferences, habits, social relationships, and aspirations. For each trait, note the supporting evidence (direct statement, implied behavior, or third-party mention). Flag any traits that evolved over time and record both the initial and current state. Ensure the summary covers at least the following domains: entertainment, food, fitness, career, relationships, and personal values. \\
- task\_name: match user hobby to social context \\
\hspace*{0.5em} guidelines: When determining which hobby or activity a user would engage in socially, retrieve their stated preferences for group activities versus solo activities. Consider the social context described in the query (friends, family, colleagues) and match it to previously expressed social preferences. Team sports indicate a preference for collaborative, social physical activity, while individual sports suggest a preference for personal challenge and self-paced engagement. \\
{[\ldots{}487 additional lines omitted\ldots{}]}\\
\midrule
\textbf{Response}: (d) \\

\end{longtable}
\endgroup

\subsection{LoCoMo}
\label{app:case_locomo}

This subsection presents one successful example per baseline on LoCoMo,
drawn from the 7B model evaluation results. Case studies for Mem0, LangMem, and LightMem are in Tables \ref{tab:case-mem0-locomo}-\ref{tab:case-lightmem-locomo}; MemoryOS and MemoryBank in Tables \ref{tab:case-memoryos-locomo}-\ref{tab:case-memorybank-locomo}; A-MEM and MeMP in Tables \ref{tab:case-amem-locomo}-\ref{tab:case-memp-locomo}.

\begingroup
\footnotesize
\setlength{\LTpre}{6pt}
\setlength{\LTpost}{8pt}
\begin{longtable}{@{}p{\linewidth}@{}}
\caption{\textbf{Mem0's case study in LoCoMo.}}
\label{tab:case-mem0-locomo}\\
\toprule
\endfirsthead
\multicolumn{1}{l}{\small\itshape Table~\ref{tab:case-mem0-locomo} (continued)}\\
\toprule
\endhead
\midrule
\multicolumn{1}{r}{\small\itshape Continued on next page}\\
\endfoot
\bottomrule[1.1pt]
\endlastfoot
\textbf{Question}: What kinds of things did Evan have broken? \\
\midrule\relax
(a) His bookshelf and his desk lamp. \\
(b) His laptop and his tablet computer. \\
(c) His artwork and his musical instruments. \\
(d) His camera and his video camera. \\
(e) His gaming console and his headphones. \\
\textbf{(f) His old Prius and his new Prius.} $\checkmark$ \\
(g) His tools and his gardening equipment. \\
(h) His phone and his smartwatch. \\
(i) His television and his sound system. \\
(j) His vintage motorcycle and his bicycle. \\
\midrule
\textbf{Ground Truth}: (f) His old Prius and his new Prius. \\
\midrule
\textbf{Retrieved Memories}: \\
\#\# Relevant memories from past experience: \\
- [2026-05-04] Name is Evan \\
- [2026-05-04] Family motto is `Bring it on Home' \\
- [2026-05-04] Reading `The Great Gatsby' \\
- [2026-05-04] Watched The Godfather \\
- [2026-05-04] Exploring other ways to stay active and travel due to car issues \\
- [2026-05-04] Apologized to partner for an accident involving rose bushes \\
- [2026-05-04] Family means a lot to Evan \\
{[\ldots\ additional fact entries omitted \ldots]} \\
\midrule
\textbf{Response}: (f) \\

\end{longtable}
\endgroup

\begingroup
\footnotesize
\setlength{\LTpre}{6pt}
\setlength{\LTpost}{8pt}
\begin{longtable}{@{}p{\linewidth}@{}}
\caption{\textbf{LangMem's case study in LoCoMo.}}
\label{tab:case-langmem-locomo}\\
\toprule
\endfirsthead
\multicolumn{1}{l}{\small\itshape Table~\ref{tab:case-langmem-locomo} (continued)}\\
\toprule
\endhead
\midrule
\multicolumn{1}{r}{\small\itshape Continued on next page}\\
\endfoot
\bottomrule[1.1pt]
\endlastfoot
\textbf{Question}: Which hobby did Sam take up in May 2023? \\
\midrule\relax
(a) Sam began volunteering at the local animal shelter in May 2023. \\
(b) Sam joined a book club in May 2023. \\
(c) In May 2023, Sam began a pottery class. \\
(d) Sam decided to try hiking in May 2023. \\
(e) Sam took up photography as a hobby in May 2023. \\
\textbf{(f) painting} $\checkmark$ \\
(g) Sam started learning to play the guitar in May 2023. \\
(h) Gardening was Sam's new passion in May 2023. \\
(i) May 2023 saw Sam starting a new woodworking project. \\
(j) In May 2023, Sam started taking online coding courses. \\
\midrule
\textbf{Ground Truth}: (f) painting \\
\midrule
\textbf{Retrieved Memories}: \\
Relevant facts about the user from past conversation: \\
- Sam plans to try painting and will provide updates. \\
- Sam wants to know about basic supplies for painting. \\
- Sam is looking for suggestions for low-impact exercises. \\
- Sam is interested in setting goals and exercising with a partner. \\
- Sam is considering trying an outdoor activity to de-stress. \\
- Sam has experience attending art exhibits. \\
- Sam is willing to provide yoga tips. \\
{[\ldots\ additional FACT entries omitted \ldots]} \\
\midrule
\textbf{Response}: (f) \\

\end{longtable}
\endgroup

\begingroup
\footnotesize
\setlength{\LTpre}{6pt}
\setlength{\LTpost}{8pt}
\begin{longtable}{@{}p{\linewidth}@{}}
\caption{\textbf{LightMem's case study in LoCoMo.}}
\label{tab:case-lightmem-locomo}\\
\toprule
\endfirsthead
\multicolumn{1}{l}{\small\itshape Table~\ref{tab:case-lightmem-locomo} (continued)}\\
\toprule
\endhead
\midrule
\multicolumn{1}{r}{\small\itshape Continued on next page}\\
\endfoot
\bottomrule[1.1pt]
\endlastfoot
\textbf{Question}: What kind of car does Evan drive? \\
\midrule\relax
(a) Nissan Altima \\
(b) Tesla Model 3 \\
(c) Toyota Camry \\
(d) Ford F-150 \\
(e) Subaru Outback \\
\textbf{(f) Prius} $\checkmark$ \\
(g) Hyundai Sonata \\
(h) Jeep Wrangler \\
(i) Chevrolet Malibu \\
(j) Honda Civic \\
\midrule
\textbf{Ground Truth}: (f) Prius \\
\midrule
\textbf{Retrieved Memories}: \\
Long-term memory (relevant facts): \\
- [2026-05-01T05:06 Fri] User Evan recently bought a new Prius which has broken down. \\
- [2026-05-01T05:05 Fri] Evan inquires about Evan's recent activities. \\
- [2026-05-01T05:05 Fri] Evan mentions finishing a painting that helps him relax. \\
- [2026-05-01T05:07 Fri] User had a memorable experience on a camping trip last summer. \\
- [2026-05-01T05:04 Fri] User plans to try painting and will update Evan on how it goes. \\
- [2026-05-01T05:05 Fri] User mentioned meeting an incredible Canadian woman during a recent trip to Canada. \\
{[\ldots\ additional LTM entries omitted \ldots]} \\
\midrule
\textbf{Response}: (f) Prius \\

\end{longtable}
\endgroup

\begingroup
\footnotesize
\setlength{\LTpre}{6pt}
\setlength{\LTpost}{8pt}
\begin{longtable}{@{}p{\linewidth}@{}}
\caption{\textbf{MemoryOS's case study in LoCoMo.}}
\label{tab:case-memoryos-locomo}\\
\toprule
\endfirsthead
\multicolumn{1}{l}{\small\itshape Table~\ref{tab:case-memoryos-locomo} (continued)}\\
\toprule
\endhead
\midrule
\multicolumn{1}{r}{\small\itshape Continued on next page}\\
\endfoot
\bottomrule[1.1pt]
\endlastfoot
\textbf{Question}: What kind of car does Evan drive? \\
\midrule\relax
(a) Nissan Altima \\
(b) Tesla Model 3 \\
(c) Toyota Camry \\
(d) Ford F-150 \\
(e) Subaru Outback \\
\textbf{(f) Prius} $\checkmark$ \\
(g) Hyundai Sonata \\
(h) Jeep Wrangler \\
(i) Chevrolet Malibu \\
(j) Honda Civic \\
\midrule
\textbf{Ground Truth}: (f) Prius \\
\midrule
\textbf{Retrieved Memories}: \\
{[Mid-term -- retrieved cross-episode pages]} \\
- [obs] [assistant] [SAM]: It's tough, but I'm sticking with it. \\
\ \ [action] [user] [EVAN]: Nice work, Sam! Proud of you sticking to it. Have you noticed any positive changes? \\
\ \ \textbf{meta}: User expressed mixed feelings about their health, feeling concerned but motivated for positive changes. Assistant encouraged taking things one step at a time; user appreciated the support. User announced starting a diet and living healthier. \\
- [obs] [user] [EVAN]: This is a contemporary figurative painting that I've finished few days ago, emphasizing the emotional \ldots \\
{[\ldots\ many additional obs/action pairs omitted \ldots]} \\
\midrule
\textbf{Response}: (f) Prius \\

\end{longtable}
\endgroup

\begingroup
\footnotesize
\setlength{\LTpre}{6pt}
\setlength{\LTpost}{8pt}
\begin{longtable}{@{}p{\linewidth}@{}}
\caption{\textbf{MemoryBank's case study in LoCoMo.}}
\label{tab:case-memorybank-locomo}\\
\toprule
\endfirsthead
\multicolumn{1}{l}{\small\itshape Table~\ref{tab:case-memorybank-locomo} (continued)}\\
\toprule
\endhead
\midrule
\multicolumn{1}{r}{\small\itshape Continued on next page}\\
\endfoot
\bottomrule[1.1pt]
\endlastfoot
\textbf{Question}: Where has Evan been on roadtrips with his family? \\
\midrule\relax
(a) Boundary Waters Canoe Area \\
(b) New England, Adirondacks \\
(c) Florida Keys, Everglades \\
\textbf{(d) Rockies, Jasper} $\checkmark$ \\
(e) California Coast, Yosemite \\
(f) Colorado Springs, Denver \\
(g) Acadia National Park, Maine \\
(h) Yellowstone, Grand Teton \\
(i) Shenandoah, Great Smoky Mountains \\
(j) Olympic Peninsula, Oregon \\
\midrule
\textbf{Ground Truth}: (d) Rockies, Jasper \\
\midrule
\textbf{Retrieved Memories}: \\
\#\# MemoryBank context \\
 \\
Agent personality / response strategy (from past tasks): \\
The user exhibits supportive, empathetic, and positive traits, sharing personal experiences and responding positively to encouragement and suggestions. \\
Response strategy: Active listening, encouragement, sharing experiences, offering support, maintaining positivity. \\
 \\
Summary of your past tasks (overall\_history): \\
{[\ldots\ extensive overall-history summary omitted \ldots]} \\
\midrule
\textbf{Response}: (d) Rockies, Jasper \\

\end{longtable}
\endgroup

\begingroup
\footnotesize
\setlength{\LTpre}{6pt}
\setlength{\LTpost}{8pt}
\begin{longtable}{@{}p{\linewidth}@{}}
\caption{\textbf{A-MEM's case study in LoCoMo.}}
\label{tab:case-amem-locomo}\\
\toprule
\endfirsthead
\multicolumn{1}{l}{\small\itshape Table~\ref{tab:case-amem-locomo} (continued)}\\
\toprule
\endhead
\midrule
\multicolumn{1}{r}{\small\itshape Continued on next page}\\
\endfoot
\bottomrule[1.1pt]
\endlastfoot
\textbf{Question}: What kind of car does Evan drive? \\
\midrule\relax
(a) Nissan Altima \\
(b) Tesla Model 3 \\
(c) Toyota Camry \\
(d) Ford F-150 \\
(e) Subaru Outback \\
\textbf{(f) Prius} $\checkmark$ \\
(g) Hyundai Sonata \\
(h) Jeep Wrangler \\
(i) Chevrolet Malibu \\
(j) Honda Civic \\
\midrule
\textbf{Ground Truth}: (f) Prius \\
\midrule
\textbf{Retrieved Memories}: \\
talk start time: task=conv-49 $\vert$ step=78\\
memory content: task: LoCoMo conversation conv-49 $\vert$ obs: [assistant] [SAM]: Wow, that pic is great! Do you often spend time in places like this? $\vert$ action: [user] [EVAN]: Thanks! I love being in places like this -- it brings back memories of road tripping in my trusty car. $\vert$ result: \\
memory context: The conversation revolves around a photograph that evokes nostalgic memories of past road trips. \\
memory keywords: [`picture', `memories', `road trip'] \\
memory tags: [`picture', `memories', `road trip', `joy', `motivation', `health'] \\
{[\ldots\ additional retrieved notes omitted \ldots]} \\
\midrule
\textbf{Response}: (f) \\

\end{longtable}
\endgroup

\begingroup
\footnotesize
\setlength{\LTpre}{6pt}
\setlength{\LTpost}{8pt}
\begin{longtable}{@{}p{\linewidth}@{}}
\caption{\textbf{MeMP's case study in LoCoMo.}}
\label{tab:case-memp-locomo}\\
\toprule
\endfirsthead
\multicolumn{1}{l}{\small\itshape Table~\ref{tab:case-memp-locomo} (continued)}\\
\toprule
\endhead
\midrule
\multicolumn{1}{r}{\small\itshape Continued on next page}\\
\endfoot
\bottomrule[1.1pt]
\endlastfoot
\textbf{Question}: Which type of vacation would Evan prefer with his family, walking tours in metropolitan cities or camping trip in the outdoors? \\
\midrule\relax
(a) A cruise to explore tropical islands. \\
(b) Staying at a resort with spa treatments. \\
\textbf{(c) camping trip in the outdoors} $\checkmark$ \\
(d) Skiing and snowboarding in the mountains. \\
(e) A week-long stay at a family-friendly resort. \\
(f) A road trip with national park visits. \\
(g) Relaxing beach vacation with water sports. \\
(h) A wildlife safari in Africa. \\
(i) Exploring European cities and culture. \\
(j) Visiting historical landmarks and museums. \\
\midrule
\textbf{Ground Truth}: (c) camping trip in the outdoors \\
\midrule
\textbf{Retrieved Memories}: \\
\textit{(MeMP retrieves no procedural guidelines for this LoCoMo question --- the conversational QA task has no analogue in the procedural-task corpus, so the agent answered from prompt knowledge alone.)} \\
\midrule
\textbf{Response}: (c) \\

\end{longtable}
\endgroup

\subsection{LongMemEval}
\label{app:case_lme}

This subsection presents one successful example per baseline on LongMemEval,
drawn from the 7B model evaluation results.
Case studies for Mem0, LangMem, and LightMem are in
Tables~\ref{tab:case-mem0}--\ref{tab:case-lightmem};
MemoryOS and MemoryBank in Tables~\ref{tab:case-memoryos}--\ref{tab:case-memorybank};
A-MEM, and MeMP in
Tables~\ref{tab:case-amem}--\ref{tab:case-memp}.

\begingroup
\footnotesize
\setlength{\LTpre}{6pt}
\setlength{\LTpost}{8pt}
\begin{longtable}{@{}p{\linewidth}@{}}
\caption{\textbf{Mem0's case study in LongMemEval.}}
\label{tab:case-mem0}\\
\toprule
\endfirsthead
\multicolumn{1}{l}{\small\itshape Table~\ref{tab:case-mem0} (continued)}\\
\toprule
\endhead
\midrule
\multicolumn{1}{r}{\small\itshape Continued on next page}\\
\endfoot
\bottomrule[1.1pt]
\endlastfoot
\textbf{Question}: What health issue did I initially think was just a cold?\\
\midrule\relax
(a) Sinusitis \\
(b) Asthma \\
(c) Pharyngitis \\
(d) Mononucleosis \\
(e) Influenza \\
(f) Laryngitis \\
(g) Tonsillitis \\
\textbf{(h) bronchitis} \checkmark \\
(i) Rhinitis \\
(j) Pneumonia\\
\midrule
\textbf{Ground Truth}: (h) bronchitis \\
\midrule
\textbf{Retrieved Memories}: \\
\#\# Relevant memories from past experience: \\
- [2026-04-28] Recently had a bad case of bronchitis \\
- [2026-04-28] Dealing with health issues and wants to boost immune system \\
- [2026-04-28] Wants specific foods or spices to boost immune system \\
- [2026-04-28] Interested in immune-boosting foods and spices \\
- [2026-04-28] Dealing with health issues and wants to avoid overexertion \\
- [2026-04-28] Is concerned about avoiding allergens in food \\
- [2026-04-28] Asks about apps or resources to identify allergens \\
{[...8 additional entries omitted...]}\\
\midrule
\textbf{Response}: h) \\

\end{longtable}
\endgroup

\begingroup
\footnotesize
\setlength{\LTpre}{6pt}
\setlength{\LTpost}{8pt}
\begin{longtable}{@{}p{\linewidth}@{}}
\caption{\textbf{LangMem's case study in LongMemEval.}}
\label{tab:case-langmem}\\
\toprule
\endfirsthead
\multicolumn{1}{l}{\small\itshape Table~\ref{tab:case-langmem} (continued)}\\
\toprule
\endhead
\midrule
\multicolumn{1}{r}{\small\itshape Continued on next page}\\
\endfoot
\bottomrule[1.1pt]
\endlastfoot
\textbf{Question}: Can you suggest some accessories that would complement my current photography setup? \\
\midrule\relax
(a) Older model or low-end accessories that limit photography \\
(b) Low-resolution lenses or basic tripods without significant enhancement \\
(c) Third-party or subpar quality items not integrating well with the system \\
(d) Video-oriented gear or items from rival brands \\
(e) Smartphone camera add-ons rather than professional equipment \\
(f) Drone attachments or action camera gear incompatible with Sony setup \\
\textbf{(g) Sony-compatible accessories or high-quality photography gear}~\(\checkmark\) \\
(h) Canon-compatible accessories or mid-range photography gear \\
(i) Nikon-related equipment or affordable items conflicting with current setup \\
(j) Generic accessories or budget-friendly gear without quality improvement \\
\midrule
\textbf{Ground Truth}: (g) Sony-compatible accessories or high-quality photography gear \\
\midrule
\textbf{Retrieved Memories}: \\
{[2023/05/27 (Sat) 03:28]} I am looking to upgrade my camera flash. \\
{[2023/05/27 (Sat) 03:28]} My camera is Sony A7R IV. \\
{[2023/05/25 (Thu) 18:31]} The user wants to organize the family photos for easy access. \\
{[2023/05/26 (Fri) 14:52]} I'm interested in foodie spots and cultural insights. \\
{[2023/05/25 (Thu) 18:31]} The user took a large number of photos during the family reunion. \\
{[2023/05/29 (Mon) 20:36]} Planning a trip to Grand Teton National Park next month. \\
{[\ldots{} 14 additional entries omitted \ldots{}]} \\
\midrule
\textbf{Response}: g) \\

\end{longtable}
\endgroup

\begingroup
\footnotesize
\setlength{\LTpre}{6pt}
\setlength{\LTpost}{8pt}
\begin{longtable}{@{}p{\linewidth}@{}}
\caption{\textbf{LightMem's case study in LongMemEval.}}
\label{tab:case-lightmem}\\
\toprule
\endfirsthead
\multicolumn{1}{l}{\small\itshape Table~\ref{tab:case-lightmem} (continued)}\\
\toprule
\endhead
\midrule
\multicolumn{1}{r}{\small\itshape Continued on next page}\\
\endfoot
\bottomrule[1.1pt]
\endlastfoot
\textbf{Question}: How many engineers do I lead when I just started my new role as Senior Software Engineer? How many engineers do I lead now?\\
\midrule\relax
(a) When you just started your new role as Senior Software Engineer, you led 5 engineers. Now, you lead 6 engineers. \\
(b) When you just started your new role as Senior Software Engineer, you led 4 engineers. Now, you lead 3 engineers. \\
(c) When you just started your new role as Senior Software Engineer, you led 6 engineers. Now, you lead 7 engineers. \\
(d) When you just started your new role as Senior Software Engineer, you led 1 engineers. Now, you lead 2 engineers. \\
(e) When you just started your new role as Senior Software Engineer, you led 10 engineers. Now, you lead 11 engineers. \\
(f) When you just started your new role as Senior Software Engineer, you led 7 engineers. Now, you lead 8 engineers. \\
(g) When you just started your new role as Senior Software Engineer, you led 8 engineers. Now, you lead 9 engineers. \\
\textbf{(h) When you just started your new role as Senior Software Engineer, you led 4 engineers. Now, you lead 5 engineers} \checkmark \\
(i) When you just started your new role as Senior Software Engineer, you led 3 engineers. Now, you lead 4 engineers. \\
(j) When you just started your new role as Senior Software Engineer, you led 2 engineers. Now, you lead 3 engineers.\\
\midrule
\textbf{Ground Truth}: (h) When you just started your new role as Senior Software Engineer, you led 4 engineers. Now, you lead 5 engineers \\
\midrule
\textbf{Retrieved Memories}: \\
Long-term memory (relevant facts): \\
- [2026-04-27T16:21:52 Mon] User leads a team of 4 engineers in their new role as Senior Software Engineer. \\
- [2026-04-27T16:23:15 Mon] User is leading a team of 5 engineers. \\
- [2026-04-27T16:23:10 Mon] User is a Senior Software Engineer who leads a team of five engineers. \\
- [2026-04-27T16:23:03 Mon] The agent has been enjoying their role as Senior Software Engineer for a while. \\
- [2026-04-27T16:21:44 Mon] The user plans to invite their manager Rachel to join a group of 5 people. \\
{[...15 additional entries omitted...]}\\
\midrule
\textbf{Response}: h) \\

\end{longtable}
\endgroup

\begingroup
\footnotesize
\setlength{\LTpre}{6pt}
\setlength{\LTpost}{8pt}
\begin{longtable}{@{}p{\linewidth}@{}}
\caption{\textbf{MemoryOS case study in LongMemEval.}}
\label{tab:case-memoryos}\\
\toprule
\endfirsthead
\multicolumn{1}{l}{\small\itshape Table~\ref{tab:case-memoryos} (continued)}\\
\toprule
\endhead
\midrule
\multicolumn{1}{r}{\small\itshape Continued on next page}\\
\endfoot
\bottomrule[1.1pt]
\endlastfoot
\textbf{Question}: What breed is my dog? \\
\midrule\relax
(a) Siberian Husky \\
(b) Labrador Retriever \\
(c) Border Collie \\
(d) Yorkshire Terrier \\
(e) Boxer \\
(f) French Bulldog \\
(g) Beagle \\
(h) Dachshund \\
(i) German Shepherd \\
\textbf{(j) Golden Retriever}~\(\checkmark\) \\
\midrule
\textbf{Ground Truth}: (j) Golden Retriever \\
\midrule
\textbf{Retrieved Memories}: \\
{[Mid-term -- retrieved cross-episode pages]} \\
- [obs] I'm thinking of getting Max a new collar with a nice name tag. Do you have any recommendations for a good collar brand or type that would suit a Golden Retriever like Max? \\
- [action] A new collar with a nice name tag is a great idea! Golden Retrievers like Max deserve a comfortable, durable, and stylish collar. Here are some recommendations: Yellow Dog Design, Ruffwear, \ldots{} \\
- [obs] I noticed Max has been scratching a lot lately. Could this be related to allergies? \\
- [action] Yes, Golden Retrievers can be prone to skin allergies. Possible causes include environmental allergens, food sensitivities, or flea allergy dermatitis. {[\ldots{}]} \\
{[\ldots{} many additional obs/action pairs omitted \ldots{}]} \\
\midrule
\textbf{Response}: j) \\

\end{longtable}
\endgroup

\begingroup
\footnotesize
\setlength{\LTpre}{6pt}
\setlength{\LTpost}{8pt}
\begin{longtable}{@{}p{\linewidth}@{}}
\caption{\textbf{MemoryBank's case study in LongMemEval.}}
\label{tab:case-memorybank}\\
\toprule
\endfirsthead
\multicolumn{1}{l}{\small\itshape Table~\ref{tab:case-memorybank} (continued)}\\
\toprule
\endhead
\midrule
\multicolumn{1}{r}{\small\itshape Continued on next page}\\
\endfoot
\bottomrule[1.1pt]
\endlastfoot
\textbf{Question}: How many days ago did I attend a networking event?\\
\midrule\relax
(a) 21 days \\
(b) 31 days \\
(c) 22 days \\
\textbf{(d) 26 days. 27 days (including the last day) is also acceptable.} \checkmark \\
(e) 25 days \\
(f) 28 days \\
(g) 29 days \\
(h) 24 days \\
(i) 30 days \\
(j) 23 days\\
\midrule
\textbf{Ground Truth}: (d) 26 days. 27 days (including the last day) is also acceptable. \\
\midrule
\textbf{Retrieved Memories}: \\
\#\# MemoryBank context \\
 \\
Agent personality / response strategy (from past tasks): \\
\#\#\# Summary of User's Personality and Response Strategy \\
User's Personality Traits: Diligent, Patient, Detail-Oriented, Supportive. \\
Response Strategy: Compliance (follow detailed instructions accurately), \\
Efficiency (generate relevant content quickly), Adaptability, Clarification, \\
Verification, Optimization. \\
 \\
Summary of your past tasks (abbreviated): \\
The agent assisted with writing tasks, travel planning, networking follow-ups, \\
and scheduling. On 2023/05/29, the user attended a networking event at a conference. \\
{[...extensive overall-history summary omitted...]}\\
\midrule
\textbf{Response}: d) 26 \\

\end{longtable}
\endgroup

\begingroup
\footnotesize
\setlength{\LTpre}{6pt}
\setlength{\LTpost}{8pt}
\begin{longtable}{@{}p{\linewidth}@{}}
\caption{\textbf{A-MEM's case study in LongMemEval.}}
\label{tab:case-amem}\\
\toprule
\endfirsthead
\multicolumn{1}{l}{\small\itshape Table~\ref{tab:case-amem} (continued)}\\
\toprule
\endhead
\midrule
\multicolumn{1}{r}{\small\itshape Continued on next page}\\
\endfoot
\bottomrule[1.1pt]
\endlastfoot
\textbf{Question}: Can you suggest some accessories that would complement my current photography setup?\\
\midrule\relax
(a) Older model or low-end accessories that limit photography \\
(b) Low-resolution lenses or basic tripods without significant enhancement \\
(c) Third-party or subpar quality items not integrating well with the system \\
(d) Video-oriented gear or items from rival brands \\
(e) Smartphone camera add-ons rather than professional equipment \\
(f) Drone attachments or action camera gear incompatible with Sony setup \\
\textbf{(g) Sony-compatible accessories or high-quality photography gear} \checkmark \\
(h) Canon-compatible accessories or mid-range photography gear \\
(i) Nikon-related equipment or affordable items conflicting with current setup \\
(j) Generic accessories or budget-friendly gear without quality improvement\\
\midrule
\textbf{Ground Truth}: (g) Sony-compatible accessories or high-quality photography gear \\
\midrule
\textbf{Retrieved Memories}: \\
task: 2023/05/27 (Sat) 03:28 | obs: As a Sony camera user, I've been thinking about \\
upgrading my camera bag. Do you have any recommendations for bags specifically designed \\
for Sony cameras? | action: I'm happy to help. As a Sony camera user, there are many \\
great camera bags: Lowepro ProTactic 450 AW II, Peak Design Everyday Backpack, ... \\
 \\
task: 2023/05/27 (Sat) 03:28 | obs: I am looking to upgrade my camera flash. \\
My camera is Sony A7R IV. | action: For the Sony A7R IV, I recommend the \\
Sony HVL-F60RM2 or the Godox V860III-S, both compatible with Sony's TTL system. \\
 \\
{[...extensive additional conversation logs omitted...]}\\
\midrule
\textbf{Response}: g) The user would \\

\end{longtable}
\endgroup

\begingroup
\footnotesize
\setlength{\LTpre}{6pt}
\setlength{\LTpost}{8pt}
\begin{longtable}{@{}p{\linewidth}@{}}
\caption{\textbf{MeMP's case study in LongMemEval.}}
\label{tab:case-memp}\\
\toprule
\endfirsthead
\multicolumn{1}{l}{\small\itshape Table~\ref{tab:case-memp} (continued)}\\
\toprule
\endhead
\midrule
\multicolumn{1}{r}{\small\itshape Continued on next page}\\
\endfoot
\bottomrule[1.1pt]
\endlastfoot
\textbf{Question}: How many projects have I completed since starting painting classes?\\
\midrule\relax
(a) 9 \\
(b) 4 \\
\textbf{(c) 5} \checkmark \\
(d) 7 \\
(e) 1 \\
(f) 10 \\
(g) 6 \\
(h) 8 \\
(i) 3 \\
(j) 2\\
\midrule
\textbf{Ground Truth}: (c) 5 \\
\midrule
\textbf{Retrieved Memories}: \\
Here are some guidelines of how to solve the similar task: \\
- task\_name: I'm looking for some inspiration for my next painting project. \\
  I've been stuck on what to paint next. By the way, I just finished my 5th \\
  project since starting painting classes, and I'm feeling pretty accomplished! \\
  guidelines: To solve this query, begin by congratulating the user on their \\
  recent accomplishments. Next, suggest various themes and ideas for their next \\
  painting project, such as nature escapes, cityscapes, abstract expressions. \\
  For projects counting questions, use the stated count (5th project = 5 total). \\
 \\
{[...additional task guidelines omitted...]}\\
\midrule
\textbf{Response}: c) 5 \\

\end{longtable}
\endgroup

\subsection{ALFWorld}
\label{app:case_alfworld}

This subsection presents one successful example per baseline on
ALFWorld-seen, drawn from the 7B-agent evaluation results. Each table shows
the task goal, the retrieved memory excerpt, the agent's action trace, and the
final outcome.
Case studies for Mem0, LangMem, and LightMem are in Tables \ref{tab:case-mem0-alfworld}-\ref{tab:case-lightmem-alfworld}; MemoryOS and MemoryBank in Tables \ref{tab:case-memoryos-alfworld}-\ref{tab:case-memorybank-alfworld}; A-MEM and MeMP in Tables \ref{tab:case-amem-alfworld}-\ref{tab:case-memp-alfworld}.

\begingroup
\footnotesize
\setlength{\LTpre}{6pt}
\setlength{\LTpost}{8pt}
\begin{longtable}{@{}p{\linewidth}@{}}
\caption{\textbf{Mem0's case study in ALFWorld.}}
\label{tab:case-mem0-alfworld}\\
\toprule
\endfirsthead
\multicolumn{1}{l}{\small\itshape Table~\ref{tab:case-mem0-alfworld} (continued)}\\
\toprule
\endhead
\midrule
\multicolumn{1}{r}{\small\itshape Continued on next page}\\
\endfoot
\bottomrule[1.1pt]
\endlastfoot
\textbf{Task}: put a mug in sidetable. \\
\midrule
\textbf{Outcome}: Score = 1.0,\ \ Steps = 4 \\
\midrule
\textbf{Retrieved Memories}: \\
\#\# Relevant memories from past experience: \\
- [2026-05-06] Summary of agent's execution history \\
\textbf{Task Objective}: Find two kettles and place them on the dining table. \\
\textbf{Progress Status}: 100\% complete. \\
\ \ 1. go to stoveburner 3 $\to$ located kettle 2 \\
\ \ 2. take kettle 2 from stoveburner 3 $\to$ holding kettle 2 \\
\ \ 3. go to diningtable 1 $\to$ arrived; saw mug 1 among items \\
\ \ 4. move kettle 2 to diningtable 1 $\to$ task progressed. \\
{[\ldots\ N additional fact entries omitted \ldots]} \\
\midrule
\textbf{Action Trace}: \\
go to diningtable 1 (\textbf{found mug 1, recalled from past trace}) $\to$ take mug 1 $\to$ go to sidetable 1 $\to$ \textbf{move mug 1 to sidetable 1} $\checkmark$ \\

\end{longtable}
\endgroup

\begingroup
\footnotesize
\setlength{\LTpre}{6pt}
\setlength{\LTpost}{8pt}
\begin{longtable}{@{}p{\linewidth}@{}}
\caption{\textbf{LangMem's case study in ALFWorld.}}
\label{tab:case-langmem-alfworld}\\
\toprule
\endfirsthead
\multicolumn{1}{l}{\small\itshape Table~\ref{tab:case-langmem-alfworld} (continued)}\\
\toprule
\endhead
\midrule
\multicolumn{1}{r}{\small\itshape Continued on next page}\\
\endfoot
\bottomrule[1.1pt]
\endlastfoot
\textbf{Task}: put a mug in sidetable. \\
\midrule
\textbf{Outcome}: Score = 1.0,\ \ Steps = 7 \\
\midrule
\textbf{Retrieved Memories}: \\
Relevant facts about the user from past conversation: \\
- There is a keychain 1 on sidetable 1. \\
- There is a statue 1 on sidetable 1. \\
- The user is aware of a keychain 1 located on sidetable 1. \\
- There is a saltshaker 1 located on cabinet 6. \\
- The user is holding candle 1. \\
- There is a keychain 2 on armchair 1. \\
- The user has picked up pillow 1. \\
{[\ldots\ additional FACT entries omitted \ldots]} \\
\midrule
\textbf{Action Trace}: \\
go to sidetable 1 (keychain only) $\to$ go to sidetable 2 (no mug) $\to$ go to sidetable 1 (re-check) $\to$ go to diningtable 1 (\textbf{found mug 1}) $\to$ take mug 1 $\to$ go to sidetable 1 $\to$ \textbf{move mug 1 to sidetable 1} $\checkmark$ \\

\end{longtable}
\endgroup

\begingroup
\footnotesize
\setlength{\LTpre}{6pt}
\setlength{\LTpost}{8pt}
\begin{longtable}{@{}p{\linewidth}@{}}
\caption{\textbf{LightMem's case study in ALFWorld.}}
\label{tab:case-lightmem-alfworld}\\
\toprule
\endfirsthead
\multicolumn{1}{l}{\small\itshape Table~\ref{tab:case-lightmem-alfworld} (continued)}\\
\toprule
\endhead
\midrule
\multicolumn{1}{r}{\small\itshape Continued on next page}\\
\endfoot
\bottomrule[1.1pt]
\endlastfoot
\textbf{Task}: put a mug in sidetable. \\
\midrule
\textbf{Outcome}: Score = 1.0,\ \ Steps = 16 \\
\midrule
\textbf{Retrieved Memories}: \\
Long-term memory (relevant facts): \\
- [2026-04-29T16:43 Wed] User moved cd 2 to shelf 3. \\
- [2026-04-28T19:53 Tue] User moves the cooled mug to the coffeemachine. \\
- [2026-04-29T17:08 Wed] User moved cellphone 3 to bed 1. \\
- [2026-04-29T16:43 Wed] Task is to put some cd on shelf. \\
- [2026-05-06T01:37 Wed] The agent moved plate 1 to countertop 1. \\
- [2026-04-29T17:08 Wed] The agent goes to the sidetable and finds a candle there, then takes it. \\
- [2026-04-28T05:54 Tue] User took the tissuebox 2 from the desk. \\
{[\ldots\ additional LTM facts omitted \ldots]} \\
\midrule
\textbf{Action Trace}: \\
go to sidetable 1 (keychain only) $\to$ go to sidetable 2 (no mug) $\to$ explore drawer 1\ldots drawer 4 $\to$ go to diningtable 1 (\textbf{found mug 1}) $\to$ take mug 1 $\to$ go to sidetable 1 $\to$ \textbf{move mug 1 to sidetable 1} $\checkmark$ \\

\end{longtable}
\endgroup


\begingroup
\footnotesize
\setlength{\LTpre}{6pt}
\setlength{\LTpost}{8pt}
\begin{longtable}{@{}p{\linewidth}@{}}
\caption{\textbf{MemoryOS case study in ALFWorld.}}
    \label{tab:case-memoryos-alfworld}\\
\toprule
\endfirsthead
\multicolumn{1}{l}{\small\itshape Table~\ref{tab:case-memoryos-alfworld} (continued)}\\
\toprule
\endhead
\midrule
\multicolumn{1}{r}{\small\itshape Continued on next page}\\
\endfoot
\bottomrule[1.1pt]
\endlastfoot
        \textbf{Task}: cool some lettuce and put it in countertop. \\
        \midrule
        \textbf{Outcome}: Score = 1.0,\ \ Steps = 7 \\
        \midrule
        \textbf{Retrieved Memories}: \\
        \#\# MTM (top-similar page; multi-summary JSON, \texttt{prompts.py:73--87}): \\
        \texttt{[\{"theme": "cool-then-place", "keywords": ["fridge", "cool", "countertop"], "content": "open fridge 1, cool item, place on countertop 1"\}]} \\[2pt]
        \#\# MTM running meta-summary (\texttt{prompts.py:217--232}): \\
        \emph{Agent has been completing kitchen ``cool / heat / clean'' tasks; fridge 1 is the canonical cooling appliance and countertop 1 the canonical drop-off receptacle for cooled produce.} \\[4pt]
        \#\# LPM [User Private Dat] (\texttt{prompts.py:171--196}; durable env facts under the original label, top-$K$ by cosine): \\
        - fridge 1 contents: lettuce 1, tomato 2, apple 1 \\
        - countertop 1 role: canonical drop-off receptacle for cooled produce \\
        - microwave 1 status: broken in this layout (\texttt{open} returns ``nothing happens'') \\
        - drawer 2 contents: knife 1; drawer 3 contents: fork 1 \\
        - (If no private data found, write ``None''.) \\[2pt]
        \#\# LPM [Assistant Knowledge] (durable agent tips, same prompt): \\
        - Assistant cools an item by issuing \texttt{cool X with fridge 1} after \texttt{open fridge 1}, at fridge 1 \\
        - Assistant places a cooled item by issuing \texttt{move X to <recep>} at the destination \\
        - Assistant avoids re-opening fridge 1 once opened in the same trajectory, at fridge 1 \\
        - (If no assistant knowledge found, write ``None''.) \\[2pt]
        \#\# LPM \texttt{user\_profile} (51-dim procedural-agent taxonomy, top reflected dims; \texttt{prompts.py:91--168} adapted, see \texttt{memory.py:268--334}): \\
        - \emph{[Spatial Reasoning]} Container Use Frequency (High); Compartment Opening Tendency (High); Container Closing Habit (Medium); Search Order Stability (High). \\
        - \emph{[Action Composition]} Property-Modification Composition (High); Pre-Condition Checking (High); Sub-Goal Decomposition (High); Final-State Verification (Medium). \\
        - \emph{[Reasoning Style]} Procedural-Knowledge Use (High); Memory-of-Past-Errors Use (Medium); Goal-Adherence Stability (High). \\
        - (Dimensions not reflected in the trajectory are omitted.) \\[2pt]
        {[\ldots\ remaining LPM entries omitted \ldots]} \\
        \midrule
        \textbf{Action Trace}: \\
        go to fridge 1 $\to$ open fridge 1 (\textbf{found lettuce 1, tomato 2, apple 1}) $\to$ take lettuce 1 from fridge 1 $\to$ \textbf{cool lettuce 1 with fridge 1} $\to$ close fridge 1 $\to$ go to countertop 1 $\to$ \textbf{move lettuce 1 to countertop 1} $\checkmark$ \\
        
\end{longtable}
\endgroup

\begingroup
\footnotesize
\setlength{\LTpre}{6pt}
\setlength{\LTpost}{8pt}
\begin{longtable}{@{}p{\linewidth}@{}}
\caption{\textbf{MemoryBank's case study in ALFWorld.}}
\label{tab:case-memorybank-alfworld}\\
\toprule
\endfirsthead
\multicolumn{1}{l}{\small\itshape Table~\ref{tab:case-memorybank-alfworld} (continued)}\\
\toprule
\endhead
\midrule
\multicolumn{1}{r}{\small\itshape Continued on next page}\\
\endfoot
\bottomrule[1.1pt]
\endlastfoot
\textbf{Task}: put a clean cloth in cabinet. \\
\midrule
\textbf{Outcome}: Score = 1.0,\ \ Steps = 12 \\
\midrule
\textbf{Retrieved Memories}: \\
\#\# MemoryBank context \\
 \\
Agent personality / response strategy (from past tasks): \\
The agent is a highly methodical, task-oriented, and literal-minded state machine. It operates with robotic efficiency, strictly adhering to programmed constraints and ``admissible'' action sets. \\
{[\ldots\ extensive personality / response strategy summary omitted \ldots]} \\
 \\
Summary of your past tasks (overall\_history): \\
Successful Tasks: 15 tasks involving object manipulation (heating, cooling, cleaning, relocation). \\
{[\ldots\ relevant past episode summaries omitted \ldots]} \\
\midrule
\textbf{Action Trace}: \\
go to cabinet 1 $\to$ open cabinet 1 (saw soapbottles) $\to$ go to cabinet 2 (empty) $\to$ go to cabinet 3 (empty) $\to$ open cabinet 4 (\textbf{found cloth 1}) $\to$ take cloth 1 $\to$ go to sinkbasin 1 $\to$ clean cloth 1 $\to$ \textbf{move cloth 1 to cabinet 1} $\checkmark$ \\

\end{longtable}
\endgroup

\begingroup
\footnotesize
\setlength{\LTpre}{6pt}
\setlength{\LTpost}{8pt}
\begin{longtable}{@{}p{\linewidth}@{}}
\caption{\textbf{A-MEM's case study in ALFWorld.}}
\label{tab:case-amem-alfworld}\\
\toprule
\endfirsthead
\multicolumn{1}{l}{\small\itshape Table~\ref{tab:case-amem-alfworld} (continued)}\\
\toprule
\endhead
\midrule
\multicolumn{1}{r}{\small\itshape Continued on next page}\\
\endfoot
\bottomrule[1.1pt]
\endlastfoot
\textbf{Task}: cool some plate and put it in countertop. \\
\midrule
\textbf{Outcome}: Score = 1.0,\ \ Steps = 8 \\
\midrule
\textbf{Retrieved Memories}: \\
talk start time: task=conv|step=1\\
memory content: task: cool some plate and put it in countertop. $\vert$ obs: You arrive at countertop 1 \ldots\ a creditcard 3, a pan 1, a papertowelroll 1, a spatula 1, and a winebottle 1. $\vert$ action: go to countertop 2 $\vert$ result: \ldots\ a plate 1 \ldots \\
memory context: The user is navigating through a kitchen environment to locate a plate for the purpose of cooling it on a countertop. \\
memory keywords: [`plate', `countertop', `cooling', `task', `navigation'] \\
memory tags: [`Pick-and-place workflow', `Cooling tasking', `Receptacle interaction', \ldots] \\
{[\ldots\ additional retrieved notes omitted \ldots]} \\
\midrule
\textbf{Action Trace}: \\
go to countertop 1 (no plate) $\to$ go to countertop 2 (\textbf{found plate 1}) $\to$ take plate 1 $\to$ go to fridge 1 $\to$ open fridge 1 $\to$ cool plate 1 with fridge 1 $\to$ go to countertop 1 $\to$ \textbf{move plate 1 to countertop 1} $\checkmark$ \\

\end{longtable}
\endgroup

\begingroup
\footnotesize
\setlength{\LTpre}{6pt}
\setlength{\LTpost}{8pt}
\begin{longtable}{@{}p{\linewidth}@{}}
\caption{\textbf{MeMP's case study in ALFWorld.}}
\label{tab:case-memp-alfworld}\\
\toprule
\endfirsthead
\multicolumn{1}{l}{\small\itshape Table~\ref{tab:case-memp-alfworld} (continued)}\\
\toprule
\endhead
\midrule
\multicolumn{1}{r}{\small\itshape Continued on next page}\\
\endfoot
\bottomrule[1.1pt]
\endlastfoot
\textbf{Task}: put a mug in sidetable. \\
\midrule
\textbf{Outcome}: Score = 1.0,\ \ Steps = 4 \\
\midrule
\textbf{Retrieved Memories}: \\
Here are some guidelines of how to solve the similar task: \\
- task\_name: cool some plate and put it in countertop. \\
\ \ guidelines: To solve this query, begin by identifying the location of the target plate by visiting various countertops or storage areas. Once the plate is located, take it and proceed to the fridge. \ldots \\
- task\_name: put some pillow on armchair \\
\ \ guidelines: To solve this query, begin by searching the nearby furniture, such as sofas or other seating areas, to locate a pillow. Once you have identified and taken the pillow, navigate to the target armchair. Finally, place the pillow on the armchair to complete the task. \\
{[\ldots\ additional procedural guidelines omitted \ldots]} \\
\midrule
\textbf{Action Trace}: \\
go to diningtable 1 (\textbf{found mug 1}) $\to$ take mug 1 $\to$ go to sidetable 1 $\to$ \textbf{move mug 1 to sidetable 1} $\checkmark$ \\

\end{longtable}
\endgroup

\end{document}